\documentclass[preprint,12pt]{elsarticle}

\usepackage[letterpaper,margin=1in]{geometry}
\usepackage{amsmath,amssymb,amsthm}
\usepackage{booktabs}
\usepackage{array}
\usepackage{graphicx}
\usepackage{xcolor}
\usepackage{etoolbox}
\usepackage{adjustbox}
\usepackage{rotating}   %

\usepackage[colorlinks=true,allcolors=blue]{hyperref}
\graphicspath{{pdf/}}


\newif\ifscanofloattab
\AtBeginEnvironment{table}{\scanofloattabtrue}
\AtBeginEnvironment{table*}{\scanofloattabtrue}
\BeforeBeginEnvironment{tabular}{\ifscanofloattab\begin{adjustbox}{max width=\linewidth,max totalheight=0.72\textheight}\fi}
\AfterEndEnvironment{tabular}{\ifscanofloattab\end{adjustbox}\fi}
\AtBeginEnvironment{table}{\setlength{\belowcaptionskip}{4pt}}
\AtBeginEnvironment{table*}{\setlength{\belowcaptionskip}{4pt}}
\AtBeginEnvironment{sidewaystable}{\setlength{\belowcaptionskip}{4pt}}
\usepackage{environ}
\makeatletter
\newsavebox{\scano@tabbox}
\newif\ifscano@intab
\newlength{\scano@tabw}
\newlength{\scano@tabh}
\newlength{\scano@tabcap}
\ifdefined\NewEnvironmentCopy
  \NewEnvironmentCopy{scanoorigtabular}{tabular}
\else
   
\fi
\newcommand{\scano@fulltab}{%
  \RenewEnviron{tabular}[2][c]{%
    \ifscano@intab
      \begin{scanoorigtabular}[##1]{##2}\BODY\end{scanoorigtabular}%
    \else
      \scano@intabtrue\scanofloattabfalse
      \sbox\scano@tabbox{\begin{scanoorigtabular}[##1]{##2}\BODY\end{scanoorigtabular}}%
      \setlength{\scano@tabcap}{0.72\textheight}%
      \setlength{\scano@tabh}{\dimexpr\ht\scano@tabbox+\dp\scano@tabbox\relax}%
      \setlength{\scano@tabw}{\linewidth}%
      \ifdim\scano@tabh>\scano@tabcap
        \setlength{\scano@tabw}{\dimexpr\linewidth*\scano@tabh/\scano@tabcap\relax}%
      \fi
      \ifdim\wd\scano@tabbox<\scano@tabw
        \begin{tabular*}{\scano@tabw}[##1]{@{\extracolsep{\fill}}##2}\BODY\end{tabular*}%
      \else\usebox\scano@tabbox\fi
    \fi}}
\AtBeginEnvironment{table}{\scano@fulltab}
\AtBeginEnvironment{table*}{\scano@fulltab}
\makeatother
\makeatletter
\patchcmd{\@xfloat}{\def \@fps {#2}}{\def \@fps {#2p}}{}{\typeout{SCANO: float placement patch failed}}
\makeatother

\newcolumntype{P}[1]{>{\raggedright\arraybackslash}p{#1}}
\makeatletter
\AtBeginDocument{\setbox\@tempboxa\hbox{%
  \def\scano@glyphs{ABCDEFGHIJKLMNOPQRSTUVWXYZabcdefghijklmnopqrstuvwxyz0123456789.,:;!?()[]-/+=*'`--- -- `` ''}%
  \@for\scano@size:=tiny,scriptsize,footnotesize,small,normalsize,large,Large,LARGE,huge\do{%
    \csname\scano@size\endcsname
    \normalfont\scano@glyphs\itshape\scano@glyphs\normalfont\slshape\scano@glyphs\normalfont\scshape\scano@glyphs
    \normalfont\bfseries\scano@glyphs\itshape\scano@glyphs
    \normalfont\sffamily\scano@glyphs\itshape\scano@glyphs\normalfont\ttfamily\scano@glyphs\normalfont}}}
\makeatother

\newif\ifpilot \pilottrue
\ifdefined\scanodraft \pilotfalse \fi
\newif\iffinal \finaltrue
\iffinal \pilotfalse \fi
\ifpilot
  \newcommand{\todo}[1]{\ifhmode\unskip\fi}
\else\iffinal
  \newcommand{\todo}[1]{\ifhmode\unskip\fi}
\else
  \newcommand{\todo}[1]{\textcolor{red}{[TODO: #1]}}
\fi\fi
\ifpilot
  \newcommand{\pending}[1]{\textit{[#1]}}
\else
  \newcommand{\pending}[1]{\ifhmode\unskip\fi}
\fi
\makeatletter
\newcommand{\withref}[3]{\ifpilot\@ifundefined{r@#1}{#3}{#2}\else#2\fi}
\makeatother
\ifpilot
  
\else\iffinal
  
\else
  
\fi\fi

\newtheorem{proposition}{Proposition}
\newtheorem{lemma}{Lemma}

\theoremstyle{remark}
\newtheorem{remark}{Remark}
\theoremstyle{plain}

\makeatletter
\def\els@aparagraph[#1]#2{\elsparagraph[#1]{#2}}
\def\els@bparagraph#1{\elsparagraph*{#1}}
\makeatother

\journal{Elsevier}
\ifpilot \date{} \fi

\usepackage{algorithm}
\usepackage{algpseudocode}
\begin{document}

\hypersetup{pageanchor=false}
\begin{frontmatter}
\title{SCANO: Restoring Boundary-Condition Completeness in Sampled Meshes with a Sliced-Cross-Attention Neural Operator
for 3D Thermal Fields}
\author[1]{Tianyu Li}
\author[2]{Zhiwei Cao}
\author[3]{Qingang Zhang\corref{cor1}}
\ead{qingang.zhang@ntu.edu.sg}
\author[4]{Ruihang Wang}
\author[5]{Binyang Song}
\author[3]{Yonggang Wen}
\cortext[cor1]{Corresponding author}
\affiliation[1]{organization={Interdisciplinary Graduate Programme, Nanyang Technological University},
  country={Singapore}}
\affiliation[2]{organization={School of Future Technology, Fuzhou University},
  city={Fuzhou}, country={China}}
\affiliation[3]{organization={College of Computing and Data Science, Nanyang Technological University},
  country={Singapore}}
\affiliation[4]{organization={School of Information Science and Engineering, Yunnan University},
  city={Kunming}, country={China}}
\affiliation[5]{organization={School of Mechanical and Aerospace Engineering, Nanyang Technological University},
  country={Singapore}}

\begin{abstract}

Training neural operators on million-cell meshes can exceed GPU memory limits. Models are therefore trained on
subsamples and deployed on whole meshes or selected regions. However, subsampling can distort boundary information,
degrading accuracy. We ask when boundary conditioning under sampling stays consistent with whole-mesh training, and how
to aggregate and inject boundary data. Representing the solution as an integral of boundary data against type-dependent
measures, we show that subsampling distorts this conditioning in three ways: incomplete boundary coverage or a
sample-dependent global state, face weighting by count rather than area, and compression of boundary inputs into a
low-dimensional vector. This analysis yields four measurable consistency conditions checkable before training. We
satisfy them with two backbone-agnostic modules. Boundary Quadrature Attention (BQA) aggregates all boundary faces,
typed and area-weighted, into slice tokens that every query point reads at every layer. Anchored Global State (AGS)
estimates the global state from anchors independent of the sampled points. Adding both to a scalable message-passing
and slice-attention backbone yields SCANO. We evaluate the approach on controlled and public benchmarks and a data-centre
thermal simulation dataset (DCTS) containing million-cell meshes. On both benchmarks, BQA improves every backbone,
reducing whole-mesh error up to 15-fold. On all three datasets, AGS aligns regional with whole-mesh predictions, cutting
regional error at least tenfold for most backbones on the public benchmark. Together, the modules nearly halve the best original
backbone's whole-mesh error on both benchmarks.
SCANO predicts a room-scale field in under one second, over 100 times faster than whole-room anchor self-attention.

\end{abstract}

\begin{keyword}

neural operators \sep surrogate modelling \sep boundary conditions \sep sampled training \sep quadrature consistency
\sep large meshes \sep data-centre cooling

\end{keyword}
\end{frontmatter}
\hypersetup{pageanchor=true}

\section{Introduction}
\label{sec:intro}

Neural operators learn the solution map of a family of boundary value problems. Once trained, they predict a physical
field in a fraction of the time of a numerical solver~\cite{transolver,gnot,abupt}. On engineering meshes with millions
of cells, however, the memory of a training step grows with the mesh. On a million-cell data-centre room, one pass over
the full mesh exceeds the memory of a single GPU for a message-passing backbone. It also consumes most of that memory, or exceeds it, for
the other backbones (Figure~\ref{fig:bc-stats}a). One remedy is to partition the mesh into subgraphs with
halos~\cite{xmgn}. Without domain decomposition, models are trained on subsamples of the mesh. Typical samples are random point
subsets~\cite{transolver3}, random anchors~\cite{abupt} or local subgraphs grown from random seeds~\cite{graphsaint}. After training, the operator is
deployed on the full mesh or on a local region of interest.

This sampling compromise is most damaging at the boundary. Industrial boundaries contain from a few segments to hundreds
of patches. These patches differ widely in size and physical type (Section~\ref{sec:bc}). Their prescribed values,
exchange coefficients, fluxes and heat loads govern the interior solution. A neural operator therefore has to carry this
boundary information to every point of the domain, and it does so in two steps. It first \emph{aggregates} the data of
the boundary faces into a form the network can read, such as features of the boundary nodes, one pooled summary or a set
of tokens. It then \emph{injects} this information into the domain points, either once at the input or in every layer.
Under sampled training, both steps can go wrong. A training sample captures only a fraction of the patches, so any patch
it misses is absent from the aggregation in that step. Aggregating boundary faces by count rather than by area biases the
model toward finely meshed patches. Injecting the boundary as a low-dimensional vector shared by all points erases small local
sources. A global state computed from the sample shifts with every draw. Together, these distortions make the prediction
depend on how the mesh is sampled and discretised, not on the physics alone. The GPU memory limit thus raises two
questions. How can every boundary condition reach every domain point with its proper weight in every pass? And how
should the boundary information be aggregated and injected so that this holds? We therefore ask:

\begin{quote}
\emph{Which conditions does the boundary conditioning of a neural operator need to meet so that an operator trained on
samples of the mesh agrees with one trained on the whole mesh, and how can these conditions be met at a cost that
grows linearly with the size of the mesh, on any backbone?}
\end{quote}

\label{sec:related}

Existing neural operators ingest boundary conditions through several types of encoder. Some work on whole
two-dimensional meshes. Learned function extension (LFE)~\cite{lfe} reads the boundary nodes once at the input. It
encodes their type by indicators and coefficients and weights them by count. BENO~\cite{beno} averages all boundary
nodes into a single vector per step. BOON~\cite{boon} enforces one fixed boundary type by construction.
Physics-informed graph Galerkin networks~\cite{pigalerkin} impose the boundary conditions strictly on unstructured
meshes.
LP-FNO~\cite{lpfno} lifts Dirichlet data into the domain. A generalised neural operator~\cite{gennop} maps mixed
conditions onto a latent Dirichlet representation. Other models act at the level of the layers. GNOT~\cite{gnot}
cross-attends to each normalised input function in every layer. A GNOT-based surrogate of a production data hall
represents each server and cooling unit as a single record~\cite{dcgnot}. Boundary types enter these models in several forms. They
appear as one-hot node features~\cite{meshgraphnets}, global codes~\cite{unisolver}, typed nodes of a symbolic
graph~\cite{pdeformer} or text embeddings~\cite{lorsung,operatorclip}. In our construction, the boundary type of each record
selects its measure, and the face area provides its quadrature weight.

Our construction builds on operator quadrature and on estimators for sampled data. Attention discretises an integral
operator~\cite{continuumattention}. Quadrature weights can therefore enter its normalisation and
aggregation~\cite{berner,neo,pcno,quadnorm}. Splitting a token then leaves the output
unchanged~\cite{refinementsymmetry}. Surface-quadrature acoustic operators apply the same principle and weight boundary
faces by area~\cite{clbo}. For sampled domains, M$^3$~\cite{m3} corrects the supervision measure under uneven sampling.
Inclusion probabilities and control variates correct losses and aggregations on sampled
graphs~\cite{fastgcn,vrgcn,mvsgnn,graphsaint}. Our
boundary slice tokens also draw on inducing points, latent arrays and token
merging~\cite{settransformer,perceiverio,tome}. They further relate to mean and sum aggregators~\cite{gin,pna,wagstaff}.

Several operator backbones have been designed for large meshes. AB-UPT~\cite{abupt} draws anchors from the full point
cloud, independently of the queried points. Transolver-3~\cite{transolver3} trains on random subsets and caches a
whole-mesh global state at deployment. Transolver++~\cite{transolverpp} scales slice attention to million-point
geometries. GeoTransolver~\cite{geotransolver} attends to a cached geometry context in every layer. UPT~\cite{upt},
IPOT~\cite{ipot} and GINO~\cite{gino} map the mesh onto a latent set or grid. AeroGTO~\cite{aerogto} combines message
passing with global tokens. Graph networks with hybrid local--global attention~\cite{hybridgnn} insert global
attention between message-passing layers. RIGNO~\cite{rigno} and GAOT~\cite{gaot} encode arbitrary domains on regional
graphs. GINOT~\cite{ginot} encodes arbitrary geometries from surface point clouds. Geom-DeepONet~\cite{geomdeeponet}
reads the mesh nodes with their signed distances.
Inside--outside schemes~\cite{batransolver,pinto,abswift,domino} tokenise the boundary and the interior separately on
whole meshes. Multipole and hierarchical attention~\cite{mgno,fma,hano,mano,erwin,mspt,neuralfmm} compresses the far
field. None of these architectures weights the boundary records by their area in every layer. For the global state, the closest
precedents to AGS are the anchors of AB-UPT and the deployment cache of Transolver-3. AGS adds three elements: volume
weighting, one estimator for training and deployment, and anchors that pass through the same layers as the queries.

On the application side, our work connects to learned boundary kernels and to data-centre thermal surrogates. The neural
harmonic-measure operator~\cite{nhmo} learns the harmonic-measure density of Dirichlet problems. BINO~\cite{bino} learns
boundary-integral kernels. Walk-on-stars~\cite{walkonstars} estimates mixed problems by random walks. We extend this
measure-theoretic view to mixed conditions, convection and server coupling. We also characterise how sampled training
distorts it. Existing data-centre surrogates take one of two forms. Some predict a single fixed layout from proper
orthogonal decomposition modes~\cite{samadianipod}. Others rasterise the boundary conditions onto a voxel
grid~\cite{sarkar}. Our operator instead reads typed, area-weighted face records directly on the computational mesh. It
handles diverse room layouts and meshes.

To our knowledge, no prior work has studied this question systematically: how should heterogeneous, patch-wise boundary
conditions be represented, aggregated and injected when a neural operator is trained on samples of a large mesh?
Existing boundary encoders were designed and evaluated on whole meshes. Boundary-augmented neural
operators~\cite{bno} are a recent example. They model the interaction between the boundary and the whole domain and
target unseen geometries. Scalable backbones ingest the boundary as
sampled nodes or as a count-weighted context. Studies of quadrature and of sampled estimation each address only one side
of the problem.

Operator quadrature gives this question a precise form. For linear boundary-driven problems, the solution integrates the
boundary data against measures. Each boundary's type fixes its measure~\cite{harmonicmeasure,freidlin,papanicolaou}. Prescribed and exchange
values are averaged under a probability measure. Fluxes are summed under an area-additive measure. A server couples its
inlet and exhaust through a Wentzell jump condition~\cite{ventcel}. Any boundary conditioning approximates these
measures by weighting discrete boundary faces or points (\emph{boundary records}). Under sampled training, this approximation can fail in three ways. We call them
\emph{sampling dependence}, \emph{measure mismatch} and \emph{injection and compression loss}. From these failure
classes we derive four measurable conditions. They are information completeness, mesh consistency, sample consistency and
response fidelity (Section~\ref{sec:theory}). Violations of all four can be diagnosed before training. The diagnostics
use the sampler, the face density and the response spectrum. After training, two metrics measure agreement with
whole-mesh training. The \emph{sampling gap} is the excess whole-mesh error of sampled training over whole-mesh training. The \emph{drift} is the distance
between a deployment's prediction and the model's own whole-mesh prediction.

\begin{figure*}[tp]
\centering
\includegraphics[width=\textwidth]{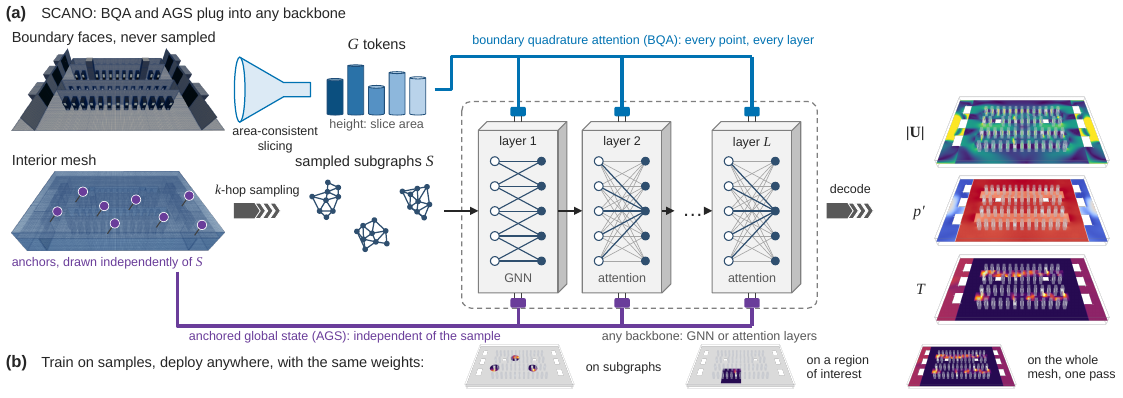}
\caption{Overview of SCANO and of the consistency conditions C1--C4 that it meets (Section~\ref{sec:conditions}). (a)
Architecture. Boundary Quadrature Attention (BQA) takes all boundary faces as unsampled typed records
$\phi_y=[\mathbf{x}_y,\mathbf{n}_y,\tau_y,g_y]$ with areas $a_y$ (C1). It pools them into $G$ area-weighted slice tokens
$z_g$ with masses $M_g$ (bar height, C2). It injects these tokens into every layer of the backbone along a normalised and an
unnormalised readout path (C4). The Anchored Global State (AGS) computes the global summary of each layer from
volume-weighted anchors. The anchors are drawn independently of the training sample $S$ (C3). (b) Deployment. A model
trained on samples is deployed on subsamples (\emph{on subgraphs}), on a region of interest with its message-passing halo, and on the whole
mesh. The whole mesh takes one pass with all cells as anchors. The example is room~1 of Figures~\ref{fig:dcts-rooms}
and~\ref{fig:dcts-roi}.}
\label{fig:method}
\end{figure*}

To meet these conditions at linear cost, we propose SCANO (Sliced-Cross-Attention Neural Operator). SCANO is built
around two plug-and-play modules. Both also attach to existing operator backbones (Figure~\ref{fig:method}). The first
module is \emph{Boundary Quadrature Attention} (BQA). BQA keeps every boundary face as an unsampled record. Each record
is typed by its role and weighted by its area. BQA aggregates the records into $G$ area-consistent slice tokens. Every
domain point attends to these tokens in every layer. A normalised path carries averaged quantities. An unnormalised path
carries fluxes and heat loads. The second module is the \emph{Anchored Global State} (AGS). AGS estimates the backbone's
global summary from volume-weighted anchors. In training, the anchors are drawn over the whole mesh, independently of
the sample. At deployment, every cell is an anchor. SCANO pairs BQA and AGS with a backbone that interleaves local
message passing and slice attention. It is designed for million-cell meshes. The two modules also integrate into six
established backbones: Transolver~\cite{transolver}, AB-UPT~\cite{abupt}, AeroGTO~\cite{aerogto}, GNOT~\cite{gnot},
GeoTransolver~\cite{geotransolver} and MeshGraphNets~\cite{meshgraphnets}.

Section~\ref{sec:results} tests each condition against its theoretical prediction. The tests cover all seven backbones,
with and without BQA and AGS, on three data sets. The first is a controlled benchmark with exact references. The second
is the public benchmark of boundary encoding~\cite{lfe}. The third is DCTS (Data Center Thermal Simulation), an
industrial suite of 500 million-cell data-centre rooms. Section~\ref{sec:dcts} examines it in depth. On the two
benchmarks, BQA reduces the whole-mesh error of every backbone. On all three data sets, AGS keeps regional predictions
aligned with the whole-mesh predictions up to the Monte Carlo error of the anchors. With both modules, AB-UPT achieves the highest
accuracy in the main comparison of each data set, and Transolver trained on room samples matches its whole-room training accuracy.
SCANO gives an accurate regional and whole-room surrogate at a cost linear in the mesh size.

This work makes three contributions. First, it establishes a consistency framework for boundary conditioning under
sampled training. From type-dependent boundary measures, we identify three failure classes. We then derive four
measurable conditions, whose violations can be diagnosed before training (Section~\ref{sec:theory}). Second, it
introduces two plug-and-play modules that meet these conditions on any backbone at linear cost. BQA attends to typed,
area-weighted boundary records in every layer. AGS keeps the domain summary consistent between training and every
deployment. Paired with a message-passing and slice-attention backbone, the two modules form SCANO for million-cell
meshes (Section~\ref{sec:method}). Third, it validates the framework from exact checks up to industrial scale. We test
the four conditions on all three data sets (Section~\ref{sec:results}). We then evaluate the modules on DCTS with paired
interventions, server inlet temperatures and end-to-end timing (Section~\ref{sec:dcts}). The controlled benchmark and
DCTS will be released with the paper.

Section~\ref{sec:theory} derives the conditions, Section~\ref{sec:method} presents SCANO and Section~\ref{sec:setup}
describes the experimental setup. Section~\ref{sec:results} tests the conditions, Section~\ref{sec:dcts} presents the
industrial case, Section~\ref{sec:discussion} discusses the results and Section~\ref{sec:conclusions} concludes.

\section{Problem formulation and theoretical analysis}
\label{sec:theory}

We first examine the boundary conditions of large meshes (Section~\ref{sec:bc}) and then formalise our central requirement. GPU memory forces training onto
subsamples. A neural operator trained on samples should still agree with one trained on the whole mesh. This section
turns that requirement into four measurable consistency conditions. The proofs and the training-free checks are in the
supplementary material. So are the map from statements to experiments (Table~\ref{tab:propositions}) and the notation
(Table~\ref{tab:notation}).

\subsection{Sampled training on large meshes and their boundary conditions}
\label{sec:bc}

\begin{figure*}[t]
\centering
\includegraphics[width=\textwidth]{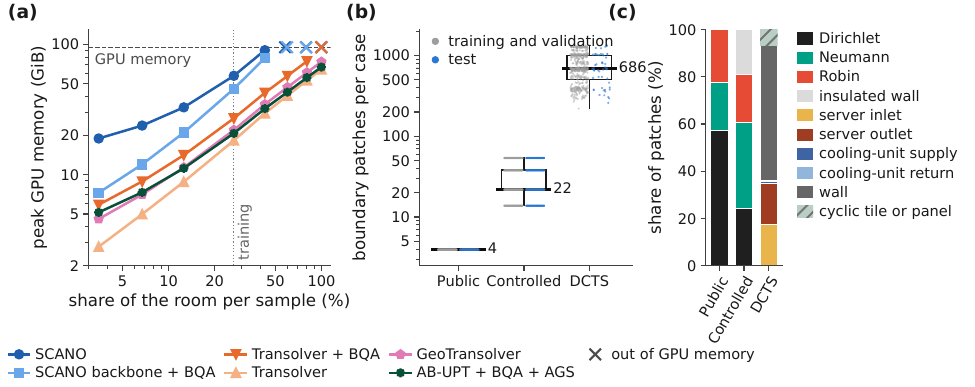}
\caption{Memory of sampled training and boundary statistics of the three data sets. (a) Peak GPU memory of one optimiser step on
a DCTS room against the share of the room per sample. Samples are two-hop subgraphs at seed ratios of
0.5--50\%, and the last point is the full room (log axes). Crosses mark runs that exceed the GPU memory. The dashed line marks
the available memory of a single GPU. The dotted line marks the training setting. (b) Number of boundary patches per
case in the training and validation (grey) and test (blue) splits, showing the median, quartiles and range. (c) Share of
the patches of each boundary type or physical role.}
\label{fig:bc-stats}
\end{figure*}

On a large mesh, the memory of a training step grows with the share of the mesh in the sample. This holds for every
model (Figure~\ref{fig:bc-stats}a). On a DCTS room of about a million cells, a step over the whole room takes most of a
GPU's memory for every backbone. For the SCANO backbone, it exceeds that memory (Section~\ref{sec:backbone}). Its message
passing stores features on every mesh edge. It therefore needs roughly three times the memory of Transolver on the same
sample. It exceeds a single GPU once a sample covers more than half of the room. Transolver + BQA also exceeds a single GPU on the whole room
without activation checkpointing. On still larger meshes, a full-mesh
pass exceeds a single GPU for every backbone. Sampled training is therefore the starting point of this study. On DCTS,
every model is trained on the same two-hop subgraphs, grown from seeds at 5\% of the room. The comparison is thus
paired.

Each training sample, however, captures only a fraction of the boundary. Engineering boundaries are large. They comprise
many patches of different physical types and sizes, and each patch carries its own data. Across our three data sets
(Table~\ref{tab:datasets}, Figure~\ref{fig:bc-stats}b,c), the patch count grows quickly. The public benchmark has four
Dirichlet, Neumann and Robin segments on a circle. The controlled benchmark has a few tens of typed wall patches. A DCTS
room has several hundred patches of six physical roles. Face areas within a single patch also span orders of magnitude.
As the geometry grows more complex, a sample observes a smaller share of the patches. The patches it misses most often
are the smallest ones, such as a single server exhaust (\emph{server outlet}) or a cooling-unit grille. These are exactly the patches whose
changes matter most to the engineer.

\subsection{The requirement: consistency with whole-mesh training}
\label{sec:consistency}

Let $\Omega_h$ be the cells of a mesh with volumes $w_x$, $u(x)$ the reference solution and $\hat u_\theta(x;S)$ the
prediction at a cell $x$ when the model is given the sample $S\subseteq\Omega_h$ with the boundary input of that
step. Training on the whole mesh minimises
\begin{equation}
  \mathcal{L}(\theta) = \sum_{x\in\Omega_h} w_x\,\ell\bigl(\hat u_\theta(x;\Omega_h),\,u(x)\bigr);
  \label{eq:loss-mesh}
\end{equation}
sampled training draws $S$ from a design that includes $x$ with probability $\pi_x$ and minimises
$\mathcal{L}_S(\theta) = \sum_{x\in S} (w_x/\pi_x)\,\ell(\hat u_\theta(x;S),u(x))$. The model is \emph{sample-invariant}
at $\theta$ if $\hat u_\theta(x;S)=\hat u_\theta(x;\Omega_h)$ for every sample $S$ that contains $x$.

\begin{proposition}[consistency of sampled training]
\label{prop:sampling}
Let the loss be additive over points with a normalisation
that does not depend on the sample. (i) If the model is sample-invariant and $\pi_x>0$ wherever the whole-mesh term is
not zero, then $\mathbb{E}_S\mathcal{L}_S=\mathcal{L}$, and the gradients agree if the invariance holds near $\theta$;
with weights renormalised over the sample the objectives agree up to $O(1/|S|)$. (ii) If it is not,
$\mathbb{E}_S\mathcal{L}_S=\sum_x w_x\mathbb{E}[\ell(\hat u_\theta(x;S),u(x))\,|\,x\in S]$, in general a different
objective however long the model is trained; for a global state computed from the sample, the gap does not shrink as the
sample grows when the limit of the state differs from the whole-mesh state. (iii) If the only sample-dependent part is a
global state estimated from $N_a$ anchors that are drawn independently of the sample and carry their whole-mesh
features, the model is sample-invariant given the anchors, and training with random anchors differs from deployment
with the whole-mesh state only by the Monte Carlo error of the anchors, of order $1/N_a$ for a loss twice
differentiable in the state. (iv) In general, loss weights alone do not give consistency.
\end{proposition}
Part~(i) is the Horvitz--Thompson identity~\cite{horvitzthompson,sarndal}.

Loss weights alone cannot make sampled training consistent (part~(iv)). Consistency must come from the architecture. The
boundary information that reaches a point must not depend on the sample. Neither may the global state that the point
reads. A model whose state is independent of the sample then keeps its accuracy and its predictions across deployments,
up to the error of its anchors. A state computed from the sample drifts with every draw (Section~\ref{sec:test-c3}).

\subsection{A measure representation of mixed boundary data}
\label{sec:measure}

Consider $-\nabla\cdot(\kappa\nabla u) + \mathbf{b}\cdot\nabla u = 0$ in a bounded Lipschitz domain $\Omega$ with constant
$\kappa$, $\nabla\cdot\mathbf{b} = 0$, $\mathbf{b}\cdot\mathbf{n} = 0$ on walls, and a boundary split into Dirichlet
($D$), Neumann ($N$) and Robin ($R$) patches with $D\cup R$ of positive measure. A diffusion started at $x$ with drift
$-\mathbf{b}$, reflected at walls and at $N$ and $R$ patches (and killed at a rate on $R$), and stopped on $D$ gives
\begin{equation}
  u(x) = \int_{D} u_D\,\mathrm{d}\mu^{\mathrm{exit}}_x + \int_{R} u_\infty\,\mathrm{d}\mu^{\mathrm{kill}}_x
       + \int_{N} \frac{q}{\kappa}\,\mathrm{d}\nu^{\mathrm{loc}}_x ,
  \label{eq:measure}
\end{equation}
with Dirichlet values $u_D$, external Robin values $u_\infty$ and Neumann fluxes $q$. The exit and killing measures
together have mass one. The local-time measure $\nu^{\mathrm{loc}}_x$ is
unnormalised~\cite{harmonicmeasure,freidlin,papanicolaou,basshsu}. Prescribed and external values thus enter as an average.
Fluxes enter as an area-additive sum. The type selects the measure, so values without their type cannot determine the
solution. The finite-volume operator of the controlled benchmark is an M-matrix with the same structure, term by term
(Section~\ref{app:measure} of the supplementary material).

\begin{remark}[a server as a jump condition]
\label{prop:jump}
A server's exhaust returns its mean inlet temperature raised by $\Delta T_s=P/(\rho_{\mathrm{air}} c_p \dot V)$. This is
a non-local coupling of two patches. With the flow frozen, it takes the form of the pure-jump term of a Wentzell
(Venttsel') boundary condition~\cite{ventcel}. A path that reaches the exhaust restarts on the server's inlet. The jump
thus keeps the convex structure, and $\Delta T_s$ enters additively, like a flux. For buoyant RANS flow in a data
centre, this representation holds for temperature under frozen flow. Beyond that, it guides the design.
\end{remark}

\subsection{How consistency breaks: three failure classes}
\label{sec:failures}

\begin{figure*}[t]
\centering
\includegraphics[width=\textwidth]{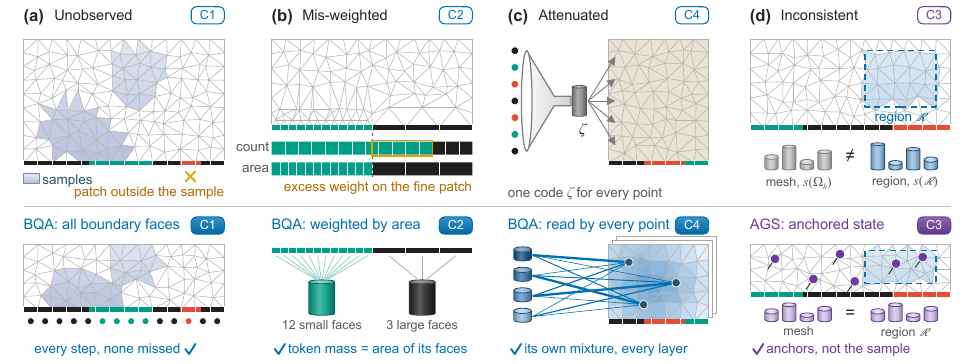}
\caption{Distortions of the boundary measure under sampled training (top row), and how SCANO meets conditions C1--C4 (bottom row, filled
tags). The mesh has Dirichlet (black), Neumann (teal) and Robin (red) patches. (a) \emph{Unobserved}: sampling the
boundary nodes (shaded) leaves out the patches that are not sampled. BQA reads all boundary faces in every step (C1).
(b) \emph{Mis-weighted}: weighting by count shifts the unit mass toward finely meshed patches (orange outline). BQA
weights the records by face area $a_y$ in slice tokens of mass $M_g$ (C2). (c) \emph{Attenuated}: global modulation
(FiLM, adaLN) broadcasts one rank-$d$ code $\zeta$ to all points. With BQA, each point attends to the slice tokens with
its own weights in every layer (C4). (d) \emph{Inconsistent}: a global state computed from the sample shifts with the
query region $\mathcal{R}$. AGS estimates the state from anchors independent of the sample (pins; C3).}
\label{fig:failures}
\end{figure*}

To first order, boundary conditioning replaces the continuous measures of \eqref{eq:measure} with discrete weights on
boundary records. These weights should not depend on the training sample (Proposition~\ref{prop:sampling}). Across
meshes, they should not depend on how the boundary is split into faces either. Under sampled training, four
distortions can violate this requirement. They fall into three failure classes (Figure~\ref{fig:failures}). Under
\emph{sampling dependence}, a patch goes \emph{unobserved}, or a state computed from the sample makes the prediction
\emph{inconsistent}. Under \emph{measure mismatch}, the records are \emph{mis-weighted}. Under \emph{injection and
compression loss}, the boundary signal is \emph{attenuated}. Each distortion leaves a signature. It can be predicted
before training or measured on the data. The following statements make this precise.

\begin{proposition}[visibility bound]
\label{prop:visibility}
For a linear problem $u=\sum_p K_p g_p$ with independent
patch data $g_p$ of covariances $\Sigma_p$ and a design independent of the data, let $H(S)$ be the patches none of whose
records reaches the input of sample $S$. Every prediction $\hat u$ that depends only on the input satisfies
\begin{equation}
  \mathbb{E}\,\|\hat u-u\|_Q^2 \;\ge\; \mathbb{E}_S \sum_{p\in H(S)} \operatorname{tr}\bigl(K_p\Sigma_pK_p^{*}\bigr)\big|_Q .
  \label{eq:visibility}
\end{equation}
When the boundary enters only as sampled nodes, a patch of $n_p$ faces is hidden from a uniform subset of fraction
$\rho$ with probability close to $(1-\rho)^{n_p}$, and the information lost at $x$ is the data change times the hidden
patch's measure, largest nearby and downstream.
\end{proposition}
In a data-centre room, the patches hidden most often are the server exhausts. They are many, small and adjacent. They
are also the heat sources (tested in Section~\ref{sec:test-c1}).

\begin{proposition}[drift of a sample-dependent global state]
\label{prop:drift}
Consider a global state pooled over the sample, $s_j(S)=\sum_{x\in S} c(x)\, \omega_j(x) f(x)/\sum_{x\in S} c(x)\,
\omega_j(x)$ (assignment weights $\omega_j$, features $f$), with context weights $c(x)=1$ (count) or $c(x)=w_x$
(volume). Under a design with inclusion probabilities $\pi_x$, $s_j(S)$ tends to the state weighted by $\pi_x c(x)$.
This limit equals the volume-weighted whole-mesh state for every feature if and only if $\pi_x c(x)\propto w_x$. It
equals the backbone's own whole-mesh state, with weights $c(x)$, if and only if $\pi_x$ is uniform. Restricting the
sample to a region shifts the limit unless the features are constant over the mesh. To first order, the change of the state between two deployments is
linear in the difference of the measures $\pi_x c(x)$, and the drift of the prediction follows it; a state from
independently drawn anchors is fixed by the whole-mesh cache and fluctuates as $N_a^{-1/2}$ with fresh anchors.
\end{proposition}
The limit is that of a ratio estimator~\cite{sarndal} (tested in Section~\ref{sec:test-c3}).

\begin{proposition}[count against area weights]
\label{prop:count}
Consider an aggregation at $x$ that assigns to record $y$ (value $g_y$) the weight $e^{\lambda_x(y)}$, and let
$K(x,\cdot)$ be the area density of the boundary measure $\mu_x$ it approximates, a probability measure on the
aggregated records. If the logits equal the log density,
$\lambda_x=\log K(x,\cdot)$, the aggregation weighted by the face areas $a_y$ is a self-normalised Riemann sum of $\int
g\,\mathrm{d}\mu_x$, second order for smooth data; in the limit of dense records, the count-weighted aggregation errs by
\begin{equation}
  \frac{\int g\varrho\,\mathrm{d}\mu_x}{\int \varrho\,\mathrm{d}\mu_x} - \int g\,\mathrm{d}\mu_x
  = \frac{\operatorname{Cov}_{\mu_x}(g,\varrho)}{\mathbb{E}_{\mu_x}[\varrho]}
  \label{eq:count-error}
\end{equation}
for a record density $\varrho$, vanishing only when $\varrho$ is constant or uncorrelated with $g$ under $\mu_x$.
\end{proposition}
Attention can be viewed as a discretised integral operator~\cite{continuumattention,neo}. This view places quadrature
weights directly into the identity above. It predicts a computable error plateau for count weighting on graded records.
Area weighting, in contrast, stays invariant when a record is copied into pieces that share its area, for any learned
logits. It converges when faces are split or the mesh is refined (Proposition~\ref{prop:records}(iii);
Section~\ref{sec:test-c2}).

\begin{proposition}[a global code limits the response]
\label{prop:code}
If the model reads the boundary data only through a code
$\zeta=E(g)\in\mathbb{R}^d$ with a continuous encoder, the worst error on a sphere of radius $r$ around $g_0$ is at
least $\sigma_{d+1}r-O(r^3)$, with $\sigma_i$ the singular values of the derivative of the solution operator, and for a
linearised model the mean squared error under isotropic perturbations of size $t$ is at least $t^2\sum_{i>d}\sigma_i^2$.
\end{proposition}
This is a classical width bound~\cite{devorehowardmicchelli,eckartyoung}. It has stable and Lipschitz
variants~\cite{cohenstable,lipschitzwidths}. The bound applies to every model that reads the boundary only through a
global code. Examples are feature-wise linear modulation (FiLM~\cite{film}) and adaptive layer normalisation (adaLN;
Section~\ref{sec:test-c4}). By the same mechanism, adjacent small patches pooled into one token cannot be resolved one
by one (Proposition~\ref{prop:unmix}).

\subsection{An error decomposition}
\label{sec:decomposition}

Write the prediction as $\hat u_\theta(x;s,\beta)$, with $s$ the global state the point reads and
$\beta=\sum_y a_y\delta_{\phi_y}$ the weighted empirical measure of the boundary records that reach it (features
$\phi_y$, weights $a_y$; for boundary values given as node features, those of the sampled boundary faces). A deployment,
whether the whole mesh, a subgraph, a region or another mesh, determines $(s_{\mathrm{dep}},\beta_{\mathrm{dep}})$; the
reference $(s^\ast,\beta^\ast)$ uses every cell and every record with its area. Let $\theta^\ast$ minimise
\eqref{eq:loss-mesh} over the model class, $\hat\theta$ be the sampled-training solution, and $\|\cdot\|_w$ be the
volume norm.

\begin{lemma}[error decomposition]
\label{lem:decomposition}
If $\hat u_{\hat\theta}$ is Lipschitz in the state (Euclidean norm over layers and tokens) and in the boundary input
(bounded-Lipschitz distance $d_{\mathrm{BL}}$ of measures), with constants $L_s$ and $L_\beta$ that hold uniformly in the
other argument, then for every deployment
\begin{equation}
  \bigl\|\hat u_{\hat\theta}(\cdot;s_{\mathrm{dep}},\beta_{\mathrm{dep}})-u\bigr\|_w \;\le\;
  \underbrace{\varepsilon^\ast}_{\text{class}} \;+\; \underbrace{\gamma(\hat\theta)}_{\text{training}}
  \;+\; \underbrace{L_s\,\|s_{\mathrm{dep}}-s^\ast\|}_{\text{state}}
  \;+\; \underbrace{L_\beta\,d_{\mathrm{BL}}(\beta_{\mathrm{dep}},\beta^\ast)}_{\text{boundary}},
  \label{eq:decomposition}
\end{equation}
with $\varepsilon^\ast=\|\hat u_{\theta^\ast}(\cdot;s^\ast,\beta^\ast)-u\|_w$ and
$\gamma(\hat\theta)=\|\hat u_{\hat\theta}(\cdot;s^\ast,\beta^\ast)-u\|_w-\varepsilon^\ast$.
\end{lemma}
\emph{Proof.} The triangle inequality through $\hat u_{\hat\theta}(\cdot;s^\ast,\beta^\ast)$ and the Lipschitz
bounds.\hfill$\square$

Propositions~\ref{prop:sampling}--\ref{prop:code} characterise each term of~\eqref{eq:decomposition}. The Lipschitz
constants of a trained network are not controlled a priori. The lemma therefore serves to organise the total error, so
that each term can be isolated in experiments. The whole-mesh error of a model trained on the whole mesh estimates
$\varepsilon^\ast+\gamma$. The \emph{sampling gap} between sampled and whole-mesh training isolates the bias of the
sampled objective. The drift across deployments captures the state and boundary terms. The decay of that drift with the
anchor count isolates the state term.

\subsection{Complete boundary conditioning}
\label{sec:conditions}

Boundary conditioning is \emph{complete} when four design criteria hold. Each criterion comes with a diagnostic that
needs no training and with an empirical check on the trained model.
\begin{description}
  \item[C1 Information completeness.] Under every deployment, the input determines all boundary conditions, that is,
        the geometry, the type of each field and the data of every patch. It is measured by the error floor of subset
        deployments (Proposition~\ref{prop:visibility}) and by the response to type switches
        (Proposition~\ref{prop:records}(v)).
  \item[C2 Mesh consistency.] The aggregation of the records is a consistent quadrature of the boundary integral,
        invariant to copying a record into pieces that share its area, and convergent when faces are split or
        refined. It is measured by the drift
        under record transformations and refinement (Propositions~\ref{prop:count} and~\ref{prop:records}(iii)).
  \item[C3 Sample consistency.] The prediction at a point does not depend on the other points of its sample, up to
        the Monte Carlo error of the anchors. It is measured by the drift across deployments and its decay with the
        anchor count (Propositions~\ref{prop:sampling} and~\ref{prop:drift}).
  \item[C4 Response fidelity.] The injection path does not cap the response to the boundary, so no code has a lower
        dimension than the data. It is measured by the response error by patch area and kind of intervention
        (Propositions~\ref{prop:code} and~\ref{prop:unmix}).
\end{description}
C1 and C2 together eliminate the boundary term in~\eqref{eq:decomposition}. C3 eliminates the state term. Under
design-consistent weights, it also removes the objective bias in the training term. C4 removes the rank floor on the
class term. A trivial predictor that ignores the boundary would look perfectly consistent. Deployment consistency is
therefore always evaluated together with the response to boundary interventions. Section~\ref{sec:results} tests C1--C4
in this order.

\label{sec:record-theory}
These conditions translate directly into requirements on the components. At the record level, every record must carry
its boundary type. Area weighting must provide the quadrature measure. The slice-token means must include the face
areas, so that copy invariance and refinement convergence hold (Propositions~\ref{prop:records} and~\ref{prop:slices},
Section~\ref{supp:records}). BQA meets C1 and C2. It reads unsampled, area-weighted records in every layer, so C1 and
the record-level invariance of C2 hold by construction. AGS meets C3 on any backbone. It estimates the global state from
anchors that do not depend on the sample (Section~\ref{sec:method}). C3 then holds up to the Monte Carlo error of the
anchors whenever each anchor carries its exact whole-mesh features, for instance when anchors bypass local message passing. On
message-passing backbones, a training sample also truncates the receptive field near its border. This affects both the
query and anchor subgraphs. A halo as deep as the message-passing receptive field removes this truncation for the queries
at deployment. In training, border truncation accounts for little of the sampling gap, and evaluating SCANO's state from fresh
anchors instead of every cell changes its error only slightly (Sections~\ref{supp:r7-halo} and~\ref{supp:r7-deploy}). C4
holds exactly when each record forms its own token, as on the two benchmarks. With $G$ slice tokens, the tokens form a
code whose width is $G$ times the token width, so Proposition~\ref{prop:code} applies with that width. $G$ then sets the
resolution budget of the boundary response, at a cost linear in $G$ (Proposition~\ref{prop:unmix}).

\section{Methodology: SCANO}
\label{sec:method}

SCANO conditions a neural operator on its boundary through two modules. The first is \emph{boundary quadrature
attention} (BQA, Sections~\ref{sec:records} and~\ref{sec:slices}). The second is an \emph{anchored global state} (AGS,
Section~\ref{sec:global-state}). Both are carried by the SCANO backbone (Section~\ref{sec:backbone} and
Figure~\ref{fig:method}). BQA reads every boundary face as a record in every pass and every layer. AGS estimates the
global state from anchors drawn independently of the sample. Boundary conditioning thus stays complete in every training
pass, whatever the sample. The modules also attach to other backbones (Section~\ref{sec:hosts}). We denote the resulting
models \emph{backbone + BQA} and \emph{backbone + BQA + AGS}. The SCANO backbone with BQA alone computes its state from
the sample. We call it the \emph{SCANO backbone + BQA}.

\subsection{Typed boundary records}
\label{sec:records}

Only the domain points are sampled. Every boundary face $y$ enters SCANO in every step as a record $\phi_y =
[\mathbf{x}_y, \mathbf{n}_y, \tau_y, g_y]$. The record holds the face centroid, the outward normal, the boundary type
and the value. The face area $a_y$ is not an input feature. It is the quadrature measure under which the records are
aggregated. Splitting a face therefore refines the quadrature without changing the information. The type $\tau_y$ fixes
the physical meaning of $g_y$. On a prescribed boundary, $g_y$ is the imposed datum. On a zero-gradient boundary, it is
an initial iterate. On a server exhaust, it is an initial estimate that includes the heat load. Value and type are
therefore always read together. On DCTS, $\tau_y$ encodes the role of the patch. There $g_y$ holds the solver's
first-iteration values. These equal the prescribed data wherever data are imposed.

A server exhaust realises the jump coupling of Remark~\ref{prop:jump}. Its records carry the server's heat load and
temperature rise $\Delta T_s$. On every backbone with BQA, and in the LFE extender, these records are weighted by the
face area times the normalised exhaust velocity magnitude $|\mathbf{U}|$. This product is proportional to the mass flux
through the face. We use it as the measure of the jump, a modelling choice that follows the exhaust flow. The heat load
enters the unnormalised readout described below. In all methods evaluated on DCTS, each domain point also receives its offset
to the nearest exhaust (Section~\ref{sec:dcts-protocol}).

\subsection{Area-consistent slice tokens and two readouts}
\label{sec:slices}

Reading every record from every point costs the product of points and faces. This is prohibitive for a million-cell room
with half a million boundary faces. SCANO therefore aggregates the records once per forward pass into $G$ boundary
tokens. For each attention head, the lifted record $h_y$ is assigned softly to $G$ slices. Each slice receives a mass
and a token:
\begin{gather}
  \omega_g(y) = \operatorname{softmax}_g\!\bigl(W h_y / \vartheta\bigr), \qquad M_g = \sum_y a_y\, \omega_g(y), \\
  z_g = \frac{\sum_y a_y\, \omega_g(y)\, h_y}{M_g + \epsilon\, M_{\mathrm{tot}}}, \qquad M_{\mathrm{tot}} = \sum_g M_g .
\end{gather}
The masses thus partition the boundary area. Each token is an area-weighted moment of the records, as in attention
viewed as a discretised integral operator~\cite{continuumattention,neo}. In every layer $l$, each domain point $x$
attends to the tokens along two paths. A normalised path serves averages. An unnormalised path serves totals such as a
heat flux or
a server's temperature rise:
\begin{gather}
  o_l(x) = \sum_g \operatorname{softmax}_g\!\Bigl(\frac{\mathrm{q}_l(x)\cdot \mathrm{k}_l(z_g)}{\sqrt{d_{\mathrm{k}}}} + \log M_g\Bigr)\,
  \mathrm{v}_l(z_g), \\
  o^{N}_l(x) = \sum_g \frac{M_g}{M_{\mathrm{ref}}}\,
  \exp\Bigl(\frac{\mathrm{q}^{N}_l(x)\cdot \mathrm{k}^{N}_l(z_g)}{\sqrt{d_{\mathrm{k}}}}\Bigr)\, \mathrm{v}^{N}_l(z_g).
\end{gather}
Here $W$ is a learned projection and $\vartheta$ a learned temperature. $d_{\mathrm{k}}$ is the width of an attention head.
$M_{\mathrm{ref}}$ is a fixed reference area of the order of a case's boundary area. The log-mass weights each slice by
the area it represents. A softmax normalises its weights and cannot represent totals. The unnormalised path supplies
them. The tokens depend on the records alone. The domain therefore reads the boundary without altering it.

The implementation verifies three properties of this construction. First, replacing any record by several copies that share
its area leaves both readouts unchanged in every backbone (\emph{copy invariance}). Second, hard assignment by patch
yields patch-level records. One slice per face yields per-record attention with a log-area logit (\emph{two limits}).
Third, the cost is one pass over the faces and one over the points, each multiplied by $G$ (\emph{linear cost}).

The slice count $G$ sets the resolution budget of BQA. The two benchmarks have only a few thousand boundary faces.
There, every face forms its own token and is read along the normalised path. On DCTS, heat loads enter as totals. We set
$G = 64$ with both readouts. Whole-room accuracy and the response to a doubled rack load stay constant from $G = 16$ to
$1{,}024$. Only the training time grows with $G$ (Table~\ref{tab:dcts-budget}). Our comparisons also include plain slice
aggregation (\emph{Slice aggregation} in the tables). It pools records by count into tokens, and the domain slices read these tokens only once. BQA differs in
three ways. It weights the records by area. It injects them at every point in every layer. It also provides an
unnormalised path for totals.

\subsection{An anchored global state}
\label{sec:global-state}

The global summary of the backbone is the second route by which the sample can enter the prediction. A slice-attention
backbone summarises the domain in each layer $l$ by a global state of one token per slice. These tokens are computed over a context set
$\mathcal{C}$ with context weights $c(x)$,
\begin{equation}
  s_{l,j} = \frac{\sum_{x\in\mathcal{C}} c(x)\, \omega_{l,j}(x)\, f_l(x)}
                    {\sum_{x\in\mathcal{C}} c(x)\, \omega_{l,j}(x)},
\end{equation}
and every queried point reads them. Standard backbones use the training sample as $\mathcal{C}$. The summary then shifts
with the sampling design. This produces the \emph{inconsistent} failure (Figure~\ref{fig:failures}d).

AGS decouples the context set from the query sample. In each training step, $\mathcal{C}$ is an anchor sample. The
training sampler draws it independently of the query sample, or uses the full mesh when the mesh is small. The anchors
are weighted by cell volume. Each global token is thus a self-normalised Monte Carlo estimate of the volume-weighted
domain mean. When the backbone has message passing, the anchor sample passes through the same message-passing layers as
the queries. It does so on its own subgraph, without edges to the queries. The global state is therefore built from
features of the same kind as the queries. Without message passing, a uniform subset of cells serves as the anchors.

At whole-mesh deployment, the anchor sample is the entire mesh. The global state is computed once, and the full mesh is
predicted in a single pass. On a subgraph or a region of interest, the model draws a fresh anchor sample, as in
training. For anchors that carry their whole-mesh features, only the Monte Carlo error of the anchors remains. Their
count $N_a$ is thus a second budget beside $G$ (Proposition~\ref{prop:drift}). AGS estimates one domain summary for
training and for all deployments (Proposition~\ref{prop:sampling}(iii)). This differs from AB-UPT~\cite{abupt} given
only a subsample. AB-UPT then draws its anchors within the sample, so its estimate depends on the sample. It also
differs from Transolver-3~\cite{transolver3}. Transolver-3 caches a whole-mesh state only at deployment. Its state
therefore departs from training unless the samples cover the mesh uniformly.

\subsection{The SCANO backbone}
\label{sec:backbone}

In every layer, the SCANO backbone interleaves local message passing over the mesh with the slice attention of
Section~\ref{sec:global-state}. It also applies the two boundary readouts of Section~\ref{sec:slices}. The message
passing uses the graph-network layer of the finite-volume graph network~\cite{fvgn,genfvgn}. It sits in an
encode--process--decode architecture~\cite{meshgraphnets}. Message passing gives each point the fine-scale resolution of
its mesh neighbourhood. Slice summaries alone lack this resolution. A point whose receptive field lies inside the sample
sees exactly the neighbourhood it has on the full mesh. $k$-hop subgraphs, compact regions and the whole mesh all keep
the required mesh graph. All baselines are matched to SCANO's parameter count of about $0.8\,\mathrm{M}$. Removing the
message-passing layers reduces SCANO to a Transolver-style backbone with BQA and AGS. This model serves as our
message-passing ablation.

\subsection{SCANO on other backbones}
\label{sec:hosts}

BQA and AGS also work as plug-and-play modules beside the native layers of a backbone (Figure~\ref{fig:method}a). BQA is
applied after each layer through the residual update $f_l(x) \leftarrow f_l(x) + o_l(x) + o^N_l(x)$ on a
layer-normalised state. The backbone keeps its native positional encoding and initialisation. We attach BQA to
Transolver, AB-UPT, AeroGTO and GeoTransolver. We also attach it to GNOT, whose native cross-attention already reads
every record weighted by count. Finally, we attach it to MeshGraphNets, which otherwise sees the boundary only through
sampled boundary nodes. AGS replaces the native global state in every backbone that has one. These states are the
slice-attention summaries of Transolver, AeroGTO and GeoTransolver, the attention summary of GNOT and the anchor set of
AB-UPT. On the graded mesh of the controlled benchmark, AGS draws AB-UPT's anchors over all mesh nodes. It weights them
by cell volume. Keeping the count weights instead gives AB-UPT + BQA with whole-mesh anchors by count.

Where BQA adds parameters, we reduce the backbone's width to keep the total parameter count matched. Each backbone with
BQA is then compared with the original (native) backbone under the same records, sampler and training protocol
(Section~\ref{supp:impl} of the supplementary material).

\section{Experimental setup}
\label{sec:setup}

We first describe the three data sets and then the compared models, the training protocol and the metrics.

\subsection{Data sets}
\label{sec:data}

\begin{figure*}[t]
\centering
\includegraphics[width=\textwidth]{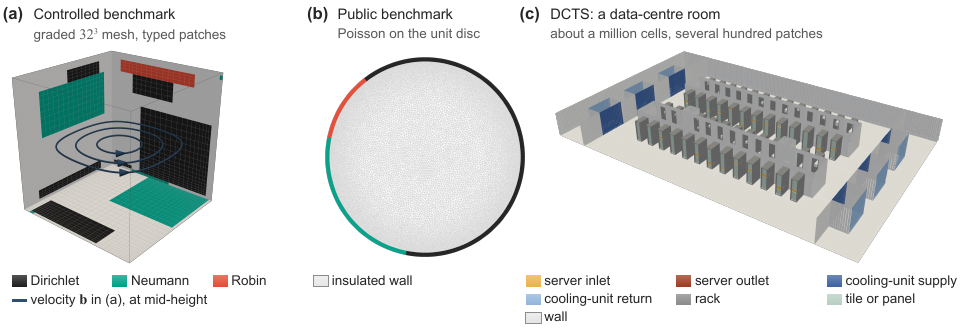}
\caption{Mesh and boundary of a representative case of each data set. (a) Controlled benchmark. The unit cube is shown open on
its three near walls. The graded boundary faces are coloured by patch type. The panel also shows the velocity $\mathbf{b}$ at
mid-height (median case of Figure~\ref{fig:r0s-fields}). (b) Public benchmark. The interior triangulation is shared by
all cases. The boundary segments are coloured by type (median case of Figure~\ref{fig:olbc-fields}). (c) DCTS room. The
CFD boundary mesh is coloured by the role of each component. The ceiling and two near walls are left out (room~1 of
Figure~\ref{fig:dcts-rooms}).}
\label{fig:datasets}
\end{figure*}

Table~\ref{tab:datasets} summarises the three data sets. Figure~\ref{fig:datasets} shows one boundary of each.

\begin{table}[t]
\centering
\footnotesize
\setlength{\tabcolsep}{3pt}
\caption{The three data sets. \emph{Patches} is the number of distinct boundary conditions per case (median and range;
Figure~\ref{fig:bc-stats}). \emph{Records} are the boundary faces or points that BQA reads. \emph{Cases} gives the
training, validation and test splits. \emph{Coverage} is the fraction of the mesh in one training sample.}
\label{tab:datasets}
\begin{tabular}{P{0.16\linewidth}P{0.25\linewidth}P{0.25\linewidth}P{0.27\linewidth}}
\toprule
& Controlled benchmark & Public benchmark & DCTS \\
\midrule
Problem & linear convection--diffusion in the unit cube, recirculating velocity $\mathbf{b}$ & Poisson on the unit disc & steady buoyant RANS (Boussinesq, realizable $k$--$\varepsilon$) \\
Mesh & graded $32^3$ (32,768 cells); deployment mesh $40^3$, graded in reverse & 16,859 points, shared by all cases & 0.88--1.23 million cells, one mesh per room \\
Patches & 22 (14--54); Dirichlet, Neumann, Robin, insulated wall & 4 segments; Dirichlet, Neumann, Robin & 686 (219--1,306); six roles \\
Records & 6,144 faces & 1,000 points & about half a million faces \\
Cases & 2,048 / 256 / 512 & 2,744 / 256 / 256 & 400 / 50 / 50 \\
References & exact (response basis) & finite elements (benchmark) & steady RANS solutions (OpenFOAM) \\
Interventions & 256 exact pairs & --- & 20 held-out test pairs (200 for training) \\
Coverage & 2--58\% & 5--76\% & about 40\% of the cells and 5\% of the boundary faces (seeds at 5\%, two hops) \\
\bottomrule
\end{tabular}
\end{table}

\subsubsection{The controlled benchmark}
\label{sec:r0s-data}

The controlled benchmark provides what industrial data cannot. It has exact reference solutions and single-patch
boundary interventions. The count, area and type of its boundary patches are set by design. It solves the linear
convection--diffusion problem of Section~\ref{sec:measure} in the unit cube. A fixed recirculating velocity $\mathbf{b}$
carries the influence of each patch downstream. Rectangular patches sit on otherwise insulated walls. Some are small and
strongly driven openings (Figure~\ref{fig:datasets}a; Table~\ref{tab:r0s-data} of the supplementary material). Each case
is assembled from an exact response basis on a graded mesh. We verified this basis for superposition, conservation, the
maximum principle and second-order convergence. Each of 64 test cases undergoes four exact single-patch interventions in
four area strata. This gives 256 paired cases (Table~\ref{tab:r0s-data}).

By construction, this benchmark isolates each failure class. Patches are missed with a known probability
(Proposition~\ref{prop:visibility}). Axis-graded faces separate count weighting from area weighting. A reverse-graded
$40^3$ deployment mesh (\emph{Other mesh} in Table~\ref{tab:r0s-main}) tests mesh consistency (Proposition~\ref{prop:count}). The exact response spectrum sets the rank floor
(Proposition~\ref{prop:code}). A type switch changes what a patch prescribes without changing its geometry. Training
samples come from two standard families, $k$-hop subgraphs~\cite{graphsaint} and uniform
subsets~\cite{transolver3,abupt}. Methods are compared within each family. A $k$-hop
sample draws seed nodes at ratio $\rho$ over cells and boundary faces. It then adds all cells within $k$ graph hops. A
boundary face therefore enters only when it is drawn as a seed. At $\rho = 0.12$ with one hop, a sample holds 58\% of
the cells but only 12\% of the faces (Table~\ref{tab:r0s-sampling}, and Section~\ref{supp:r7-sampling} also tests a
face-inclusive sampler). A uniform sample selects a random fraction $\rho$
of cells and faces, without a mesh graph. On uniform samples, the SCANO backbone runs without message passing.

\subsubsection{The public benchmark}
\label{sec:olbc-data}
\label{sec:olbc-protocol}

The public benchmark is the mixed-boundary Poisson problem on the unit disc introduced with LFE~\cite{lfe}. Its boundary
is split into Dirichlet, Neumann and Robin segments. Their layout and values vary across cases on a shared point cloud
(Figure~\ref{fig:datasets}b). The prediction error therefore reflects the boundary conditioning directly. The boundary
points are nearly uniform in arc length, so count and area weights almost coincide. The interior points are not uniform.
The volume measure of the global state therefore still matters.

We evaluate all applicable methods of Table~\ref{tab:methods} under dense sampling. Dense samples are two-hop subgraphs
at seed ratio 0.05 and cover about 76\% of the points. Under sparse sampling, we keep the subset shared with the
controlled benchmark. The sparse samples use ratio 0.01 with two hops and 0.005 with one hop. They cover about 25\% and
5\% of the points. SCANO, SCANO without message passing and Transolver + LFE are also trained on the whole mesh. All LFE
variants in the main tables are re-implementations trained under this unified protocol. LFE trained with its official
code and published settings reaches the accuracy reported in its paper (Section~\ref{app:olbc} of the supplementary
material).

\subsubsection{DCTS: data-centre rooms}
\label{sec:dcts-million}

Air-cooled data centres operate under strict thermal guidelines for the air drawn into each server inlet~\cite{ashrae}.
Thermal engineers need to know which servers risk overheating. They also need to know how the inlet temperatures shift
when a rack's power rises or a cooling unit is throttled. Answering each such query with computational fluid dynamics
(CFD) requires solving the steady, buoyant, turbulent airflow over about a million cells. From each predicted field, we
therefore also extract the mean temperature at every server inlet.

To represent this setting, DCTS comprises 500 rooms. They share an outer building envelope but differ in their entire
internal layout. The number, placement and heat load of the racks vary (18 to 273 servers). So do the number, placement
and supply conditions of the cooling units (two to six). The arrangement of perforated panels, partitions and pillars
varies too. Each room thus has its own geometry, its own mesh and hundreds of boundary patches
(Figures~\ref{fig:bc-stats} and~\ref{fig:datasets}c).

An in-house parametric generator produces the surface geometry of each room. It also writes a case specification that
assigns a physical role to every patch. \texttt{snappyHexMesh} discretises the geometry into a hexahedron-dominant mesh.
The OpenFOAM solver \texttt{buoyantBoussinesqSimpleFoam} then solves the steady, incompressible, buoyant
Reynolds-averaged Navier--Stokes equations. It uses the Boussinesq approximation, the realizable $k$--$\varepsilon$ closure and wall
functions. Each server exhaust returns its volumetric airflow. Its temperature is the mean inlet temperature raised by
the heat-load temperature rise (Remark~\ref{prop:jump}). The cooling-unit supplies impose fixed flow rates and
temperatures (Section~\ref{supp:r7-reference} of the supplementary material audits convergence, energy balance and mesh
sensitivity).

From the converged solution, each case records three fields on all cells and boundary faces. They are the
Reynolds-averaged velocity $\mathbf{U}=(U_x,U_y,U_z)$, the kinematic pressure minus its hydrostatic part $p'$, and the
temperature $T$. Each case also stores the mesh connectivity, face areas, normals and patch roles. The data splits were
fixed at generation time. All normalisation statistics come from the training rooms only.

Field accuracy alone does not show whether a model responds to a change of the boundary. We therefore also generate
paired interventions. No two rooms share a mesh. Each test room is therefore solved again on its own mesh after one
change: the power of one rack is doubled, or the airflow of one cooling unit is halved. The difference between the two
solutions isolates the room's physical response to that single change. Twenty such pairs in ten test rooms form our main
response evaluation. They were held out from every architectural and hyperparameter decision. Another 200 pairs on
training rooms are used in training.

\label{sec:dcts-protocol}To keep the comparison fair, all methods that ingest boundary records read the same typed face records. Every method
also receives the same heat-source features. On the boundary records, these are the heat load and temperature rise of
each server. At every domain point, they are the load of the nearest server and the offset to its exhaust. The offset is
expressed in a frame aligned with the exhaust jet. All models are trained on the two-hop subgraphs of
Table~\ref{tab:datasets} (Figure~\ref{fig:visibility}).

\subsection{Compared models}
\label{sec:setup-methods}

All compared models are matched to within ten percent of SCANO's parameter count (Section~\ref{supp:r7-capacity}
tests larger budgets). They fall into three groups
(Table~\ref{tab:methods}). The first group varies boundary and state injection on the SCANO backbone. It includes SCANO,
the SCANO backbone + BQA, global modulation by FiLM or adaLN-Zero~\cite{dit,unisolver}, and a variant with boundary data
only on sampled boundary nodes (\emph{boundary as sampled nodes}). The second group contains six external operator
backbones. Each is evaluated in its original (native) form and with BQA, AGS or both (Section~\ref{sec:hosts}).
Transolver++~\cite{transolverpp} is added as published. The third group contains dedicated boundary encoders: Transolver
+ LFE~\cite{lfe}, BENO~\cite{beno} and Slice aggregation (Section~\ref{sec:slices}). In the result tables, GNOT and
GeoTransolver are listed with this group, because their native layers already attend to boundary records. Three further
ablations isolate single design choices. They are SCANO with anchors drawn outside the message-passing layers, AB-UPT +
AGS, and the SCANO backbone + BQA without node boundary values.

\begin{table}[t]
\centering
\footnotesize
\setlength{\tabcolsep}{3pt}
\caption{The compared models and how the boundary and the global state enter them. \emph{Node values} are boundary data attached
to the sampled boundary nodes. \emph{Records} are all boundary records of a case, unsampled and weighted by area or by
count. The \emph{global state} column names the domain points over which the global tokens are pooled. Under \emph{AGS},
the anchors are drawn independently of the sample in training and cover all cells at deployment. For each backbone, the
second row gives the native route that BQA and AGS replace. Check marks show the data sets on which a model is evaluated
(Section~\ref{sec:hosts}).}
\label{tab:methods}
\begin{tabular}{P{0.21\linewidth}P{0.38\linewidth}P{0.23\linewidth}ccc}
\toprule
Method & Boundary input & Global state & Contr. & Public & DCTS \\
\midrule
\multicolumn{6}{l}{\emph{The SCANO backbone: ways of injecting the boundary}} \\
SCANO & node values; BQA in every layer, records by area & AGS, independent of the sample &
$\checkmark$ & $\checkmark$ & $\checkmark$ \\
SCANO backbone + BQA & as SCANO & the sample & $\checkmark$ & $\checkmark$ & $\checkmark$ \\
Global modulation & one area-weighted summary, FiLM or adaLN-Zero in every layer & the sample & $\checkmark$ &
$\checkmark$ & \\
Boundary as sampled nodes & node values only & the sample & & $\checkmark$ & \\
No boundary values & none (geometry only) & the sample & $\checkmark$ & & \\
\midrule
\multicolumn{6}{l}{\emph{Backbones: original, and with the SCANO components}} \\
Transolver & node values & the sample, by count & $\checkmark$ & $\checkmark$ & $\checkmark$ \\
 & + BQA & AGS & & & \\
\addlinespace[0.35em]
AB-UPT & node values & anchors from the sample & $\checkmark$ & $\checkmark$ & $\checkmark$ \\
 & + BQA & AGS: anchors over the whole mesh & & & \\
\addlinespace[0.35em]
AeroGTO & node values & the sample & $\checkmark$ & $\checkmark$ & $\checkmark$ \\
 & + BQA & AGS & & & \\
\addlinespace[0.35em]
GNOT & records cross-attended in every layer, by count & the sample, by count & $\checkmark$ & $\checkmark$ &
$\checkmark$ \\
 & BQA in place of (controlled, DCTS) or beside (public) its record attention & AGS & & & \\
\addlinespace[0.35em]
GeoTransolver & records as the context of every layer, by count & the sample, by count & $\checkmark$ & $\checkmark$ &
$\checkmark$ \\
 & + BQA & AGS & & & \\
\addlinespace[0.35em]
MeshGraphNets & node values, message passing & --- & $\checkmark$ & $\checkmark$ & \\
 & + BQA & & & & \\
\addlinespace[0.35em]
Transolver++ & node values & the sample, by count & & $\checkmark$ & \\
\midrule
\multicolumn{6}{l}{\emph{Other boundary encoders}} \\
Transolver + LFE & LFE extender at the input, records by count (and by area) & the sample, by count & $\checkmark$ &
$\checkmark$ & $\checkmark$ \\
Slice aggregation & records pooled into slice tokens, read once by the domain slices & the sample, by count &
$\checkmark$ & $\checkmark$ & $\checkmark$ \\
BENO & all boundary nodes averaged into one vector per step & --- & $\checkmark$ & $\checkmark$ & \\
\bottomrule
\end{tabular}
\end{table}

\subsection{Training and model selection}
\label{sec:setup-training}

Every model is trained on mesh samples under one shared protocol per data set (Table~\ref{tab:hyper} and
Algorithm~\ref{alg:train} of the supplementary material). As is standard for these backbones, the training loss weights
all sampled points equally. Proposition~\ref{prop:sampling}(iv) shows that this choice is separate from the
architectural conditions studied here. On DCTS, the loss also includes a difference loss over the paired interventions
of the training rooms. The training samples depend only on the case and the epoch. All comparisons are therefore paired.

For each method, a first training seed is run at the learning rates $10^{-3}$ and $3\times10^{-4}$. Further seeds use
the rate with the lower whole-mesh validation error. Every configuration is thus trained with multiple seeds. Unless a
caption states otherwise, every reported metric is the mean over these seeds, the first one included. It is given with
$\pm$ half the width of its 95\% case-bootstrap interval over the test cases (Section~\ref{supp:r7-stats} adds a paired
bootstrap over cases and seeds). No design or hyperparameter choice uses
the test sets. All evaluation criteria were registered before the test results were inspected. Some runs do not leave
the initial loss plateau under a registered validation threshold. Such a run is kept and marked in the tables. An
additional seed replaces it in the mean over the converged runs.

\subsection{Deployments and metrics}
\label{sec:setup-deploy}

\begin{figure*}[t]
\centering
\includegraphics[width=\textwidth]{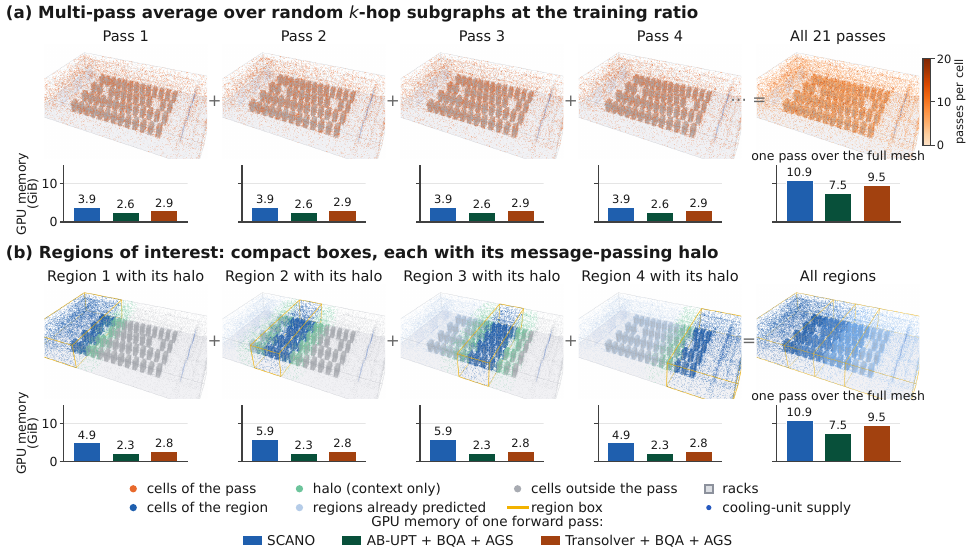}
\caption{Two ways to cover a DCTS test room with forward passes. (a) The multi-pass average. Each pass draws seed cells
at the training ratio and grows them by two hops. Orange points are the cells of the pass. Grey points are the other
cells. The last panel counts the passes that cover each cell. (b) Regions of interest that fill the room one after
another. Each region is a box, outlined in yellow. A message-passing model also reads a halo around it (green). Its
prediction is kept only inside the box. In both (a) and (b), the bars give the peak GPU memory of each pass for three backbones with BQA and
AGS, and the last column gives one pass over the full mesh. The points are a random subset of the cells.}
\label{fig:tta-roi}
\end{figure*}

Each trained model is evaluated in three deployment regimes (Figure~\ref{fig:method}b, Algorithm~\ref{alg:deploy}). These
are a subsample drawn as in training (\emph{Sample} or \emph{Subgraph} in the tables), the whole mesh in a single pass
(\emph{Whole mesh}), and compact regions of interest (\emph{Region}). On the
controlled benchmark, we add a reverse-graded mesh and a sequence of refined record sets (Table~\ref{tab:r0s-deploy}). A
region of interest consists of all cells and boundary faces inside an axis-aligned box. The box holds a prescribed
fraction of the cells. The region is the same for all methods on a data set. Message-passing models also read a
\emph{halo} of cells, as deep as their receptive field. Some models compute their global state from the sample. On DCTS, we also
evaluate each model in its native full-room deployment (\emph{own deployment} in Table~\ref{tab:dcts-main}). For sample-dependent models, this is the cell-wise average of the predictions over training-sized
subgraphs that cover the mesh. For models with AGS, it is simply the single whole-mesh pass.
Figure~\ref{fig:tta-roi} shows the average and a cover by regions on a DCTS test room. All runs
were trained, and timed after warm-up, on NVIDIA GPUs with 96 GB of memory.

Field accuracy is measured by the relative $L^2$ error over the deployed points or cells, weighted by cell volume on
volumetric meshes. On DCTS, it is averaged over the five fields $(U_x,U_y,U_z,p',T)$, each normalised by the variation of the true field about its training mean. The
deployment drift $\delta$ is the relative distance between a deployment's prediction and the model's own whole-mesh
prediction on the same nodes. On DCTS, the reference is each model's own-deployment prediction, and the drift is
normalised in the same way as the error. On another mesh, the drift counts the excess over the reference change. For each paired
intervention, the response error is
\begin{equation}
  E_{\mathrm{resp}} = \frac{\|\Delta\hat u-\Delta u\|}{\|\Delta u\|},
  \label{eq:eresp}
\end{equation}
where $\Delta u$ and $\Delta\hat u$ are the true and predicted field changes in the error norm. On DCTS, this norm is
taken for the temperature $T$ over the whole room. Drift is always reported together with the response
(Section~\ref{sec:conditions}). Identities that hold by construction are marked as such.

To measure what sampled training costs, we also train on whole meshes where possible. Full cases fit in GPU memory on
both benchmarks. On DCTS, they fit with activation checkpointing (Section~\ref{supp:impl}) for Transolver, Transolver
+ BQA and Transolver + BQA + AGS, but not for the SCANO backbone (Section~\ref{sec:bc}). Where whole-mesh training is feasible, a method's
\emph{sampling gap} is its whole-mesh error under sampled training minus that under whole-mesh training.

\section{Results: testing the conditions}
\label{sec:results}

We begin with the effect of BQA and AGS on seven operator backbones and with the ranking of all methods on the two
benchmarks (Section~\ref{sec:test-hosts}). We then test each condition of Section~\ref{sec:conditions} against its
prediction made before training (Sections~\ref{sec:test-c1}--\ref{sec:test-c4}). Next, we inspect the predicted fields
of the benchmarks (Section~\ref{sec:bench-fields}). Section~\ref{sec:dcts} then turns to DCTS.

\subsection{The components on seven backbones}
\label{sec:test-hosts}
\label{sec:hosts-results}

\begin{figure*}[t]
\centering
\includegraphics[width=\textwidth]{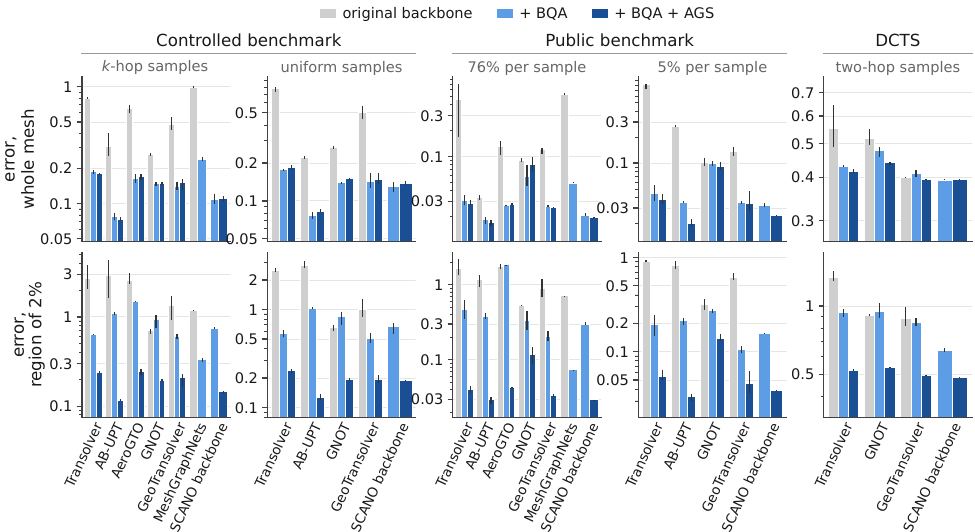}
\caption{Effect of BQA and AGS on seven operator backbones, for each data set and sampling setting (columns). The bars give the
relative $L^2$ error of the original (native) backbone, of the backbone with BQA and of the backbone with BQA + AGS. The top row is
the whole mesh in a single pass. Table~\ref{tab:dcts-main} gives each method's own deployment on DCTS. The bottom row is
a compact region that holds 2\% of the cells (log axes). Bars show the mean over training seeds. Whiskers show the
range. Table~\ref{tab:hosts} of the supplementary material lists the values.}
\label{fig:hosts-bars}
\end{figure*}

Figure~\ref{fig:hosts-bars} compares each native backbone with its variants after adding BQA and after adding both BQA and
AGS. The comparison covers all three data sets. Each configuration is evaluated on the whole mesh and in a compact
region of interest much smaller than a training sample. Table~\ref{tab:hosts} of the supplementary material lists the
values. The architecture, initialisation, learning rate and training samples of each backbone stay fixed throughout
(Section~\ref{sec:hosts}). Tables~\ref{tab:r0s-main} and~\ref{tab:olbc-main} report all methods in every deployment of
the two benchmarks.

On the two benchmarks, BQA improves every backbone it is added to. There, it cuts the whole-mesh error of most native
backbones by more than half (Figure~\ref{fig:hosts-scatter}a of the supplementary material). In most cases it also improves the
regional accuracy. On the SCANO backbone without message passing, BQA lowers the error several-fold compared with
ingesting the boundary as sampled nodes (Table~\ref{tab:olbc-main}). The gain is largest where the native backbone sees
the boundary only through sampled boundary nodes. The errors of Transolver, AeroGTO and MeshGraphNets fall by a factor
of several to more than an order of magnitude. The gain remains substantial on GeoTransolver and AB-UPT. These backbones
already attend to the boundary through a geometry context or an anchor set. On them, BQA shifts the whole distribution
of per-case errors downward (Figure~\ref{fig:hosts-violin} of the supplementary material).

On DCTS, in a single whole-room pass, BQA lowers the error of Transolver by about one fifth. It lowers that of GNOT more
modestly. GeoTransolver keeps its accuracy with BQA and AGS (Table~\ref{tab:hosts}). Its geometry context already
attends to the boundary in every layer. The relative gain is smaller on DCTS for two reasons. First, every method
already receives the heat load of the nearest server at every domain point. Second, BQA changes the velocity
$\mathbf{U}$ less than the boundary-driven temperature $T$. The velocity makes up three of the five error components. On
Transolver, BQA changes it by at most one tenth, against one fifth for $T$ (Table~\ref{tab:dcts-phys}). Without the
per-point heat-load feature, Transolver + BQA stays well ahead of the native backbone. The lead is largest in regions of
interest (Table~\ref{tab:ablations}).

BQA improves accuracy. AGS makes the backbones consistent across deployments. Added to BQA, AGS keeps the whole-mesh
accuracy of BQA alone on almost every backbone. Only GNOT on the public benchmark gives up part of it. AGS also cuts the
regional error of every backbone that maintains a global state. The reduction is several-fold on the two benchmarks. In
DCTS rooms it is one quarter to one half (Figure~\ref{fig:hosts-scatter}b). On GNOT in the controlled benchmark, BQA
alone raises the error inside a small region. This increase comes from recomputing the global state from the region. It
disappears once AGS is added. With both modules, the regional error stays close to the whole-mesh error. Backbones whose
state follows the sample degrade in regions (Figure~\ref{fig:hosts-bars}, bottom; Section~\ref{sec:test-c3}).

These two effects determine the ranking of all methods. With BQA and AGS, AB-UPT reaches the lowest error on all three
data sets. It leads every deployment in the $k$-hop family and on the reverse-graded mesh of the controlled benchmark.
Among the backbones with BQA and AGS, SCANO ranks second on both benchmarks. On DCTS, it ties with GeoTransolver within
its confidence interval. SCANO computes its whole-mesh state in a single pass of linear cost. AB-UPT needs a quadratic
anchor self-attention instead (Section~\ref{sec:dcts-cost}). SCANO's message-passing layers give a clear gain in
accuracy on both benchmarks. On the whole mesh of the public benchmark, BQA decides the ranking. Every model of
Table~\ref{tab:olbc-main} with BQA outperforms all other boundary encoders. The SCANO backbone stays ahead even when it
reads the boundary through BQA alone, without node boundary values. Among the alternative encoders, LFE comes closest on
the controlled benchmark once it uses area weights. Its published count-weighted form lags well behind. At the scale and
metric of the original LFE study, SCANO trained on samples matches the official whole-graph LFE implementation. In that
setting, as on the public benchmark and DCTS, BQA also improves Transolver more than LFE does (Section~\ref{app:olbc}
of the supplementary material).

\subsection{C1: information completeness}
\label{sec:test-c1}

\begin{figure*}[t]
\centering
\includegraphics[width=\textwidth]{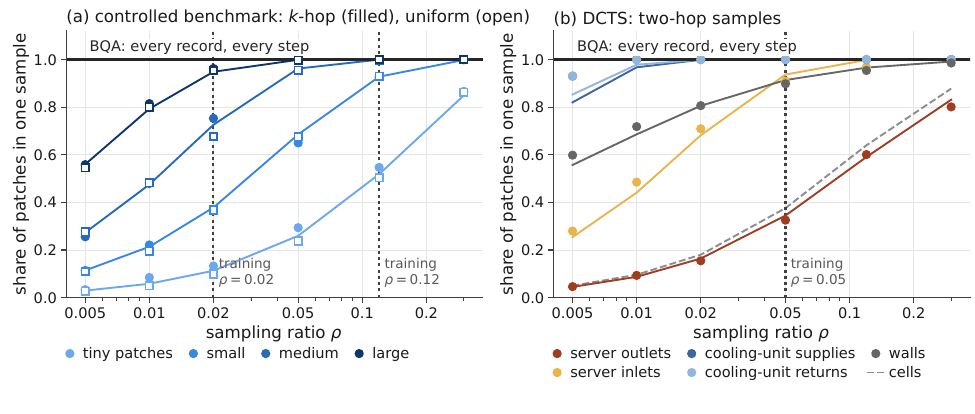}
\caption{Fraction of the boundary patches that a single training sample observes, against the sampling ratio $\rho$
(Proposition~\ref{prop:visibility}). (a) Controlled benchmark. The observed fraction is shown for each area stratum, for
$k$-hop subgraphs (filled) and uniform subsets (open). (b) DCTS. The observed fraction is shown for each component role
under two-hop training subgraphs. The dashed line is the sampled fraction of interior cells. Markers are measurements on
the sampler. Solid curves follow Proposition~\ref{prop:visibility}. Vertical dotted lines mark the training ratios. The
top line marks BQA, which reads every patch.}
\label{fig:visibility}
\end{figure*}

Proposition~\ref{prop:visibility} predicts before training what fraction of the boundary a sample observes. A patch of
$n_p$ faces is missed entirely with probability close to $(1-\rho)^{n_p}$. The smallest patches are missed most often. The share of
patches with at least one face in a sample follows this curve closely on both the controlled benchmark and DCTS
(Figure~\ref{fig:visibility}). At the training ratio, a single sample of the controlled benchmark misses about one sixth
of all patches. A sample of a DCTS room captures nearly 40\% of the interior cells. Yet it captures only one third of
the server outlets, which drive the room's thermal field.

Any model that ingests the boundary only through sampled boundary nodes inherits this omission. It shows no response at
all to interventions on patches outside its sample. These cases are marked $^{\ddagger}$ in
Tables~\ref{tab:r0s-main-resp}, \ref{tab:r0s-uniform-resp} and~\ref{tab:r0s-strong-resp} of the supplementary material.
On the public benchmark, every patch is partly present in each sample. Even there, relying on sampled boundary nodes alone reduces
accuracy. The SCANO backbone conditioned in this way has several times the error of SCANO without message passing. This
holds even with four times as many slice tokens (Table~\ref{tab:olbc-main}). BQA, in contrast, reads all records in
every step.

Information completeness also requires the boundary type of each record. A type switch changes what a patch prescribes
without changing its geometry. We trained SCANO without record types and kept everything else the same. Its response
error on the type-switch interventions of the controlled benchmark roughly tripled (Table~\ref{tab:ablations} of the
supplementary material). Proposition~\ref{prop:records}(v) of the supplementary material implies this result.

\subsection{C2: mesh consistency}
\label{sec:test-c2}

Proposition~\ref{prop:count} makes two predictions. Aggregating graded records by count leaves a covariance bias that
does not vanish under refinement. Aggregating them by area converges. A training-free test confirms both. A fixed kernel
aggregates a sequence of refined record sets on the controlled benchmark (Figure~\ref{fig:k1} of the supplementary
material). Area weighting converges at second order for all mesh gradings. Count weighting on graded records stalls at
the predicted plateau.

Trained models show the same separation on the reverse-graded mesh (\emph{Other mesh}, the last column of Table~\ref{tab:r0s-main}). SCANO
and area-weighted Transolver + LFE weight boundary records by area. Both keep their accuracy. SCANO nearly matches its
error on the training mesh. Count-weighted Transolver + LFE and GNOT degrade sharply. Weighting AB-UPT's anchors by cell
volume removes most of its remeshing loss, as Proposition~\ref{prop:drift} predicts. Within SCANO itself, we also
trained with records weighted by count instead of area. The error on the reverse-graded mesh roughly tripled in both
sampling families. On the training mesh, it rose by about a quarter on $k$-hop samples and changed little on uniform samples
(Table~\ref{tab:ablations}).

\begin{figure}[t]
\centering
\includegraphics[width=\textwidth]{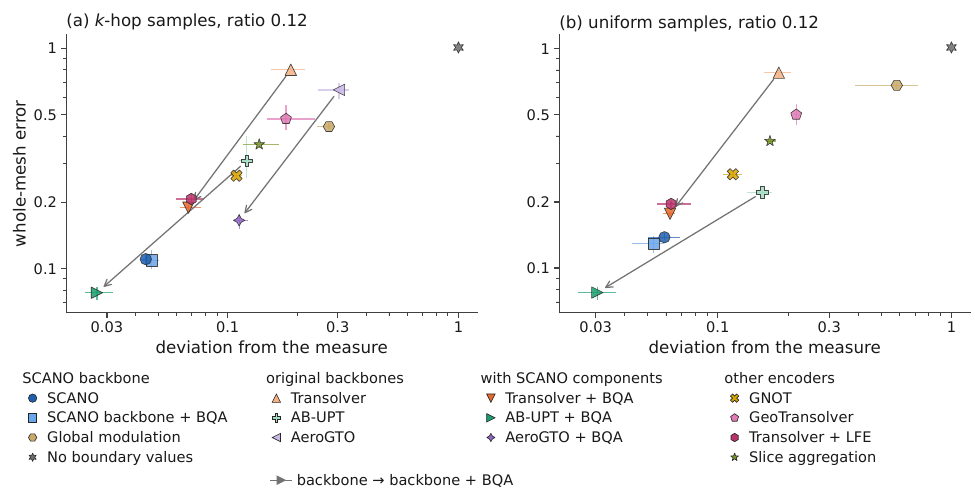}
\caption{Unit-mass test of the learned measure against the whole-mesh relative $L^2$ error on the controlled benchmark, for (a)
$k$-hop and (b) uniform samples at ratio 0.12. Raising all Dirichlet and external Robin values by a constant tests unit
mass. The horizontal axis (deviation from the measure) is $0$ for exact conservation and $1$ for no response. Markers
and bars show the mean and range over training seeds. Arrows join each original backbone to the same backbone with BQA.
Groups follow Table~\ref{tab:r0s-main}.}
\label{fig:measure}
\end{figure}

The boundary measure behind these weights can also be probed directly in a trained model. The controlled benchmark is
linear. The representation~\eqref{eq:measure} can therefore be tested on any trained surrogate without a numerical
solver. Three tests follow from it. Shifting all Dirichlet and external Robin values by a constant should shift every
interior cell by the same constant. This confirms that the weights sum to one (\emph{unit-mass test}). Perturbing a single such patch should
shift each cell by a fraction in $[0,1]$, and these fractions should sum to one over all patches. This confirms
non-negativity. Doubling a Neumann flux should double the increment it causes. This confirms additivity.
Figure~\ref{fig:measure} evaluates the unit-mass test over the test split. Models with BQA satisfy this unit-mass property
closely. Transolver, AB-UPT and AeroGTO each move several-fold closer to the exact measure once BQA is attached. Across
all trained models, the deviation from unit mass ranks the methods almost exactly as their whole-mesh error does. The
test thus probes the learned quadrature without any reference solution.

\subsection{C3: sample consistency}
\label{sec:test-c3}

\begin{figure*}[t]
\centering
\includegraphics[width=\textwidth]{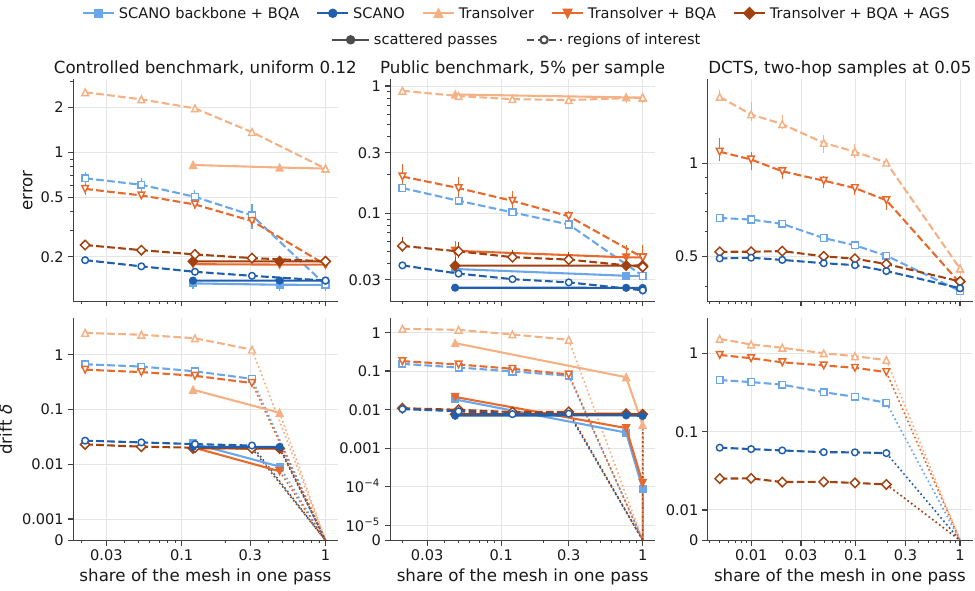}
\caption{Accuracy and consistency of regional deployments against the share of the mesh in one pass
(Proposition~\ref{prop:drift}, condition C3). The top row gives the relative $L^2$ error. The bottom row gives the drift
$\delta$ from each model's own whole-mesh prediction, which is zero at full coverage (symmetric-log vertical axis).
Solid curves are scattered passes. Dashed curves are compact regions of interest covering 2--30\% of the cells on the
benchmarks and 0.5--20\% on DCTS. Training samples are uniform at ratio 0.12 on the controlled benchmark. On the public
benchmark, they are one-hop subgraphs that cover about 5\% of the mesh. On DCTS, they are two-hop subgraphs. Markers and
bars show the mean and range over training seeds.}
\label{fig:coverage-deploy}
\end{figure*}

Proposition~\ref{prop:drift} predicts how a global state computed from the deployed sample behaves. As the sample shrinks,
the state shifts linearly with the change of the inclusion measure. It therefore drifts much faster on a compact region
than on scattered samples that cover the mesh almost uniformly. Figure~\ref{fig:coverage-deploy} confirms this on all
three data sets as a function of deployment coverage. For every model whose global state is computed from the deployed sample,
its drift from its own whole-mesh prediction grows as the coverage decreases. The growth is far steeper on compact
regions than on scattered samples. The prediction error follows the drift. SCANO's prediction in any deployment stays aligned with its
whole-mesh prediction up to the Monte Carlo variance of its anchors. Its regional error therefore equals its whole-mesh
error on the same cells. BQA lowers the error and the drift of Transolver at every coverage. Only AGS removes the
dependence on the region.

\begin{table*}[!tbp]
\centering
\scriptsize
\setlength{\tabcolsep}{3pt}
\caption{Controlled benchmark (512 test cases): volume-weighted relative $L^2$ error (seed mean $\pm$ half-width of the 95\% case-bootstrap interval) across four deployments (Section~\ref{sec:setup-deploy}): one training-sized sample, the whole mesh in one pass, compact regions of interest (30\%, 12\%, 5\% and 2\% of cells), and zero-shot transfer to the reverse-graded $40^3$ mesh (\emph{other mesh}). Bold indicates the best result per column; the uniform-sampling family is in Table~\ref{tab:r0s-uniform} of the supplementary material. $^{\S}$No boundary input.}
\label{tab:r0s-main}
\begin{tabular}{P{4.3cm}ccccccc}
\toprule
 & & & \multicolumn{4}{c}{Region, share of the cells} & \\
\cmidrule(lr){4-7}
Method & Sample & Whole mesh & 0.3 & 0.12 & 0.05 & 0.02 & Other mesh \\
\midrule
\multicolumn{8}{l}{\textbf{$k$-hop samples, ratio 0.12}} \\
\multicolumn{8}{l}{\emph{SCANO}} \\
SCANO & 0.111\,{\tiny$\pm$\,0.009} & 0.111\,{\tiny$\pm$\,0.009} & 0.118\,{\tiny$\pm$\,0.010} & 0.127\,{\tiny$\pm$\,0.013} & 0.136\,{\tiny$\pm$\,0.018} & 0.145\,{\tiny$\pm$\,0.015} & 0.115\,{\tiny$\pm$\,0.008} \\
SCANO backbone + BQA & 0.112\,{\tiny$\pm$\,0.007} & 0.109\,{\tiny$\pm$\,0.008} & 0.401\,{\tiny$\pm$\,0.046} & 0.555\,{\tiny$\pm$\,0.064} & 0.697\,{\tiny$\pm$\,0.080} & 0.750\,{\tiny$\pm$\,0.083} & 0.115\,{\tiny$\pm$\,0.008} \\
\midrule
\multicolumn{8}{l}{\emph{Boundary injection on the SCANO backbone}} \\
SCANO, anchors outside the message passing & 0.118\,{\tiny$\pm$\,0.009} & 0.118\,{\tiny$\pm$\,0.009} & 0.129\,{\tiny$\pm$\,0.011} & 0.138\,{\tiny$\pm$\,0.013} & 0.153\,{\tiny$\pm$\,0.022} & 0.167\,{\tiny$\pm$\,0.029} & 0.119\,{\tiny$\pm$\,0.009} \\
SCANO without message passing & 0.139\,{\tiny$\pm$\,0.010} & 0.139\,{\tiny$\pm$\,0.010} & 0.149\,{\tiny$\pm$\,0.011} & 0.161\,{\tiny$\pm$\,0.015} & 0.176\,{\tiny$\pm$\,0.023} & 0.193\,{\tiny$\pm$\,0.027} & 0.139\,{\tiny$\pm$\,0.010} \\
Global modulation & 0.525\,{\tiny$\pm$\,0.034} & 0.442\,{\tiny$\pm$\,0.030} & 0.527\,{\tiny$\pm$\,0.038} & 0.615\,{\tiny$\pm$\,0.049} & 0.669\,{\tiny$\pm$\,0.057} & 0.762\,{\tiny$\pm$\,0.062} & 0.723\,{\tiny$\pm$\,0.047} \\
No boundary values$^{\S}$ & 1.009\,{\tiny$\pm$\,0.033} & 1.009\,{\tiny$\pm$\,0.033} & 1.044\,{\tiny$\pm$\,0.041} & 1.091\,{\tiny$\pm$\,0.052} & 1.134\,{\tiny$\pm$\,0.065} & 1.199\,{\tiny$\pm$\,0.068} & 1.009\,{\tiny$\pm$\,0.033} \\
\midrule
\multicolumn{8}{l}{\emph{Other boundary encoders}} \\
Transolver + LFE & 0.206\,{\tiny$\pm$\,0.026} & 0.207\,{\tiny$\pm$\,0.026} & 0.334\,{\tiny$\pm$\,0.032} & 0.445\,{\tiny$\pm$\,0.044} & 0.514\,{\tiny$\pm$\,0.054} & 0.598\,{\tiny$\pm$\,0.069} & 0.467\,{\tiny$\pm$\,0.045} \\
Transolver + LFE, area & 0.166\,{\tiny$\pm$\,0.014} & 0.167\,{\tiny$\pm$\,0.014} & 0.294\,{\tiny$\pm$\,0.026} & 0.388\,{\tiny$\pm$\,0.039} & 0.452\,{\tiny$\pm$\,0.053} & 0.503\,{\tiny$\pm$\,0.054} & 0.176\,{\tiny$\pm$\,0.016} \\
GNOT & 0.263\,{\tiny$\pm$\,0.015} & 0.264\,{\tiny$\pm$\,0.015} & 0.423\,{\tiny$\pm$\,0.028} & 0.529\,{\tiny$\pm$\,0.042} & 0.609\,{\tiny$\pm$\,0.059} & 0.700\,{\tiny$\pm$\,0.072} & 0.505\,{\tiny$\pm$\,0.036} \\
GeoTransolver & 0.478\,{\tiny$\pm$\,0.029} & 0.479\,{\tiny$\pm$\,0.029} & 0.916\,{\tiny$\pm$\,0.057} & 1.102\,{\tiny$\pm$\,0.085} & 1.226\,{\tiny$\pm$\,0.132} & 1.352\,{\tiny$\pm$\,0.135} & 0.542\,{\tiny$\pm$\,0.027} \\
Slice aggregation & 0.364\,{\tiny$\pm$\,0.025} & 0.366\,{\tiny$\pm$\,0.024} & 0.789\,{\tiny$\pm$\,0.053} & 1.071\,{\tiny$\pm$\,0.084} & 1.136\,{\tiny$\pm$\,0.097} & 1.262\,{\tiny$\pm$\,0.113} & 0.487\,{\tiny$\pm$\,0.025} \\
BENO & 0.457\,{\tiny$\pm$\,0.038} & 0.458\,{\tiny$\pm$\,0.038} & 0.480\,{\tiny$\pm$\,0.044} & 0.514\,{\tiny$\pm$\,0.055} & 0.551\,{\tiny$\pm$\,0.072} & 0.585\,{\tiny$\pm$\,0.074} & 0.768\,{\tiny$\pm$\,0.033} \\
\midrule
\multicolumn{8}{l}{\emph{Backbones}} \\
Transolver & 0.846\,{\tiny$\pm$\,0.055} & 0.800\,{\tiny$\pm$\,0.045} & 1.354\,{\tiny$\pm$\,0.107} & 1.893\,{\tiny$\pm$\,0.169} & 2.329\,{\tiny$\pm$\,0.231} & 2.718\,{\tiny$\pm$\,0.254} & 0.743\,{\tiny$\pm$\,0.033} \\
Transolver + BQA & 0.186\,{\tiny$\pm$\,0.012} & 0.189\,{\tiny$\pm$\,0.012} & 0.429\,{\tiny$\pm$\,0.026} & 0.538\,{\tiny$\pm$\,0.044} & 0.597\,{\tiny$\pm$\,0.054} & 0.636\,{\tiny$\pm$\,0.069} & 0.216\,{\tiny$\pm$\,0.015} \\
\addlinespace[0.35em]
AB-UPT & 0.436\,{\tiny$\pm$\,0.044} & 0.309\,{\tiny$\pm$\,0.020} & 1.247\,{\tiny$\pm$\,0.089} & 1.970\,{\tiny$\pm$\,0.155} & 2.474\,{\tiny$\pm$\,0.213} & 2.879\,{\tiny$\pm$\,0.230} & 0.472\,{\tiny$\pm$\,0.028} \\
AB-UPT + BQA & \textbf{0.074}\,{\tiny$\pm$\,0.008} & 0.078\,{\tiny$\pm$\,0.008} & 0.507\,{\tiny$\pm$\,0.046} & 0.762\,{\tiny$\pm$\,0.076} & 0.932\,{\tiny$\pm$\,0.089} & 1.099\,{\tiny$\pm$\,0.099} & 0.228\,{\tiny$\pm$\,0.019} \\
AB-UPT, whole-mesh anchors by count & 0.256\,{\tiny$\pm$\,0.029} & 0.256\,{\tiny$\pm$\,0.029} & 0.458\,{\tiny$\pm$\,0.046} & 0.484\,{\tiny$\pm$\,0.061} & 0.529\,{\tiny$\pm$\,0.080} & 0.588\,{\tiny$\pm$\,0.098} & 0.439\,{\tiny$\pm$\,0.035} \\
AB-UPT + BQA, whole-mesh anchors by count & 0.083\,{\tiny$\pm$\,0.012} & 0.083\,{\tiny$\pm$\,0.012} & 0.097\,{\tiny$\pm$\,0.015} & 0.104\,{\tiny$\pm$\,0.017} & 0.119\,{\tiny$\pm$\,0.031} & 0.134\,{\tiny$\pm$\,0.043} & 0.203\,{\tiny$\pm$\,0.015} \\
AB-UPT + BQA + AGS & \textbf{0.074}\,{\tiny$\pm$\,0.009} & \textbf{0.073}\,{\tiny$\pm$\,0.009} & \textbf{0.086}\,{\tiny$\pm$\,0.011} & \textbf{0.091}\,{\tiny$\pm$\,0.013} & \textbf{0.106}\,{\tiny$\pm$\,0.024} & \textbf{0.116}\,{\tiny$\pm$\,0.034} & \textbf{0.094}\,{\tiny$\pm$\,0.008} \\
\addlinespace[0.35em]
AeroGTO & 0.702\,{\tiny$\pm$\,0.056} & 0.648\,{\tiny$\pm$\,0.039} & 1.391\,{\tiny$\pm$\,0.113} & 2.027\,{\tiny$\pm$\,0.182} & 2.303\,{\tiny$\pm$\,0.187} & 2.589\,{\tiny$\pm$\,0.221} & 0.685\,{\tiny$\pm$\,0.049} \\
AeroGTO + BQA & 0.164\,{\tiny$\pm$\,0.018} & 0.166\,{\tiny$\pm$\,0.018} & 0.742\,{\tiny$\pm$\,0.039} & 1.014\,{\tiny$\pm$\,0.053} & 1.221\,{\tiny$\pm$\,0.078} & 1.503\,{\tiny$\pm$\,0.113} & 0.181\,{\tiny$\pm$\,0.017} \\
\bottomrule
\end{tabular}
\end{table*}

\begin{table*}[t]
\centering
\small
\caption{Public benchmark (256 test cases, training samples covering ${\sim}76\%$ of the mesh): relative $L^2$ error (\%) and regional drift $\delta$ (\%; distance between regional and whole-mesh predictions relative to the reference) across whole-mesh, training-subgraph, and compact 5\%-region deployments (Section~\ref{sec:setup-deploy}; seed mean $\pm$ 95\% case-bootstrap half-width). Because this mesh is small, SCANO uses all mesh nodes as its anchor set. Bold indicates the column best. $^{\ddagger}$Zero by construction (whole-mesh anchors); not ranked.}
\label{tab:olbc-main}
\begin{tabular}{lcccc}
\toprule
Method & Whole mesh & Subgraph & Region & Region drift $\delta$ \\
\midrule
\multicolumn{5}{l}{\textbf{Dense sampling: one training sample covers about 76\% of the mesh}} \\
\multicolumn{5}{l}{\emph{SCANO}} \\
SCANO & 1.97\,{\scriptsize$\pm$\,0.09} & 2.20\,{\scriptsize$\pm$\,0.10} & 2.59\,{\scriptsize$\pm$\,0.20} & 0$^{\ddagger}$ \\
SCANO, anchors outside the message passing & 2.22\,{\scriptsize$\pm$\,0.11} & 2.49\,{\scriptsize$\pm$\,0.12} & 2.91\,{\scriptsize$\pm$\,0.25} & 0$^{\ddagger}$ \\
SCANO backbone + BQA & 2.08\,{\scriptsize$\pm$\,0.10} & 2.50\,{\scriptsize$\pm$\,0.11} & 22.72\,{\scriptsize$\pm$\,1.71} & 22.60\,{\scriptsize$\pm$\,1.67} \\
\midrule
\multicolumn{5}{l}{\emph{Boundary injection on the SCANO backbone without message passing}} \\
SCANO without message passing & 2.76\,{\scriptsize$\pm$\,0.12} & 3.05\,{\scriptsize$\pm$\,0.12} & 3.70\,{\scriptsize$\pm$\,0.26} & 1.68\,{\scriptsize$\pm$\,0.14} \\
SCANO backbone + BQA, no node values & 3.14\,{\scriptsize$\pm$\,0.16} & 3.30\,{\scriptsize$\pm$\,0.16} & 31.11\,{\scriptsize$\pm$\,2.44} & 30.85\,{\scriptsize$\pm$\,2.44} \\
Global modulation & 59.69\,{\scriptsize$\pm$\,2.23} & 59.78\,{\scriptsize$\pm$\,2.24} & 79.24\,{\scriptsize$\pm$\,4.87} & 58.67\,{\scriptsize$\pm$\,3.16} \\
Boundary as sampled nodes & 15.42\,{\scriptsize$\pm$\,0.59} & 19.92\,{\scriptsize$\pm$\,0.78} & 182.13\,{\scriptsize$\pm$\,20.89} & 184.58\,{\scriptsize$\pm$\,19.95} \\
Boundary as sampled nodes, four times the slices & 10.38\,{\scriptsize$\pm$\,0.40} & 14.26\,{\scriptsize$\pm$\,0.60} & 168.08\,{\scriptsize$\pm$\,16.57} & 169.54\,{\scriptsize$\pm$\,16.48} \\
\midrule
\multicolumn{5}{l}{\emph{Other boundary encoders and message passing}} \\
Transolver + LFE & 5.09\,{\scriptsize$\pm$\,0.25} & 5.08\,{\scriptsize$\pm$\,0.25} & 18.19\,{\scriptsize$\pm$\,1.16} & 15.25\,{\scriptsize$\pm$\,1.08} \\
GNOT & 9.04\,{\scriptsize$\pm$\,0.39} & 9.21\,{\scriptsize$\pm$\,0.39} & 40.90\,{\scriptsize$\pm$\,2.57} & 39.86\,{\scriptsize$\pm$\,2.65} \\
GeoTransolver & 11.98\,{\scriptsize$\pm$\,0.52} & 12.07\,{\scriptsize$\pm$\,0.52} & 77.20\,{\scriptsize$\pm$\,5.13} & 77.31\,{\scriptsize$\pm$\,5.98} \\
Slice aggregation & 11.37\,{\scriptsize$\pm$\,0.52} & 11.54\,{\scriptsize$\pm$\,0.52} & 104.19\,{\scriptsize$\pm$\,6.78} & 103.53\,{\scriptsize$\pm$\,6.70} \\
MeshGraphNets & 53.54\,{\scriptsize$\pm$\,1.93} & 56.52\,{\scriptsize$\pm$\,1.88} & 66.00\,{\scriptsize$\pm$\,4.58} & 24.45\,{\scriptsize$\pm$\,1.79} \\
BENO & 9.71\,{\scriptsize$\pm$\,0.39} & 10.24\,{\scriptsize$\pm$\,0.40} & 13.56\,{\scriptsize$\pm$\,0.93} & 1.98\,{\scriptsize$\pm$\,0.24} \\
\midrule
\multicolumn{5}{l}{\emph{Backbones}} \\
Transolver & 46.85\,{\scriptsize$\pm$\,1.29} & 48.99\,{\scriptsize$\pm$\,1.37} & 163.37\,{\scriptsize$\pm$\,14.43} & 191.20\,{\scriptsize$\pm$\,14.74} \\
Transolver + BQA & 3.06\,{\scriptsize$\pm$\,0.13} & 3.22\,{\scriptsize$\pm$\,0.13} & 37.92\,{\scriptsize$\pm$\,2.88} & 37.72\,{\scriptsize$\pm$\,2.83} \\
\addlinespace[0.35em]
Transolver++ & 55.54\,{\scriptsize$\pm$\,1.64} & 57.51\,{\scriptsize$\pm$\,1.72} & 153.49\,{\scriptsize$\pm$\,11.65} & 185.94\,{\scriptsize$\pm$\,11.62} \\
\addlinespace[0.35em]
AB-UPT & 3.35\,{\scriptsize$\pm$\,0.16} & 6.84\,{\scriptsize$\pm$\,0.40} & 106.82\,{\scriptsize$\pm$\,9.82} & 107.52\,{\scriptsize$\pm$\,9.96} \\
AB-UPT + BQA & 1.84\,{\scriptsize$\pm$\,0.09} & 2.07\,{\scriptsize$\pm$\,0.10} & 33.95\,{\scriptsize$\pm$\,3.69} & 33.78\,{\scriptsize$\pm$\,3.49} \\
AB-UPT + AGS & 2.79\,{\scriptsize$\pm$\,0.13} & 6.23\,{\scriptsize$\pm$\,0.39} & 7.14\,{\scriptsize$\pm$\,0.58} & 6.23\,{\scriptsize$\pm$\,0.56} \\
AB-UPT + BQA + AGS & \textbf{1.71}\,{\scriptsize$\pm$\,0.08} & \textbf{1.92}\,{\scriptsize$\pm$\,0.08} & \textbf{2.58}\,{\scriptsize$\pm$\,0.19} & \textbf{1.16}\,{\scriptsize$\pm$\,0.08} \\
\addlinespace[0.35em]
AeroGTO & 12.97\,{\scriptsize$\pm$\,0.47} & 18.02\,{\scriptsize$\pm$\,0.72} & 172.59\,{\scriptsize$\pm$\,18.55} & 174.40\,{\scriptsize$\pm$\,17.86} \\
AeroGTO + BQA & 2.71\,{\scriptsize$\pm$\,0.12} & 3.00\,{\scriptsize$\pm$\,0.13} & 141.65\,{\scriptsize$\pm$\,11.03} & 141.51\,{\scriptsize$\pm$\,10.99} \\
\bottomrule
\end{tabular}
\end{table*}

The regional columns of Tables~\ref{tab:r0s-main} and~\ref{tab:olbc-main} show the same pattern for all methods. On the
controlled benchmark, SCANO's error in compact regions down to 2\% of the cells stays close to its whole-mesh error. The
same holds on the reverse-graded mesh. Every method whose global state is recomputed from the queried points
suffers a several-fold increase in error in the smallest region. This holds even with BQA. The SCANO backbone + BQA differs from SCANO only in how
it estimates its global state. Yet it has several times SCANO's error there. The state computed from the sample is
therefore the cause.

On the public benchmark, the mesh is small enough for SCANO's anchor set to cover all of it. SCANO's regional drift
therefore vanishes by construction. AB-UPT + BQA + AGS with freshly drawn anchors matches its whole-mesh prediction, up
to the Monte Carlo error of that draw. AB-UPT with anchors drawn from the sample loses an order of magnitude in regional
accuracy. So does the SCANO backbone + BQA. We also trained on sparser samples that cover about 25\% and 5\% of the mesh
(Table~\ref{tab:olbc-strong} of the supplementary material). The models with AGS lose far less accuracy than the variants whose
state follows the sample. SCANO's lead over the SCANO backbone + BQA therefore widens.

\begin{figure*}[t]
\centering
\includegraphics[width=\textwidth]{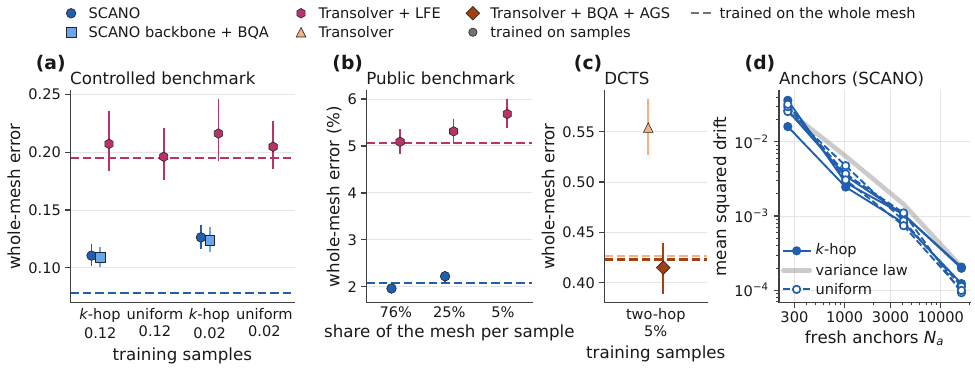}
\caption{Sampling gap against whole-mesh training, and convergence in the number of anchors. (a--c) Whole-mesh relative $L^2$
error of models trained on subsamples (markers with 95\% bootstrap bars). Dashed horizontal lines show the same
architectures trained on the full mesh. Panel (a) is the controlled benchmark. Panel (b) is the public benchmark at
several training coverages (see Table~\ref{tab:olbc-tax} for SCANO at 5\%). Panel (c) is DCTS, for two of the models whose training fits a full room in memory. (d) Mean
squared drift of SCANO's whole-mesh prediction on the controlled benchmark against the fresh anchor count $N_a$. The grey reference
line (\emph{variance law}) is the finite-population rate $1/N_a - 1/|\Omega_h|$.}
\label{fig:gap}
\end{figure*}

How closely AGS follows the whole-mesh state depends on the number of anchors. An independently anchored state departs
from the whole-mesh state only by the Monte Carlo error of sampling (Proposition~\ref{prop:sampling}(iii)). We drew
$N_a$ fresh single-cell anchors. The mean squared drift of SCANO's prediction decays at the finite-population rate of
sampling without replacement. This holds in both sampling families and for every training seed (Figure~\ref{fig:gap}d).
As $N_a$ grows, the drift converges to the error of the whole-mesh state (Table~\ref{tab:ablations} of the supplementary
material). $N_a$ is thus a budget with a known rate of convergence.

Anchoring is not the only way to remove the dependence on the sample at deployment. A model whose state follows the
sample in training can instead cache its global state over the whole mesh at deployment. Transolver-3~\cite{transolver3}
does this. The cost is a full-mesh pass for every new case. With this cache, the regional prediction coincides with the
whole-mesh prediction. Its error is the whole-mesh column of Tables~\ref{tab:r0s-main} and~\ref{tab:olbc-main}. On the
controlled benchmark, such a cache performs comparably to AGS. On the public benchmark, SCANO's margin over the cached
SCANO backbone + BQA widens as the training samples shrink (Table~\ref{tab:olbc-strong} of the supplementary material).
The reason is that the cache evaluates the network on whole-mesh states. The network never encountered such states during training
(Section~\ref{sec:global-state}).

Consistency at deployment leaves one question open. Does training on samples sacrifice accuracy relative to whole-mesh training?
Proposition~\ref{prop:sampling} predicts that it need not. A model with a sample-independent state, trained on
subsamples, converges toward the same architecture trained on the whole mesh (Figure~\ref{fig:gap}a--c). On DCTS
(Figure~\ref{fig:gap}c), Transolver fits on a whole room. Transolver + BQA + AGS trained on samples is slightly more accurate than when trained
on whole rooms. It also needs under one third of the training time per epoch. The native Transolver, in contrast, loses
markedly in accuracy. With BQA and AGS, sampled training on DCTS thus entails no loss of accuracy. On the public benchmark
(Figure~\ref{fig:gap}b), SCANO's sampling gap stays small at every coverage. That of Transolver + LFE grows as the samples shrink. On the
controlled benchmark (Figure~\ref{fig:gap}a), SCANO and the SCANO backbone + BQA coincide under whole-mesh training. They share the same
sampling gap. The gap therefore does not come from the global state (Section~\ref{supp:r7-halo}). Both models remain
well ahead of Transolver + LFE (Tables~\ref{tab:r0s-tax}, \ref{tab:olbc-tax}, \ref{tab:dcts-tax}
and~\ref{tab:ablations} of the supplementary material).

\subsection{C4: response fidelity}
\label{sec:test-c4}

On the controlled benchmark, the exact singular-value spectrum of the response operator gives the error floor of
Proposition~\ref{prop:code}. This floor depends on the code width $d$. It drops by more than two orders of magnitude
between $d=16$ and the full dimension of the boundary data (Figure~\ref{fig:predictions} of the supplementary material).
Trained global modulation stays far above this floor. Even at full code width, the rank floor is negligible. Yet the
whole-mesh error of global modulation is several times SCANO's on both benchmarks. Its response error also remains high
on the controlled benchmark (Tables~\ref{tab:r0s-main}, \ref{tab:r0s-main-resp} and~\ref{tab:olbc-main}). Its limitation
is therefore not the width of the code. It is the broadcast of one uniform summary to all points.

A model that ignores the boundary would drift as little as a correct one. Drift is therefore only meaningful next to the
response. Figure~\ref{fig:drift-response} and Table~\ref{tab:r0s-main-resp} of the supplementary material evaluate the
two together. They plot the regional drift against the response error to single-patch interventions. SCANO and AB-UPT +
BQA + AGS both combine low drift with a low response error. This also holds on the smallest patches and on patches
outside the training sample. The SCANO backbone + BQA matches SCANO's response but drifts by more than an order of
magnitude in a region. The native backbones both drift and respond poorly. Global modulation and BENO read the boundary
through a global code. They respond poorly whatever their drift.

\begin{figure}[t]
\centering
\includegraphics[width=\textwidth]{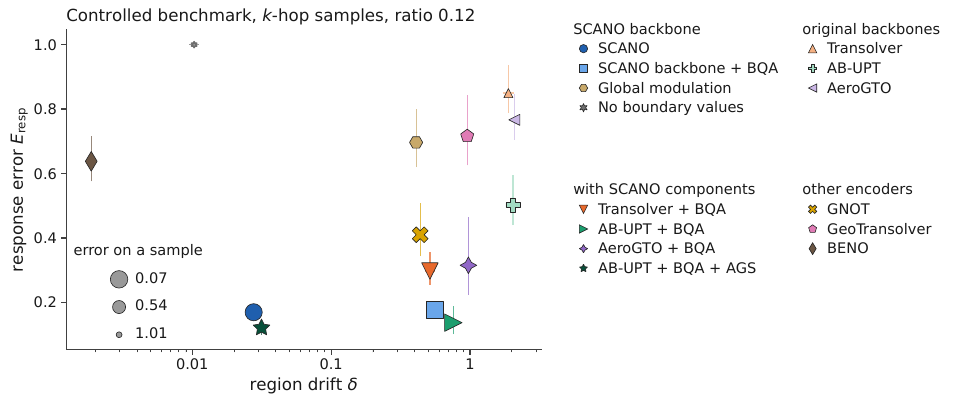}
\caption{Region drift $\delta$ against the response error $E_{\mathrm{resp}}$ on the controlled benchmark. The drift is
measured in a compact region of 12\% of the cells. The response is to single-patch interventions. Samples are $k$-hop at
ratio 0.12, and Table~\ref{tab:r0s-main-resp} of the supplementary material lists the values. Marker area grows as the
error on a sample decreases. Error bars are 95\% case-bootstrap intervals. A predictor that ignores the boundary
lies at $E_{\mathrm{resp}}=1$ by construction.}
\label{fig:drift-response}
\end{figure}

\subsection{Fields on the two benchmarks}
\label{sec:bench-fields}

\begin{figure*}[t]
\centering
\includegraphics[width=\textwidth]{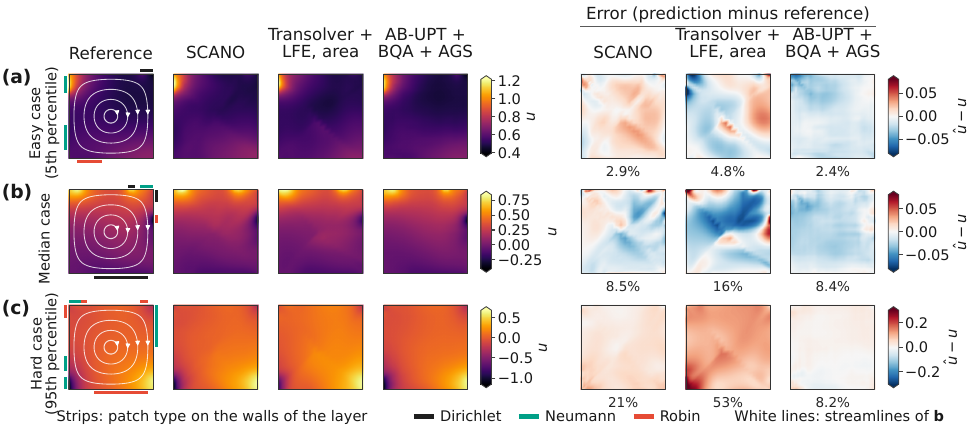}
\caption{Scalar field $u$ at mid-height and its pointwise error on the controlled benchmark. The rows are (a) the 5th-, (b) the
50th- and (c) the 95th-percentile test cases of SCANO's whole-mesh error. The first column is the finite-volume reference. It shows
streamlines of $\mathbf{b}$ and the wall patches as strips coloured by type. The next columns are the whole-mesh
predictions of SCANO, area-weighted Transolver + LFE and AB-UPT + BQA + AGS (median validation seed). The last columns
show the signed error $\hat u - u$, with the whole-mesh relative $L^2$ error below each panel.}
\label{fig:r0s-fields}
\end{figure*}

\begin{figure*}[t]
\centering
\includegraphics[width=\textwidth]{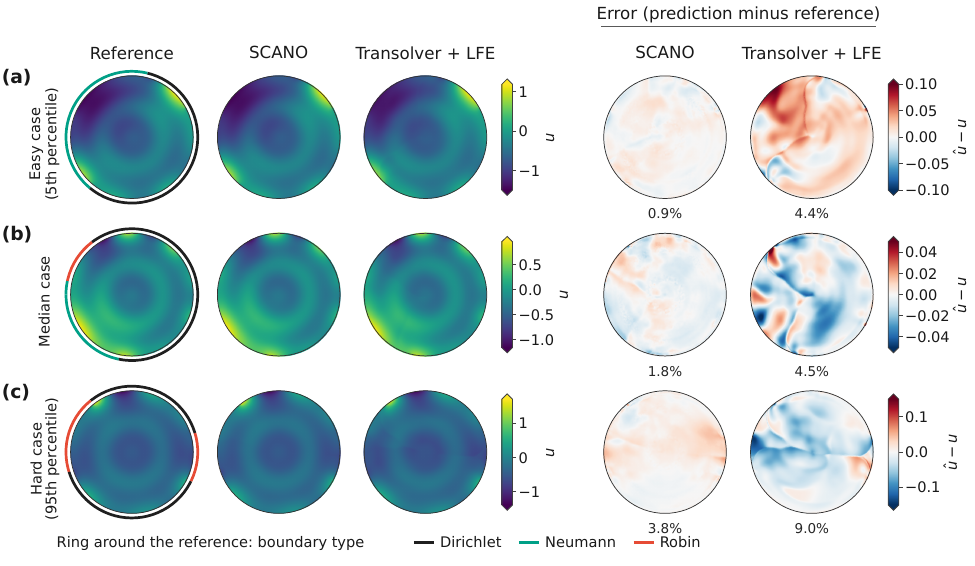}
\caption{Whole-mesh predictions and pointwise errors on the public benchmark. The rows are (a) the 5th-, (b) the 50th- and
(c) the 95th-percentile test cases of SCANO's error. Training covers about 76\% of the mesh. The first column is the finite-element
reference, with the outer ring coloured by boundary type. The next columns are the predictions of SCANO and of
Transolver + LFE, each with the seed of median validation error. The last columns show $\hat u - u$ (prediction minus reference), with the
whole-mesh relative $L^2$ error below each panel.}
\label{fig:olbc-fields}
\end{figure*}

Figure~\ref{fig:r0s-fields} shows the mid-height plane of the cube of the controlled benchmark. SCANO and AB-UPT + BQA +
AGS capture the warm and cold plumes. These plumes issue from each wall opening and are carried around by the
recirculating flow. The errors of the two models stay small and without structure. Those of area-weighted Transolver + LFE are
larger and concentrate along the layer diagonals. In the 95th-percentile case, one strongly driven patch sets the
background level of the whole plane. The two models with BQA and AGS also have the smallest offset in that level. On the
public benchmark (Figure~\ref{fig:olbc-fields}), SCANO's error likewise stays small and without structure at all three
percentiles. The error of Transolver + LFE concentrates along the boundary. It grows with the difficulty of the case.

\section{An industrial case: the cooling of data-centre rooms}
\label{sec:dcts}

This section evaluates the models on DCTS (Section~\ref{sec:dcts-million}). Its order follows the questions of a thermal
engineer. How accurate is each model over a room and in a region? What do its fields look like? How does it respond to a
single changed boundary condition? How well does it rank the server inlets? What does it cost to train and to deploy?

\subsection{Accuracy and deployment}
\label{sec:dcts-accuracy}
\label{sec:dcts-consistency}

\begin{figure*}[t]
\centering
\includegraphics[width=\textwidth]{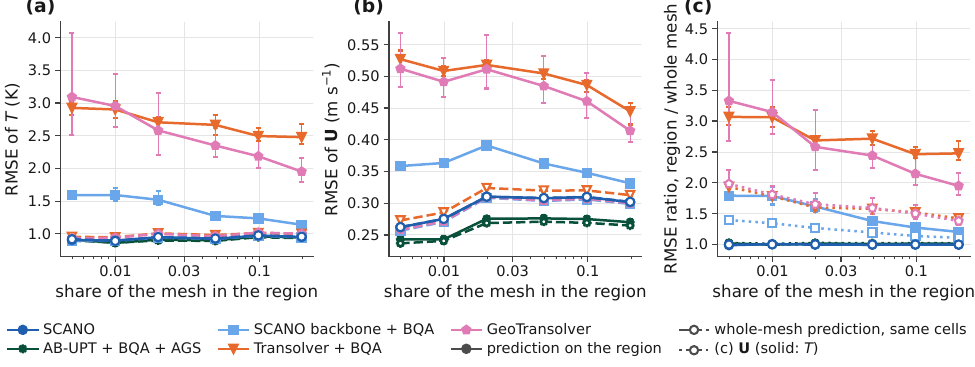}
\caption{Regional deployment against the whole-mesh prediction on the same cells, over the 50 DCTS test rooms. The horizontal
axis is the share of the mesh in the region. (a) Temperature RMSE (K). (b) Velocity RMSE ($\mathrm{m\,s^{-1}}$).
Solid lines with filled markers are standalone predictions on the region. Dashed lines with open markers are whole-mesh predictions
evaluated on the same cells. (c) Ratio of the regional to the whole-mesh RMSE, for $T$ (solid) and $\mathbf{U}$
(dotted). Markers and bars show the mean and range over training seeds.}
\label{fig:region}
\end{figure*}

On the whole mesh (Table~\ref{tab:dcts-main}), AB-UPT + BQA + AGS and AB-UPT + AGS reach the lowest overall
error. SCANO matches GeoTransolver and the SCANO backbone + BQA. It outperforms the remaining methods. SCANO's anchor
sample can pass through the message-passing layers or be drawn outside them. Passing it through those layers slightly lowers the error in every deployment. In the boundary-driven temperature $T$ over the whole room, all three SCANO-backbone
variants achieve lower error than every other method (Table~\ref{tab:dcts-phys} of the supplementary
material).
As on the benchmarks (Section~\ref{sec:test-c3}), the deployment regime decides the ranking. We tested regions from
0.5\% to 20\% of each test room. The models with AGS keep their whole-mesh accuracy on the same cells, in both $T$ and
$\mathbf{U}$. Every method whose global state is recomputed from the region degrades as the region shrinks. Its regional
error reaches about three times its whole-mesh temperature error (Figure~\ref{fig:region}). Its prediction drifts
several-fold further from its whole-mesh pass (Table~\ref{tab:dcts-main}). BQA reduces the error of Transolver in every
deployment. It also lowers that of AB-UPT slightly. AB-UPT + BQA + AGS leads every accuracy column. These columns, however, evaluate AB-UPT with
every cell as an anchor. In that deployment, its whole-mesh pass costs over a hundred times more than SCANO's. The
reason is that its anchors attend quadratically to one another. Drawing fresh anchors at the training count reduces the pass time to
within a few times SCANO's cost. The error is then slightly higher (Sections~\ref{sec:dcts-cost}
and~\ref{supp:r7-deploy}).

\begin{table*}[t]
\centering
\small
\caption{DCTS benchmark (50 test rooms): five-field mean volume-weighted relative $L^2$ error, normalised by the variation of the true field about its training mean (seed mean $\pm$ 95\% case-bootstrap half-width; Section~\ref{sec:dcts-protocol}). Columns report a single whole-room pass, each method's standard deployment (\emph{own deployment}, coinciding with the single pass for AGS models and averaging covering passes otherwise; Section~\ref{sec:setup-deploy}), a single training subgraph, a compact 5\% region, and regional drift $\delta$. Bold indicates the column best.}
\label{tab:dcts-main}
\begin{tabular}{lccccc}
\toprule
Method & Whole mesh & Own deployment & Subgraph & Region & Region drift $\delta$ \\
\midrule
\multicolumn{6}{l}{\emph{SCANO}} \\
SCANO & 0.395\,{\scriptsize$\pm$\,0.025} & 0.395\,{\scriptsize$\pm$\,0.025} & 0.391\,{\scriptsize$\pm$\,0.025} & 0.475\,{\scriptsize$\pm$\,0.030} & 0.055\,{\scriptsize$\pm$\,0.003} \\
SCANO, anchors outside the message passing & 0.397\,{\scriptsize$\pm$\,0.024} & 0.397\,{\scriptsize$\pm$\,0.024} & 0.396\,{\scriptsize$\pm$\,0.024} & 0.482\,{\scriptsize$\pm$\,0.029} & \textbf{0.023}\,{\scriptsize$\pm$\,0.002} \\
SCANO backbone + BQA & 0.393\,{\scriptsize$\pm$\,0.026} & 0.386\,{\scriptsize$\pm$\,0.026} & 0.391\,{\scriptsize$\pm$\,0.026} & 0.573\,{\scriptsize$\pm$\,0.023} & 0.318\,{\scriptsize$\pm$\,0.025} \\
\midrule
\multicolumn{6}{l}{\emph{Backbones without and with BQA}} \\
Transolver & 0.554\,{\scriptsize$\pm$\,0.027} & 0.457\,{\scriptsize$\pm$\,0.030} & 0.456\,{\scriptsize$\pm$\,0.029} & 1.163\,{\scriptsize$\pm$\,0.062} & 0.996\,{\scriptsize$\pm$\,0.069} \\
Transolver + BQA & 0.431\,{\scriptsize$\pm$\,0.023} & 0.410\,{\scriptsize$\pm$\,0.024} & 0.408\,{\scriptsize$\pm$\,0.024} & 0.878\,{\scriptsize$\pm$\,0.046} & 0.698\,{\scriptsize$\pm$\,0.059} \\
AB-UPT + AGS & 0.378\,{\scriptsize$\pm$\,0.027} & 0.378\,{\scriptsize$\pm$\,0.027} & 0.383\,{\scriptsize$\pm$\,0.026} & 0.454\,{\scriptsize$\pm$\,0.031} & 0.078\,{\scriptsize$\pm$\,0.007} \\
AB-UPT + BQA + AGS & \textbf{0.373}\,{\scriptsize$\pm$\,0.025} & \textbf{0.373}\,{\scriptsize$\pm$\,0.025} & \textbf{0.375}\,{\scriptsize$\pm$\,0.025} & \textbf{0.451}\,{\scriptsize$\pm$\,0.029} & 0.069\,{\scriptsize$\pm$\,0.004} \\
AeroGTO & 0.469\,{\scriptsize$\pm$\,0.041} & 0.463\,{\scriptsize$\pm$\,0.043} & 0.468\,{\scriptsize$\pm$\,0.043} & 1.228\,{\scriptsize$\pm$\,0.059} & 1.087\,{\scriptsize$\pm$\,0.057} \\
\midrule
\multicolumn{6}{l}{\emph{Other boundary encoders}} \\
Transolver + LFE & 0.460\,{\scriptsize$\pm$\,0.020} & 0.419\,{\scriptsize$\pm$\,0.024} & 0.417\,{\scriptsize$\pm$\,0.023} & 0.891\,{\scriptsize$\pm$\,0.050} & 0.715\,{\scriptsize$\pm$\,0.064} \\
GeoTransolver & 0.399\,{\scriptsize$\pm$\,0.025} & 0.392\,{\scriptsize$\pm$\,0.025} & 0.391\,{\scriptsize$\pm$\,0.025} & 0.813\,{\scriptsize$\pm$\,0.031} & 0.634\,{\scriptsize$\pm$\,0.043} \\
GNOT & 0.518\,{\scriptsize$\pm$\,0.016} & 0.441\,{\scriptsize$\pm$\,0.022} & 0.439\,{\scriptsize$\pm$\,0.022} & 0.890\,{\scriptsize$\pm$\,0.057} & 0.688\,{\scriptsize$\pm$\,0.069} \\
Slice aggregation & 0.423\,{\scriptsize$\pm$\,0.025} & 0.406\,{\scriptsize$\pm$\,0.027} & 0.405\,{\scriptsize$\pm$\,0.026} & 1.079\,{\scriptsize$\pm$\,0.062} & 0.942\,{\scriptsize$\pm$\,0.069} \\
\bottomrule
\end{tabular}
\end{table*}

\subsection{Flow fields}
\label{sec:dcts-fields}

\begin{figure*}[tp]
\centering
\includegraphics[width=\textwidth]{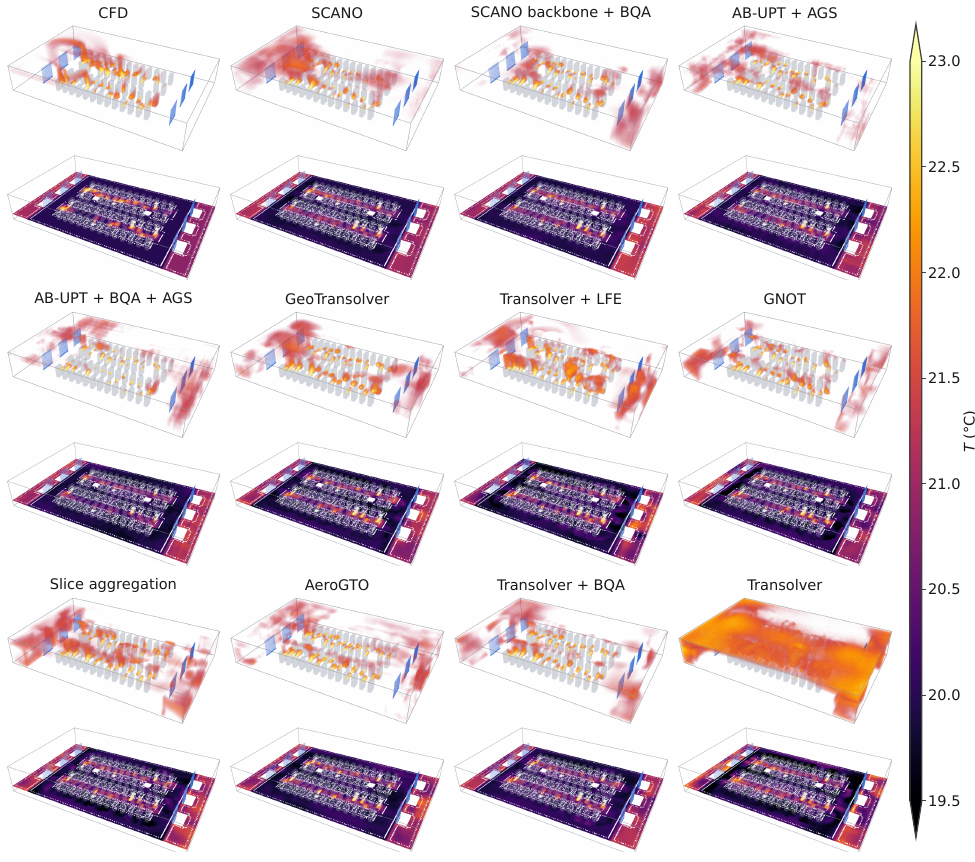}
\caption{Three-dimensional temperature field $T$ in a DCTS test room (room~1 of Figure~\ref{fig:dcts-rooms}). The panels show the
CFD reference and the whole-room predictions of the methods of Table~\ref{tab:dcts-main}. Each model uses the seed of
median validation error. The top of each panel renders the thermal plumes above the 90th percentile of the CFD
temperature. The bottom shows the horizontal cross-section at the height of the server inlets. Racks are grey and
cooling supplies blue.}
\label{fig:dcts-3d}
\end{figure*}

Figure~\ref{fig:dcts-3d} compares the three-dimensional temperature field $T$ of a whole test room. It shows the CFD
reference and the methods of Table~\ref{tab:dcts-main}. In the CFD solution, a distinct exhaust plume rises from each
rack. The plumes feed a warm return layer below the ceiling, which flows toward the cooling units. SCANO,
AB-UPT + BQA + AGS and AB-UPT + AGS reproduce these features plume by plume. The other methods that read the records
capture the plumes with more diffuse edges. The native Transolver sees the boundary only through sampled boundary nodes.
It overheats the upper part of the room. Attaching BQA to the same backbone brings back the individual rack plumes. It
also predicts the heat removed by the cooling units more accurately (Section~\ref{supp:r7-dcts}).

\begin{figure*}[t]
\centering
\includegraphics[width=\textwidth]{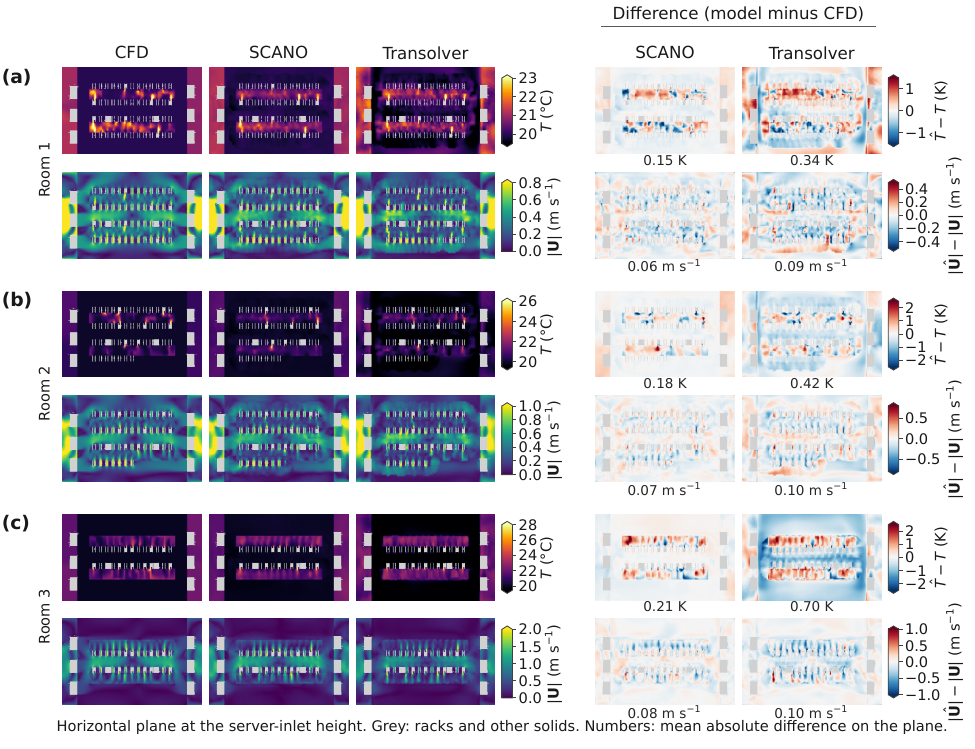}
\caption{Predictions on the horizontal plane at the height of the server inlets, in three held-out DCTS test rooms, (a) Room~1,
(b) Room~2 and (c) Room~3. For each room, the upper row shows the temperature $T$ and the lower row the velocity magnitude $|\mathbf{U}|$. The
columns are the CFD reference, SCANO and the native Transolver, followed by the pointwise differences (model minus CFD). Each model uses the seed of median validation error. Racks and solid obstacles are grey. Numbers below the difference panels give the mean
absolute difference on the plane. Table~\ref{tab:dcts-main} summarises all 50 rooms.}
\label{fig:dcts-rooms}
\end{figure*}

Figure~\ref{fig:dcts-rooms} shows the horizontal plane at the height of the server inlets in three test rooms. SCANO
reproduces the solver's floor-fed cold aisles, the hot aisles behind the racks and the individual exhaust plumes. Its
errors are small and without structure away from the racks. The native Transolver shifts the temperature of entire zones
and hot aisles. Its errors are structured along the rack rows. On the plane, its temperature errors are at least twice SCANO's in
each room.

\begin{figure*}[t]
\centering
\includegraphics[width=\textwidth]{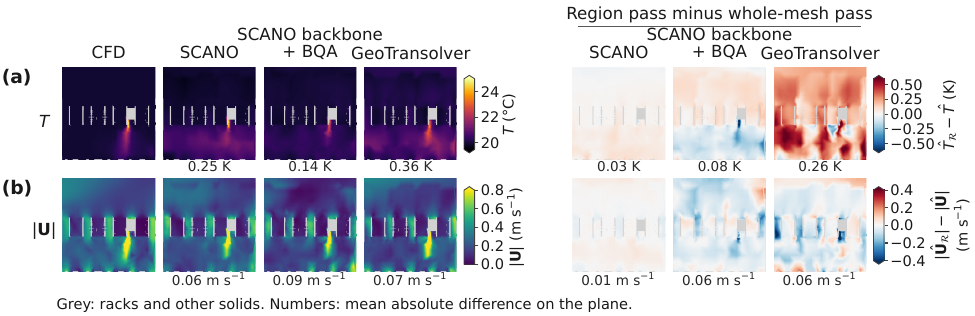}
\caption{A local query in a DCTS test room (room~1 of Figure~\ref{fig:dcts-rooms}). The region is a full-height box within 3\,m
horizontally of the rack with the hottest CFD inlets. It holds 6\% of the room volume. The panels show it at the height
of the server inlets. (a)
Temperature $T$. (b) Velocity magnitude $|\mathbf{U}|$. The left columns show the CFD reference and the standalone
regional predictions of SCANO, the SCANO backbone + BQA and GeoTransolver. The right columns show the region pass minus the
whole-mesh pass for each model (median validation seed). Numbers below the panels give the mean
absolute difference on the plane.}
\label{fig:dcts-roi}
\end{figure*}

Figure~\ref{fig:dcts-roi} illustrates a typical local query of an engineer. The task is to predict the compact region
around the rack with the hottest inlets, 6\% of the room, from a single regional pass. SCANO's global state does not
depend on the queried subset. Its regional prediction therefore closely matches its whole-mesh prediction in both $T$
and $\mathbf{U}$. The SCANO backbone + BQA recomputes its state from the region. Its regional velocity prediction shifts noticeably across
the cold aisle relative to its whole-room pass. GeoTransolver matches SCANO on the whole mesh. In the region, however, it warms the entire region and
drifts further from the CFD reference than in its whole-mesh pass (Proposition~\ref{prop:drift}). In this particular
room, the SCANO backbone + BQA happens to lie closer to the solver's temperature on the plane. This does not hold over
the test split.

\subsection{Response to a single boundary change}
\label{sec:dcts-response}

The paired interventions test whether a trained surrogate captures the response to a single changed boundary condition
on a fixed mesh. A rack's heat load enters through the jump coupling of its inlet and exhaust patches
(Remark~\ref{prop:jump}). Every method receives the same heat-source features and the same paired-intervention training
(Section~\ref{sec:dcts-protocol}).
Table~\ref{tab:dcts-decision} reports the twenty held-out test pairs. Every method captures the response to a doubled
rack power, which also enters through the domain heat-source features (Section~\ref{supp:r7-design}). Over the whole room, AB-UPT + BQA + AGS leads, closely followed by the SCANO-backbone variants and AB-UPT + AGS. Halving
the supply airflow of a cooling unit separates the methods more sharply. The split follows how boundary information
reaches the domain.
Some architectures attend to the boundary records in every layer. These are all variants of the SCANO backbone, the
backbones with BQA, GeoTransolver and GNOT. They capture the room's thermal response for every training seed. The native
Transolver and AeroGTO see the boundary only through sampled boundary nodes. They fail for some seeds. So do AB-UPT +
AGS, whose anchors include only a random subset of the boundary faces, and Transolver + LFE, which reads the records
only at the input. With BQA, Transolver and AB-UPT respond reliably for every seed. Their response error also falls by
more than half.

\begin{table}[!htbp]
\centering
\footnotesize
\setlength{\tabcolsep}{3pt}
\caption{Single-intervention thermal response and server-inlet temperature accuracy on DCTS across the methods of Table~\ref{tab:dcts-main}. \emph{Response}: whole-room temperature response error $E_{\mathrm{resp}}$~\eqref{eq:eresp} (seed mean $\pm$ s.d.\ of the median over ten held-out pairs per intervention type). \emph{Server inlets}: within-room rank correlation and MAE (K) of predicted server-inlet temperatures across the 50 test rooms (each model's own deployment, seed mean). Bold indicates the column best.}
\label{tab:dcts-decision}
\begin{tabular}{lcccc}
\toprule
 & \multicolumn{2}{c}{Response $E_{\mathrm{resp}}$} & \multicolumn{2}{c}{Server inlets} \\
\cmidrule(lr){2-3}\cmidrule(lr){4-5}
 & Rack power & Cooling flow & Rank & MAE \\
Method & doubled & halved & correlation & (K) \\
\midrule
\multicolumn{5}{l}{\emph{SCANO}} \\
SCANO & 0.643\,{\scriptsize$\pm$\,0.024} & 0.400\,{\scriptsize$\pm$\,0.025} & 0.54 & 1.40 \\
SCANO, anchors outside the message passing & 0.612\,{\scriptsize$\pm$\,0.009} & 0.393\,{\scriptsize$\pm$\,0.012} & 0.55 & 1.39 \\
SCANO backbone + BQA & 0.633\,{\scriptsize$\pm$\,0.011} & 0.386\,{\scriptsize$\pm$\,0.008} & 0.55 & 1.38 \\
\midrule
\multicolumn{5}{l}{\emph{Backbones without and with BQA}} \\
Transolver & 0.718\,{\scriptsize$\pm$\,0.054} & 0.854\,{\scriptsize$\pm$\,0.253} & 0.47 & 1.53 \\
Transolver + BQA & 0.734\,{\scriptsize$\pm$\,0.031} & 0.376\,{\scriptsize$\pm$\,0.002} & 0.49 & 1.45 \\
AB-UPT + AGS & 0.615\,{\scriptsize$\pm$\,0.003} & 0.860\,{\scriptsize$\pm$\,0.186} & \textbf{0.66} & 1.07 \\
AB-UPT + BQA + AGS & \textbf{0.608}\,{\scriptsize$\pm$\,0.029} & 0.399\,{\scriptsize$\pm$\,0.077} & 0.66 & \textbf{1.04} \\
AeroGTO & 0.766\,{\scriptsize$\pm$\,0.062} & 0.849\,{\scriptsize$\pm$\,0.262} & 0.52 & 1.60 \\
\midrule
\multicolumn{5}{l}{\emph{Other boundary encoders}} \\
Transolver + LFE & 0.658\,{\scriptsize$\pm$\,0.025} & 0.587\,{\scriptsize$\pm$\,0.358} & 0.47 & 1.46 \\
GeoTransolver & 0.653\,{\scriptsize$\pm$\,0.038} & \textbf{0.371}\,{\scriptsize$\pm$\,0.018} & 0.48 & 1.43 \\
GNOT & 0.746\,{\scriptsize$\pm$\,0.013} & 0.412\,{\scriptsize$\pm$\,0.041} & 0.45 & 1.53 \\
Slice aggregation & 0.661\,{\scriptsize$\pm$\,0.027} & 0.497\,{\scriptsize$\pm$\,0.155} & 0.48 & 1.47 \\
\bottomrule
\end{tabular}
\end{table}

For the operator of a data centre, the decisive question is which servers run hot. We therefore also evaluate the server
inlet temperatures from each model's own deployment (Table~\ref{tab:dcts-decision}). AB-UPT + BQA + AGS and AB-UPT + AGS
reach the highest rank correlation of the inlets. They also have the lowest inlet temperature error over a room. All three
SCANO-backbone variants follow, ahead of all remaining methods.

\subsection{Cost of training and deployment}
\label{sec:dcts-cost}

\begin{figure*}[t]
\centering
\includegraphics[width=\textwidth]{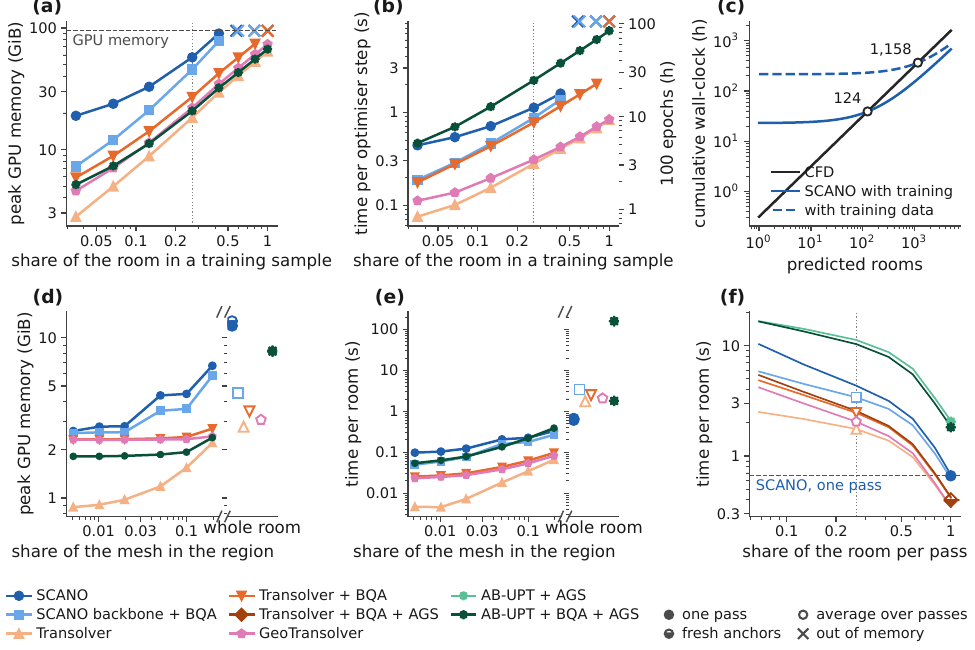}
\caption{Cost of training and deployment on million-cell DCTS rooms, on an NVIDIA GPU with 96~GB of memory. Values are medians
over the four rooms of Table~\ref{tab:dcts-time}. (a) Peak GPU memory against the sampled fraction of the room. (b) Step
time against the same fraction. The right axis converts it into the optimiser time of 100 epochs. Crosses mark runs that exceed the
memory. The dotted line marks the training setting. (c) Cumulative wall-clock hours against the number of predicted
rooms, for CFD and for SCANO. SCANO is counted with training alone and with training plus data generation. Markers show
the break-even counts. (d) Inference memory per room. (e) Inference time per room. Curves are compact regions. Markers
are full rooms, predicted in one pass with all cells as anchors, with fresh anchors, or as an average over
passes. (f) Time per room of the multi-pass average on one GPU against the share of the room in one pass. The dotted
line marks the training share. Open markers are the deployment of the models whose state follows the sample. Filled
markers are the single pass of the models with AGS, half filled with fresh anchors. The dashed line is SCANO's single
pass.}
\label{fig:cost}
\end{figure*}

In training (Figure~\ref{fig:cost}a,b), peak memory and step time both grow with the share of the room in a sample. The
hundred epochs of Table~\ref{tab:dcts-main} take about one day on a single GPU. From a seed ratio of 0.2 upward, the
SCANO backbone exceeds the GPU memory. Sampled training is therefore what makes million-cell rooms trainable with
message passing and BQA. On Transolver, a whole room fits in memory. There, sampled training costs no accuracy once BQA
and AGS are both attached (Section~\ref{sec:test-c3}).

Table~\ref{tab:r7-cost} lists the training and deployment cost of all DCTS methods. All are measured in the same way on
one GPU. At the common budget of about 0.8 million parameters, a training step at the training ratio takes from a
quarter of a second to a few seconds. SCANO's step takes about one second. It carries the message passing over the
sampled subgraph and over its anchor sample.

\begin{table}[!htb]
\centering
\footnotesize
\setlength{\tabcolsep}{3pt}
\caption{Cost of the DCTS methods of Table~\ref{tab:dcts-main} on one GPU. The table gives the parameters, the training step at the training ratio of 0.05 (median of six steps) with its peak memory, and the deployment of a whole room with its peak memory. The deployment time is the median over the four rooms of Table~\ref{tab:dcts-time}. For SCANO and the models with AGS, the deployment is one pass with the state over every cell. Where a fresh draw of the training anchor count is much faster, its time is given in parentheses. For the models whose state follows the sample, the deployment is the average over covering training-sized passes.}
\label{tab:r7-cost}
\begin{tabular}{P{0.36\linewidth}ccccc}
\toprule
 & & \multicolumn{2}{c}{Training step} & \multicolumn{2}{c}{Whole room} \\
\cmidrule(lr){3-4}\cmidrule(lr){5-6}
Method & Parameters ($10^3$) & Time (s) & Memory (GiB) & Time (s) & Memory (GiB) \\
\midrule
\multicolumn{6}{l}{\emph{SCANO}} \\
SCANO & 811 & 1.13 & 57 & 0.61 & 13 \\
SCANO, anchors outside the message passing & 811 & 0.95 & 46 & 0.96 & 23 \\
SCANO backbone + BQA & 811 & 0.93 & 46 & 3.40 & 5 \\
\midrule
\multicolumn{6}{l}{\emph{Backbones without and with BQA}} \\
Transolver & 811 & 0.35 & 18 & 1.76 & 3 \\
Transolver + BQA & 859 & 0.82 & 27 & 2.48 & 4 \\
AB-UPT + AGS & 821 & 2.51 & 15 & 173.30 (2.05) & 9 \\
AB-UPT + BQA + AGS & 856 & 2.25 & 21 & 158.97 (1.79) & 9 \\
AeroGTO & 798 & 1.23 & 39 & 5.31 & 6 \\
\midrule
\multicolumn{6}{l}{\emph{Other boundary encoders}} \\
Transolver + LFE & 787 & 0.75 & 22 & 2.36 & 3 \\
GeoTransolver & 805 & 0.34 & 22 & 2.07 & 3 \\
GNOT & 794 & 0.25 & 23 & 0.86 & 3 \\
Slice aggregation & 818 & 0.40 & 23 & 2.54 & 5 \\
\bottomrule
\end{tabular}
\end{table}

\begin{table}[!htb]
\centering
\caption{Time per DCTS test room on one NVIDIA GPU with 96 GB of memory and the CPU cores attached to it. We timed four test
rooms of 0.90--1.03 million cells. Each lies at the middle of one quarter of the cell-count distribution. The table
gives the median and range over the four, in seconds. Software: OpenFOAM v2506, OpenMPI 5.0.8, PyTorch 2.10. Meshing,
extraction and conversion run on one process. The other CFD steps run in parallel with MPI.}
\label{tab:dcts-time}
\begin{tabular}{lcc}
\toprule
 & CFD (reference) & SCANO \\
\midrule
Mesh (blockMesh, snappyHexMesh, decomposition)$^a$ & 235 [225--239] & same mesh \\
5,000 SIMPLE iterations, decomposition and reconstruction$^a$ & 904 [816--965] & --- \\
\quad of which the iterations (solver clock) & 844 [766--891] & --- \\
One solver iteration, for the boundary values$^a$ & --- & 54 [43--61] \\
Geometry and boundary attributes, as built for the dataset$^b$ & --- & 185 [173--198] \\
Reading the converted case & --- & 0.10 [0.09--0.10] \\
Inference, one pass over the whole mesh on the GPU$^c$ & --- & 0.60 [0.58--0.69] \\
\midrule
Total & 1,136 [1,050--1,201] & 467 [458--499] \\
\bottomrule
\end{tabular}
\par\smallskip\footnotesize
$^a$ Includes the start-up of the MPI steps (mesh 110 [99--132], solver 24 [14--40], the solver iteration 23 [14--30]).
$^b$ An upper bound. It includes reading the converged field and a wall-distance field that SCANO does not use. $^c$ The
SCANO model of Table~\ref{tab:dcts-main}, timed afterwards on one otherwise idle GPU of the same kind. Median of seven
warm passes; loading the model and the first pass are excluded.
\end{table}

At deployment (Figure~\ref{fig:cost}d--f), SCANO predicts a full room in one pass of well under a second. Its global
state is evaluated over all cells. A compact region of interest with its message-passing halo takes a fraction of that
time and memory (Figure~\ref{fig:tta-roi}). Models whose state follows the sample instead average over several passes
at the training ratio. On one GPU, this average takes about one to five seconds per room (Table~\ref{tab:r7-cost}).
Larger passes make it faster (Figure~\ref{fig:cost}f). Over a whole-room pass, however,
the error of these models grows. It grows most for Transolver without BQA (Tables~\ref{tab:dcts-main}
and~\ref{tab:c132-tta}). The models with
AGS need no average. Each predicts the whole room in a single pass. The passes of the multi-pass average can be distributed across the GPUs
of a node with nearly linear speed-up. On four GPUs, they match SCANO's wall-clock time at three to six times its total
GPU time (Figure~\ref{fig:cost-gpus}). AB-UPT + BQA + AGS uses a quadratic anchor self-attention. It takes under three
minutes per room with all cells as anchors and under two seconds with fresh anchors (Table~\ref{tab:r7-cost}).

End to end, a SCANO prediction needs four steps: the mesh, one solver iteration for the initial boundary values,
feature extraction and one forward pass. The CFD solver starts a new room from its initial fields. SCANO replaces its
thousands of iterations by one and cuts the wall-clock time to less than half that of the solve
(Table~\ref{tab:dcts-time}). The mesh takes about half of SCANO's time. The CFD needs the same mesh. The extraction of
the boundary attributes takes most of the rest. The forward pass itself takes under a second. The training cost
is recovered after about 120 rooms, and data generation plus training after 1,200.

\section{Discussion}
\label{sec:discussion}

\subsection{How the two components act on the error}
\label{sec:discussion-error}

The experiments of Section~\ref{sec:results} attribute each gain in accuracy and consistency to one term
of~\eqref{eq:decomposition}. BQA removes the boundary term. Every patch reaches every domain point with the
weight of its area. BQA most benefits backbones that otherwise see the boundary only through sampled nodes. It also
lets trained models recover the measure structure of the boundary data. AGS removes the state term. Regional predictions
stay aligned with whole-mesh predictions, up to the Monte Carlo error of the anchors. A state recomputed from the region drifts as
the region shrinks. With both modules, the bias of a sample-dependent state disappears. Transolver + BQA + AGS trained
on DCTS subgraphs then matches the same model trained on whole rooms (Section~\ref{sec:test-c3}). When boundary conditioning
is complete, the sampling compromise forced by GPU memory entails no loss of accuracy. The SCANO backbone needs more memory for a
whole-room step than a single GPU has. For it, sampled training at linear cost yields an accurate surrogate for any
region of a room. This surrogate captures the response to a halved cooling airflow. It answers a room query in well
under a second of network time, and end to end in under half the time of a CFD solve.

\subsection{Scope of the construction}
\label{sec:scope}

Our construction applies wherever boundary conditions can be written as typed face records. This covers Dirichlet,
Neumann and Robin conditions, inflows and outflows. It also covers coupled heat sources such as servers, which enter as
jump conditions with a measure of their own. C1 holds by design for any backbone that reads the boundary through BQA. C3
holds whenever the global state is estimated independently of the query sample. The measure representation is exact for
linear transport. On the buoyant RANS flow of DCTS, it serves as a design guide, and we verified its consequences
empirically. Likewise, the slice budget $G$ bounds the resolution of single-patch responses in general. On DCTS, a
modest $G$ already matches a much larger token budget. The cost of violating C1--C4 grows with the number of small,
independently varying patches and with the field's sensitivity to them. We focus on steady states. The
two benchmarks show that the conclusions carry beyond DCTS.

\section{Conclusions}
\label{sec:conclusions}

A single training pass over a million-cell engineering mesh approaches or exceeds the memory of a GPU. Neural operators
are therefore trained on subsamples of the mesh as a compromise. They are then deployed on the full mesh or on any
region of interest. Each pass thus sees only a fraction of the boundary conditions that govern the solution. We wrote
the solution map as an integral against type-dependent boundary measures. This showed that the conditioning can fail
in three ways under sampled training. It also led to four measurable consistency conditions. SCANO meets these conditions at
linear cost through two plug-and-play modules that fit any operator backbone. Boundary Quadrature Attention (BQA)
aggregates typed, area-weighted face records into slice tokens. Every point reads these tokens in every layer. The
Anchored Global State (AGS) estimates the global state from anchors independent of the sample in training. At
deployment, it uses all cells.

Our experiments confirmed all four predictions made before training. These concern the fraction of the boundary that a
sample observes, the refinement plateau of count weighting, the deployment drift of a sample-dependent state and the
rank floor of a global code. Encoders that read the boundary through a global code fell well short of per-point
boundary readouts. We integrated BQA into
seven backbones. On the two benchmarks, it reduced the whole-mesh error of every backbone over its native baseline. We integrated AGS into six backbones.
On all three data sets, it kept regional predictions aligned with the whole-mesh predictions up to the error of the anchors. It also enabled
sampled training of Transolver + BQA + AGS on DCTS to match training on whole rooms.

Boundary conditioning for neural operators is therefore more than a choice of input encoding. It is a question of
consistency with sampled training. Once boundary conditions are written as a quadrature of type-dependent measures, the
failure modes follow for any backbone. So do the consistency conditions and the architectural remedies. Together they
form a quantitative checklist governed by the slice-token and anchor budgets. We expect
this framework to benefit boundary-driven systems with localised sources on large meshes. Examples are electronics
cooling and building ventilation.

\section*{Data and code availability}
DCTS, the controlled benchmark with its generator, and the code of SCANO and of the training-free checks will be made
available upon acceptance of the paper.

\section*{CRediT authorship contribution statement}
Tianyu Li: Conceptualization, Methodology, Software, Validation, Visualization, Writing~-- original draft,
Writing~-- review \& editing.
Zhiwei Cao: Data curation, Formal analysis, Writing~-- review \& editing.
Qingang Zhang: Data curation, Formal analysis, Writing~-- review \& editing.
Ruihang Wang: Data curation, Formal analysis, Writing~-- review \& editing.
Binyang Song: Methodology, Validation, Supervision, Writing~-- review \& editing.
Yonggang Wen: Supervision, Funding acquisition, Project administration, Writing~-- review \& editing.

\section*{Declaration of competing interest}
The authors declare that they have no known competing financial interests or personal relationships that could have
appeared to influence the work reported in this paper.

\section*{Acknowledgements}
This research is supported by the Ministry of Education (MOE), Singapore, under its Academic Research Fund Tier 1
project `Towards Resilient Data Centre Operations via Physics-Informed Predictive Maintenance' (Award No. RG104/23).
This research is also supported by the RIE2025 Industry Alignment Fund -- Industry Collaboration Projects (IAF-ICP)
(Award I2301E0026), administered by A*STAR, as well as supported by Alibaba Group and NTU Singapore through
Alibaba-NTU Global e-Sustainability CorpLab (ANGEL).
The authors thank the National Supercomputing Centre (NSCC), Singapore, for additional support.
T.L. gratefully acknowledges the financial support of the Interdisciplinary Graduate Programme (IGP), Graduate
College, Nanyang Technological University, Singapore, through the NTU Research Scholarship.

\section*{Role of the funding source}
The funding sources provided computational resources for this work and had no other role in the study design, in the
collection, analysis and interpretation of the data, in the writing of the article, or in the decision to submit it
for publication.

\section*{Declaration of generative AI and AI-assisted technologies in the manuscript preparation process}
During the preparation of this work the authors used Claude (Anthropic) to assist in editing the text.
After using this tool, the authors reviewed and edited the content as needed and take full responsibility for the
content of the published article.

\bibliographystyle{elsarticle-num}
\bibliography{references}

\clearpage
\setcounter{page}{1}
\renewcommand{\thepage}{S\arabic{page}}
\setcounter{section}{0}\setcounter{table}{0}\setcounter{figure}{0}\setcounter{equation}{0}
\setcounter{proposition}{0}\setcounter{algorithm}{0}
\renewcommand{\thesection}{S\arabic{section}}\renewcommand{\theHsection}{S\arabic{section}}
\renewcommand{\thetable}{S\arabic{table}}\renewcommand{\theHtable}{S\arabic{table}}
\renewcommand{\thefigure}{S\arabic{figure}}\renewcommand{\theHfigure}{S\arabic{figure}}
\renewcommand{\theequation}{S\arabic{equation}}\renewcommand{\theHequation}{S\arabic{equation}}
\renewcommand{\theproposition}{S\arabic{proposition}}\renewcommand{\theHproposition}{S\arabic{proposition}}
\renewcommand{\thealgorithm}{S\arabic{algorithm}}\renewcommand{\theHalgorithm}{S\arabic{algorithm}}
\begin{center}
{\Large\bfseries Supplementary material}
\end{center}

Section, table, equation and reference numbers without the prefix S refer to the manuscript; citations use its
reference list.

\section{Notation}
\label{supp:notation}

Table~\ref{tab:notation} lists the symbols of the manuscript in the order in which its sections introduce them.

\begin{table}[htbp]
\centering
\footnotesize
\setlength{\tabcolsep}{4pt}
\caption{Symbols of the manuscript, grouped by the section that introduces them.}
\label{tab:notation}
\begin{tabular}{P{0.22\linewidth}P{0.74\linewidth}}
\toprule
Symbol & Meaning \\
\midrule
\multicolumn{2}{l}{\emph{Sampled training (Section~\ref{sec:consistency})}} \\
$\Omega_h$, $w_x$ & the cells of a mesh and the volume of cell $x$ \\
$u$, $\hat u_\theta(x;S)$ & reference solution; prediction at $x$ of the model with parameters $\theta$ given the sample $S$ \\
$S$, $\pi_x$ & sample of a training step or a deployment; probability that the design includes $x$ \\
$\rho$, $\rho_A$, $k$ & sampling ratio of the seeds, ratio of the anchor seeds, number of hops grown around them \\
$\ell$, $\mathcal{L}$, $\mathcal{L}_S$ & pointwise loss; whole-mesh and sampled objectives; minimisers $\theta^\ast$ and $\hat\theta$ \\
$N_a$, $A$ & number of anchors and the anchor set, drawn independently of $S$ \\
\midrule
\multicolumn{2}{l}{\emph{Boundary data and measures (Sections~\ref{sec:measure}--\ref{sec:failures})}} \\
$D$, $N$, $R$ & Dirichlet, Neumann, Robin patches; data $u_D$, $q$, $u_\infty$; diffusivity $\kappa$, velocity $\mathbf{b}$, Robin coefficient $\alpha$ \\
$g$, $g_p$, $g_y$ & boundary data of a case, of patch $p$ and of record $y$ \\
$\mu^{\mathrm{exit}}_x$, $\mu^{\mathrm{kill}}_x$, $\nu^{\mathrm{loc}}_x$ & exit, killing and local-time measures of a
diffusion started at $x$; $t_D$ its exit time through $D$; $L^R$, $L^N$ its local times on $R$ and $N$ \\
$P$, $\dot V$, $\Delta T_s$, $\rho_{\mathrm{air}}$, $c_p$ & server power, volumetric flow, temperature rise, air density and
heat capacity; $\nu_s$ normalised area measure of a server inlet \\
$H(S)$, $K_p$, $\Sigma_p$ & patches hidden from sample $S$; response operator and data covariance of patch $p$; $\|\cdot\|_Q$ norm over the queried cells \\
$s_j(S)$, $\omega_j$, $c(x)$, $f$ & global state of slice $j$ computed from the sample $S$, slice weights, context weights and features \\
$\varrho$, $K(x,\cdot)$, $\lambda_x$ & density of the boundary records; area density of the boundary measure seen from $x$;
aggregation logits \\
$\zeta$, $d$, $\sigma_i$ & a global code of the boundary data, its width, and the singular values of the solution
operator's derivative \\
\midrule
\multicolumn{2}{l}{\emph{Error decomposition and conditions (Sections~\ref{sec:decomposition}--\ref{sec:conditions})}} \\
$s$, $\beta$ & the global state a point reads and the boundary input that reaches it, as the measure $\sum_y a_y
\delta_{\phi_y}$ \\
$s^\ast$, $\beta^\ast$ & state from every cell and boundary input from every record with its area \\
$\varepsilon^\ast$, $\gamma(\hat\theta)$ & class error and training term; $L_s$, $L_\beta$, $L_\psi$ Lipschitz constants;
$L_\Omega$ domain size; $d_{\mathrm{BL}}$ bounded-Lipschitz distance; $\mathcal{Y}$, $d_{\mathcal{Y}}$ record space and its metric \\
C1--C4 & information completeness, mesh consistency, sample consistency, response fidelity \\
\midrule
\multicolumn{2}{l}{\emph{SCANO (Section~\ref{sec:method})}} \\
$y$, $\phi_y$, $a_y$ & boundary face, its record $[\mathbf{x}_y,\mathbf{n}_y,\tau_y,g_y]$ (position, normal, type,
value) and its area \\
$h_y$, $W$, $\vartheta$ & lifted record, slice projection and temperature of the slice assignment \\
$h_\Gamma$, $h_b$ & size of a patch and spacing of the boundary records \\
$l$, $L$ & layer index and number of layers \\
$G$, $\omega_g(y)$, $M_g$, $z_g$ & number of boundary tokens, slice weights, slice masses and tokens;
$M_{\mathrm{tot}}=\sum_g M_g$; $\epsilon$ regulariser \\
$o_l$, $o^N_l$, $M_{\mathrm{ref}}$ & normalised and unnormalised readouts in layer $l$; reference area of the
second; queries, keys and values $\mathrm{q}_l, \mathrm{k}_l, \mathrm{v}_l$ of head width $d_{\mathrm{k}}$ \\
$s_{l,j}$, $\omega_{l,j}$, $c(x)$, $\mathcal{C}$ & global state of slice $j$ in layer $l$, slice weights of the domain,
context weights and context set \\
\midrule
\multicolumn{2}{l}{\emph{DCTS fields (Section~\ref{sec:dcts-million})}} \\
$\mathbf{U}=(U_x,U_y,U_z)$, $|\mathbf{U}|$ & Reynolds-averaged velocity and its magnitude \\
$p'$, $T$ & kinematic pressure without its hydrostatic part; temperature \\
\midrule
\multicolumn{2}{l}{\emph{Metrics (Section~\ref{sec:setup-deploy})}} \\
$\mathcal{R}$, $\mathcal{R}^+$ & a compact region of interest and the region with its message-passing halo \\
$\mathcal{D}$, $Q_{\mathcal{D}}$ & a deployment and its deployed cells \\
$\delta$ & drift: relative distance of a prediction from the model's own whole-mesh prediction \\
$\Delta u$, $\Delta\hat u$, $E_{\mathrm{resp}}$ & true and predicted change for a paired intervention; response
error \\
\bottomrule
\end{tabular}
\end{table}

\section{Implementation and hyperparameters}
\label{supp:impl}

Every method is trained with the procedure of its data set (Section~\ref{sec:setup-training}). It is evaluated with
multiple training seeds. Table~\ref{tab:hyper} lists the settings of the procedure and of SCANO. The backbones keep the
hyperparameters of their authors. Their width is set so that every model lies within ten percent of SCANO's parameter
count. Algorithms~\ref{alg:train} and~\ref{alg:deploy} give one training step and the deployments.

When BQA and AGS are added to the other backbones (Section~\ref{sec:hosts}), their placement follows each backbone's
design. GeoTransolver attends to a geometry context of its own. On it, BQA adds a per-point readout of the boundary
beside that context. GNOT's cross-attention already reads every record in every layer, weighted by count. On the public
benchmark, the records are evenly spaced, and BQA is added beside this attention. On the controlled benchmark and on
DCTS, the records are graded, and BQA replaces it. There, two readouts of the same records weighted by count and by area disagree
(Proposition~\ref{prop:count}). On MeshGraphNets, BQA is applied after every third message-passing step. AB-UPT locates
points with rotary encodings. The queries and keys of BQA are rotated by the same encodings. Under AGS, its anchors are
drawn from every node, boundary faces included. They are weighted by cell volume on the controlled benchmark and by
count on the other two data sets. AeroGTO's coordinates are normalised by the fixed extent of the domain instead of the
sample's bounding box. Its tokens thus attend to the same anchors. The output projection of the unnormalised readout of
BQA is initialised to zero. Its exponent is bounded at $e^{30}$. The bound leaves the readout unchanged below it and
finite in single precision.

For whole-room training on DCTS (Section~\ref{sec:setup-deploy}), the baselines of Table~\ref{tab:dcts-tax} recompute
the layer activations in the backward pass instead of storing them. The arithmetic of the forward and backward passes
stays identical to standard training. So do the losses, the gradients and the trained weights. Only the peak memory
falls and the step takes longer. A whole-room step of Transolver + BQA + AGS then needs roughly 32~GiB, which fits on a
single GPU, and 3.5~s. Figures~\ref{fig:bc-stats}a and~\ref{fig:cost} time standard steps that store the activations.
All models trained on samples use such steps. The wall-clock times per epoch in Table~\ref{tab:ablations} include the
recomputation.

\begin{table}[htbp]
\centering
\footnotesize
\setlength{\tabcolsep}{3pt}
\caption{Training procedure and SCANO's settings on the three data sets. Each method uses the better learning rate of $10^{-3}$
and $3\times10^{-4}$ on the validation set (Section~\ref{sec:setup-training}). The table gives SCANO's rate.}
\label{tab:hyper}
\begin{tabular}{P{0.25\linewidth}P{0.22\linewidth}P{0.22\linewidth}P{0.22\linewidth}}
\toprule
& Controlled benchmark & Public benchmark & DCTS \\
\midrule
\multicolumn{4}{l}{\emph{Training}} \\
Sample of a step & seeds at ratio 0.12 grown by one hop, or a uniform subset of 12\% (sparse sampling: 0.02) & seeds at 0.05 grown by two hops (sparse sampling: 0.01 and two hops, 0.005 and one hop) & seeds at 0.05 grown by two hops \\
Loss & relative $L^2$ over the sample & mean squared error & mean squared error, plus the same on the change of $T$ over a paired intervention (weight 1) \\
Epochs & 80 & 60 & 100 \\
Optimiser & \multicolumn{3}{P{0.70\linewidth}}{AdamW, weight decay $10^{-4}$, gradient norm clipped at 1, one case per step} \\
Learning rate (SCANO) & $10^{-3}$ & $3\times10^{-4}$ & $10^{-3}$ \\
Schedule & linear warm-up over a tenth of the steps, cosine decay to a tenth & warm-up over 1,000 steps, cosine decay to zero & as the public benchmark \\
Weight average & exponential, 0.9995 & none & none \\
\midrule
\multicolumn{4}{l}{\emph{SCANO backbone}} \\
Layers, width, heads & \multicolumn{3}{P{0.70\linewidth}}{4 layers of width 96 with 4 heads, 32 domain slices, MLP ratio 2} \\
Message passing & \multicolumn{3}{P{0.70\linewidth}}{one graph-network sub-layer of three hops per layer: receptive field and halo of 12 hops} \\
Parameters & 777,313 & 777,121 & 810,601 \\
\midrule
\multicolumn{4}{l}{\emph{Boundary quadrature attention (BQA)}} \\
Records & 6,144 boundary faces & 1,000 boundary points & about $5\times10^5$ boundary faces \\
Measure $a_y$ & face area & arc length & face area (scaled by normalised $|\mathbf{U}|$ on server outlets) \\
Type & Dirichlet, Neumann, Robin & Dirichlet, Neumann, Robin & physical role of the patch \\
Tokens $G$ & one per record & one per record & 64 \\
Readouts & normalised & normalised & normalised and unnormalised, $M_{\mathrm{ref}}=4{,}000\,\mathrm{m}^2$ \\
Slicing & \multicolumn{3}{P{0.70\linewidth}}{temperature $\vartheta$ learned per head (initialised at 0.5); $\epsilon = 10^{-8}$} \\
\midrule
\multicolumn{4}{l}{\emph{Anchored global state (AGS)}} \\
Anchors in training & a second $k$-hop sample at the training ratio, through the message passing (uniform samples: 4,096 cells, uniform) & the whole mesh, through the message passing (samples of 5\%: 1,024 points, stratified, outside it) & a second two-hop sample, seeds at 1.25\% of the cells, through the message passing \\
Weights & cell volume & area of the point's cell & cell volume \\
Deployment & \multicolumn{3}{P{0.70\linewidth}}{whole mesh: every cell; subgraph or region: fresh anchors as in training} \\
\bottomrule
\end{tabular}
\end{table}

\begin{algorithm}[htbp]
\caption{One training step of SCANO on a sample of the mesh.}
\label{alg:train}
\begin{algorithmic}[1]
\Require case: cells $\Omega_h$ with volumes $w_x$, boundary records $\{(\phi_y,a_y)\}$, reference $u$; sampling
ratio $\rho$ and hops $k$; anchor ratio $\rho_A$
  \State $S \gets$ seeds drawn at ratio $\rho$ (cells and boundary faces; Section~\ref{sec:r0s-data}), grown by $k$ hops of the mesh graph
  \State $A \gets$ seeds drawn from $\Omega_h$ at ratio $\rho_A$ independently of $S$, grown by $k$ hops ($A=\Omega_h$ for
$\rho_A=1$); weights $c(x)=w_x$ \Comment{without message passing: cells drawn uniformly}
  \State lift the records, $h_y \gets \mathrm{MLP}(\phi_y)$; per head $\omega_g(y)$, masses $M_g$ and tokens $z_g$ as in
Section~\ref{sec:slices} \Comment{once per pass}
  \State $f_0(x) \gets$ embedding of the node features, $x\in S\cup A$
\For{$l = 1,\dots,L$}
  \State $f_l(x) \gets$ message-passing sub-layer on the graph of $S$ and, separately, on the graph of $A$
  \Comment{no edge joins $S$ and $A$}
  \State global state $s_{l,j} \gets \sum_{x\in A} c(x)\,\omega_{l,j}(x) f_l(x)
         \big/ \sum_{x\in A} c(x)\,\omega_{l,j}(x)$
  \State $f_l(x) \gets f_l(x) + \text{slice attention}(f_l(x); s_l)$, $x\in S\cup A$
  \State $f_l(x) \gets f_l(x) + o_l(x) + o^N_l(x)$, $x\in S\cup A$ \Comment{readouts of the tokens
$z_g$ ($o^N_l$ on DCTS)}
\EndFor
  \State $\hat u(x) \gets$ decoder$(f_L(x))$ for $x\in S$; loss over $S$; one AdamW step
\end{algorithmic}
\end{algorithm}

\begin{algorithm}[htbp]
\caption{Deployment of a trained model.}
\label{alg:deploy}
\begin{algorithmic}[1]
\Require trained model; mesh $\Omega_h$; records; a region of interest $\mathcal{R}\subseteq\Omega_h$ (optional)
\If{whole mesh}
  \State one pass with $S = A = \Omega_h$: the state over every cell, the tokens over every record
\ElsIf{region $\mathcal{R}$}
  \State $\mathcal{R}^{+} \gets \mathcal{R}$ grown by the receptive field of the message passing (12 hops)
  \State one pass over $\mathcal{R}^{+}$ with a fresh anchor sample of $\Omega_h$ (Algorithm~\ref{alg:train}, lines
2--10); keep the prediction on $\mathcal{R}$
\EndIf
  \Statex \emph{A model whose state is computed from the sample
(multi-pass average):}
  \State permute $\Omega_h$, cut it into chunks at the training ratio, grow each chunk by the
training hops
  \State predict every chunk independently and average the predictions per cell; on several GPUs the chunks
are dealt round-robin, each GPU keeps a sum and a count per cell, and the sums are added at the end (the one-GPU result
up to round-off)
\end{algorithmic}
\end{algorithm}

\section{The controlled benchmark: further details}
\label{app:r0s}

This section specifies the controlled benchmark of Section~\ref{sec:r0s-data}. Table~\ref{tab:r0s-data} gives the
problem, its meshes, patches and interventions. Table~\ref{tab:r0s-sampling} gives the two families of training samples
and the boundary coverage they provide. Table~\ref{tab:r0s-deploy} lists the deployments on which the trained models are
evaluated. The results for sparse sampling and the sampling gaps against whole-mesh training are in
Section~\ref{supp:r0s} (Tables~\ref{tab:r0s-strong}--\ref{tab:r0s-tax}).

\begin{table}[t]
\centering
\small
\caption{The controlled benchmark. Face counts refer to the training mesh. The area strata are used to report errors and
responses.}
\label{tab:r0s-data}
\begin{tabular}{P{0.24\linewidth}P{0.69\linewidth}}
\toprule
Problem & $\Omega=[0,1]^3$; $-\nabla\cdot(\kappa\nabla u)+\mathbf{b}\cdot\nabla u=0$ with $\kappa=1$, no source;
$\mathbf{b}=\mathrm{Pe}\,\nabla\times(0,\psi,0)$, $\psi=\sin(\pi x)\sin(\pi z)\,(1+\tfrac12\cos\pi y)/(1.5\pi)$,
$\mathrm{Pe}=10$ \\
Discretisation & finite volumes, central convection, face fluxes divergence-free; cell P\'eclet number at most about
0.26 on the truth mesh \\
Meshes & graded towards every wall (tanh stretching, factor 1.5); truth $64^3$ (262,144 cells, 24,576 faces); training $32^3$
(32,768 cells, 6,144 faces; wall cells about five times thinner than centre cells); deployment mesh $40^3$
graded in reverse (coarse at the walls); every mesh's grid lines are a subset of the truth mesh's \\
Patches & $N_p\in\{8,16,32,48\}$ patches per layout, a quarter of the layouts each; rectangles on a $16\times16$ lattice per
wall, $w\times h$ lattice cells with $w,h\in\{1,2,4,8\}$ and aspect ratio at most 4 \\
Area strata & tiny 4--8 faces, small 16--32, medium 64, large 128--256 \\
Types & Dirichlet $\max(2,\mathrm{round}(0.3N_p))$, Robin $\mathrm{round}(0.25N_p)$, Neumann the rest; Robin
coefficient $\alpha\in\{1,10\}$; insulated wall outside the patches \\
Values & constant, plus a linear profile on half of the patches; Dirichlet amplitude
$\mathcal{N}(0,1)\cdot\min\bigl(3,(A_{\mathrm{med}}/A_p)^{1/4}\bigr)$; Neumann flux $\pm q_{\mathrm{tot}}/A_p$, $q_{\mathrm{tot}}$ log-uniform in
$[0.02,0.5]$, positive with probability 0.8; Robin external value $\mathcal{N}(0,1)$ \\
Layouts & 128 for training, 16 held out, 32 siblings for the type switches \\
Cases & training 2,048 and validation 256 (training layouts); test 256 on training layouts, 256 on held-out layouts
and 256 interventions (64 base cases on 32 layouts, half of them held out, times four interventions) \\
Interventions & Dirichlet value $+1$ training standard deviation; Neumann total $\times2$; Robin external value $+1$
training standard deviation; one Dirichlet patch switched to an insulated wall \\
\bottomrule
\end{tabular}
\end{table}

\begin{table}[t]
\centering
\small
\caption{Training samples on the controlled benchmark. The columns give the share of the cells and of the boundary faces in one
sample and the number of nodes per sample. The last column gives the share of patches that a sample misses entirely.
A model that receives the boundary only as sampled nodes misses these patches (512 draws on the training mesh).}
\label{tab:r0s-sampling}
\begin{tabular}{llcccc}
\toprule
Family & Setting & Cells & Faces & Nodes & Patches missed \\
\midrule
$k$-hop, training ratio & ratio 0.12, one hop & 58\% & 12\% & 19.8k & 17\% \\
$k$-hop with the faces of its cells & ratio 0.12, one hop & 58\% & 58\% & 22.6k & 1\% \\
uniform, training ratio & ratio 0.12 & 12\% & 12\% & 4.7k & 18\% \\
$k$-hop, sparse sampling & ratio 0.02, one hop & 13\% & 2\% & 4.3k & 52\% \\
uniform, sparse sampling & ratio 0.02 & 2\% & 2\% & 0.8k & 52\% \\
\bottomrule
\end{tabular}
\end{table}

\begin{table}[t]
\centering
\small
\caption{Deployments on the controlled benchmark (the last row only in the measure studies).}
\label{tab:r0s-deploy}
\begin{tabular}{P{0.26\linewidth}P{0.68\linewidth}}
\toprule
Deployment & Settings \\
\midrule
Sample as in training & the training family and ratio; 5 draws \\
Larger samples & ratio $\times4$ and $\times10$ (at most the whole mesh); 3 draws \\
Whole mesh, one pass & SCANO reads its whole-mesh global state \\
Region of interest & compact regions holding 0.02, 0.05, 0.12 and 0.3 of the cells, with the halo of the message passing
and fresh anchors for the models with AGS; 3 draws \\
Another mesh & the $40^3$ mesh graded in reverse (coarse at the walls); its grid lines are a subset of those of the
truth mesh, so its reference is the volume average of the truth over each cell (exact restriction) \\
Sequence of record sets & 64 test cases; 8,192 fixed query cells, half of them near the walls; 4,096 fixed anchors;
sequence of record resolutions and gradings against a uniform $256\times256$ reference per wall; unweighted $L^2$ over
the queries \\
\bottomrule
\end{tabular}
\end{table}

\section{Proofs and training-free checks}
\label{app:theory}

This section gives short proofs or proof sketches of the propositions of Section~\ref{sec:theory}. Table~\ref{tab:propositions}
summarises the statements, their relation to known results and the experiments that test the trained models.
Section~\ref{supp:checks} gives the training-free checks behind them (Table~\ref{tab:theory-checks}).

\begin{sidewaystable}   %
\centering
\footnotesize
\setlength{\tabcolsep}{4pt}
\caption{The statements of Section~\ref{sec:theory}, with their content, their relation to known results and the experiment that
tests each of them. The proof sketches follow, and Table~\ref{tab:theory-checks} lists the training-free checks.}
\label{tab:propositions}
\label{tab:hypotheses}
\begin{tabular}{P{0.10\linewidth}P{0.37\linewidth}P{0.17\linewidth}P{0.30\linewidth}}
\toprule
 & Content & Relation to known results & Tested by \\
\midrule
Lemma~\ref{lem:decomposition} & the error of a model trained on samples, under any deployment, is at most its class and
training error plus a state and a boundary term; C1--C4 remove the terms & elementary, new use & sampling gap against
each method trained on the whole mesh; drift across deployments (Section~\ref{sec:test-c3}) \\
Prop.~\ref{prop:sampling} & sampled training is consistent with whole-mesh training if the model is sample-invariant
and the weights are design-consistent; AGS keeps error and drift low across deployments &
classical estimator, new use & deployments, the anchor count and the sampling gap on all three data sets
(Section~\ref{sec:test-c3}) \\
Remark~\ref{prop:jump} & a server is a jump condition; its load enters additively & classical condition, new use &
response to a changed rack power (Section~\ref{sec:dcts-response}) \\
Prop.~\ref{prop:visibility} & hidden patches set an error floor; reading the boundary only on sampled nodes misses whole patches, whereas BQA reads
every patch & classical argument, new computation & patch coverage of the samplers and models that read the boundary only on sampled nodes
(Section~\ref{sec:test-c1}); backbones with and without BQA (Section~\ref{sec:test-hosts}) \\
Prop.~\ref{prop:drift} & a global state computed from the sample drifts with the design & ratio-estimator limit, new
use; first order new & drift against the covered share, regions and scattered passes (Section~\ref{sec:test-c3}) \\
Prop.~\ref{prop:count} & count weighting errs by a covariance and drifts with grading; area weighting converges at
second order & identity, new use & sequence of record sets; trained models on the reverse-graded mesh, and SCANO
trained with count weights (Section~\ref{sec:test-c2}) \\
Prop.~\ref{prop:code} & a global code of any width below the data dimension caps the response & classical &
global modulation against the floor of its code width, on both benchmarks (Section~\ref{sec:test-c4}) \\
Prop.~\ref{prop:unmix} & pooled adjacent patches cannot be unmixed & exact for one layer, heuristic beyond &
training-free check \\
Prop.~\ref{prop:records} & face records with area as the measure; patch records are the far-field monopole; the type
decides between average and sum; records need types & classical (i), new (ii), (iv), corollary (v) & training-free
checks; the measure test on trained models (Section~\ref{sec:test-c2}); the type-switch interventions of the
controlled benchmark, with SCANO trained without types (Section~\ref{sec:test-c1}) \\
Prop.~\ref{prop:slices} & slice tokens are copy-invariant; their readouts are a convex mixture of slice measures and an
area-additive sum & new & verified to round-off; training-free check \\
\bottomrule
\end{tabular}
\end{sidewaystable}

\subsection{Proof of Proposition~\ref{prop:sampling}}
\label{app:prop-c}

(i) For a sampling design, $\mathbb{E}\,\mathbf{1}\{x\in S\}=\pi_x$. Hence
$\mathbb{E}_S\mathcal{L}_S(\theta)=\sum_x (w_x/\pi_x)\,\mathbb{E}\bigl[\mathbf{1}\{x\in
S\}\,\ell(\hat u_\theta(x;S),u(x))\bigr]$. Under sample invariance, the prediction on the event $x\in S$ is $\hat u_\theta(x;\Omega_h)$. Each expectation therefore
equals $\pi_x\,\ell(\hat u_\theta(x;\Omega_h),u(x))$, and the sum is $\mathcal{L}(\theta)$. Points with $\pi_x=0$
contribute to neither side when $w_x\ell_x=0$ there. If the invariance holds on a neighbourhood of $\theta$, the two
objectives coincide as functions there. The gradient then commutes with the finite expectation. The H\'ajek form divides
by $\sum_{x\in S}w_x/\pi_x$ instead of its expectation. As a ratio estimator, it is consistent with a bias of order
$1/|S|$.

(ii) Without invariance the same computation gives
\[
  \sum_x w_x\,\mathbb{E}\bigl[\ell(\hat u_\theta(x;S),u(x))\mid x\in S\bigr],
\]
whatever the length of training. Suppose a global state $s(S)$ computed from the sample has a design-dependent limit
$s(\Pi)$. Suppose this limit differs from the whole-mesh state $s(\Omega_h)$. Then every supervised point is predicted
with the wrong state, and the gap does not shrink with the sample. If the state is a design-consistent ratio estimate,
$s(S)\to s(\Omega_h)$. The gap is then the ratio-estimator error, of order $1/|S|$. Proposition~\ref{prop:drift} gives
the limit.

(iii) Given anchors $A$ drawn independently of $S$, $\hat u_\theta(x;S,A)=\hat u_\theta(x;A)$ for every $S\ni x$. Part
(i) therefore holds conditionally on $A$. The anchor state is a H\'ajek estimate of the whole-mesh state. For $N_a$
anchors drawn without replacement, its bias is of order $1/N_a$. Its variance is of order $1/N_a-1/|\Omega_h|$. We
expand the loss to second order in the state (the delta method). This gives an objective gap of order $1/N_a$ between
training with random anchors and deployment with the whole-mesh cache.

(iv) Consider a design of nearly fixed size. The plain mean over the sample has expectation close to
$\sum_x\pi_x\ell_x\big/\sum_x\pi_x$. This is the whole-mesh objective with point weights $\pi_x$ instead of $w_x$. It is
the supervision measure studied in~\cite{m3}. It says nothing about whether the prediction at a supervised point is the
whole-mesh prediction. More generally, point weights $\varsigma_x$ give $\sum_x\varsigma_x\pi_x\,\mathbb{E}[\ell(\hat
u_\theta(x;S),u(x))\mid x\in S]$. Without sample invariance, this matches the whole-mesh loss for every $\theta$ only in
the degenerate case of Section~\ref{supp:scope}.

\emph{Check.} We use Poisson sampling on a graded one-dimensional mesh, with a quadratic loss in a global state. The
invariant model with Horvitz--Thompson weights is unbiased. A sample state by count or volume without $1/\pi_x$ leaves a
gap of order one. A design-consistent state leaves a gap of order $1/|S|$. Decoupled anchors leave a gap of order
$1/N_a$ over a 64-fold range of $N_a$. The plain mean targets the $\pi$-weighted objective.

\subsection{The other propositions}
\label{app:other-proofs}
\label{app:records}
\label{app:slices}

The derivations and proofs of the other statements of Section~\ref{sec:theory} are in Sections~\ref{supp:proofs}
and~\ref{supp:records}.

\section{Training-free checks}
\label{supp:checks}

Table~\ref{tab:theory-checks} lists the checks behind the statements of Section~\ref{sec:theory}. Each check evaluates a
statement on an exact kernel, on the discrete operator of the controlled benchmark or on the samplers. No trained model
is involved. Each check runs within minutes on a CPU. The derivations follow in
Sections~\ref{app:measure}--\ref{supp:records}.

\begin{table}[t]
\centering
\footnotesize
\setlength{\tabcolsep}{4pt}
\caption{Training-free checks of the propositions of Section~\ref{sec:theory} (column Prop.: the proposition, the
measure representation~\eqref{eq:measure}, or a part of Propositions~\ref{prop:records} and~\ref{prop:slices}).
Exponents are fitted over the far field; ``round-off'' means agreement to double-precision rounding. The hidden-patch
row uses one DCTS room, with its faces sampled uniformly at random.}
\label{tab:theory-checks}
\begin{tabular}{P{0.17\linewidth}P{0.11\linewidth}P{0.66\linewidth}}
\toprule
Check & Prop. & Outcome \\
\midrule
measure representation & \eqref{eq:measure}, \ref{prop:records}(v) & M-matrix with non-negative weights; Dirichlet and Robin weights sum to one with a Neumann patch present; Monte Carlo with drift $-\mathbf{b}$ matches the finite-volume weights, $+\mathbf{b}$ does not; a type switch changes the ratio of two exit probabilities by a factor between 0.30 and 1.15 over the domain \\
visibility bound & \ref{prop:visibility} & Bayes risk over the bound: 1.00 for independent constant data, 1.33 when the visible faces of a patch do not determine its data; 0.58 with correlated data (the bound no longer applies) \\
sampled objective & \ref{prop:sampling}, \ref{prop:drift} & invariant model with Horvitz--Thompson weights unbiased (within 1.1 standard errors); coupled state by count or by volume: gap of order one; anchors: gap times $N_a$ (finite-population corrected) between 0.065 and 0.090 for $N_a$ from 32 to 2,048 \\
global-code floor & \ref{prop:code} & worst error at the antipodal coincidence 0.93 against $\sigma_{d+1}r=0.35$; Eckart--Young bound attained \\
count against area & \ref{prop:count} & covariance identity to round-off; with the exact Poisson kernel, for smooth data the area sum is exact to round-off at distance 0.5 from the wall and in error below about half a record spacing; for piecewise-constant data it is in error already at 0.5 and of order one at the wall \\
pooled patches & \ref{prop:unmix} & $\sigma_{\min}(\Psi)\propto r_{pq}$ (fitted slope 0.99), always below $L_\psi r_{pq}/\sqrt2$ \\
far-field moments & \ref{prop:records}(i) & far-field exponent of the relative error $-2.00$ for the monopole and $-4.00$ with second moments; face records converge as $n_\Gamma^{-2}$ \\
patch-record bias & \ref{prop:records}(ii) & far-field bias equal to $|\Gamma|/|\mathbf{A}|-1$ for $|\Gamma|\,\bar{\mathbf{n}}$; residual exponent near $-1$ with the area vector and near $-2$ with the first-moment tensor \\
record copies and grading & \ref{prop:count}, \ref{prop:records}(iii) & copy invariance of area weighting to round-off; area weighting of second order (fitted 2.00); count weighting at the predicted plateau \\
two readouts & \ref{prop:records}(iv), (v) & Dirichlet weights sum to one to round-off; Neumann mass additive and growing with the patch; its spread over the cells exceeds a factor 100, whereas a normalised readout needs it constant; the two-path readout is exact to round-off \\
hidden-patch count & \ref{prop:visibility} & hidden patches at $\rho=0.05$: predicted 65.0 (standard deviation 4.7) of 373 non-empty patches, observed 64 and 66 in two draws; at $\rho=0.3$: predicted 14.3, observed 14 and 12 \\
\bottomrule
\end{tabular}
\end{table}

\section{Derivations of the measure representation and of the propositions of Section~\ref{sec:theory}}
\label{supp:proofs}

\subsection{The measure representation and Remark~\ref{prop:jump}}
\label{app:measure}

Let $X_t$ be the reflected diffusion with generator $\kappa\Delta-\mathbf{b}\cdot\nabla$,
$\mathrm{d}X=-\mathbf{b}\,\mathrm{d}t+\sqrt{2\kappa}\,\mathrm{d}W+\mathbf{n}_{\mathrm{in}}\,\mathrm{d}L$, reflected on
the insulated walls and on the $N$ and $R$ patches ($L$ the boundary local time), stopped at the first hitting time
$t_D$ of $D$, and killed on $R$ with the additive functional $\mathrm{d}\Lambda=(\alpha/\kappa)\,\mathrm{d}L^R$ for
the Robin condition $-\kappa\partial_n u=\alpha(u-u_\infty)$. The Feynman--Kac formula gives
\begin{equation*}
  u(x)=\mathbb{E}_x\bigl[e^{-\Lambda_{t_D}}u_D(X_{t_D})\bigr]
      +\mathbb{E}_x\Bigl[\int_0^{t_D} e^{-\Lambda_t}\tfrac{\alpha}{\kappa}u_\infty\,\mathrm{d}L^R_t\Bigr]
      +\mathbb{E}_x\Bigl[\int_0^{t_D} e^{-\Lambda_t}\tfrac{q}{\kappa}\,\mathrm{d}L^N_t\Bigr],
\end{equation*}
which is \eqref{eq:measure}. Its three terms are the discounted exit law on $D$, the killing law on $R$ and the
discounted expected local time on $N$. With $u_D\equiv1$, $u_\infty\equiv1$ and $q\equiv0$, the solution is the constant
one. The exit and killing measures therefore have total mass one at every $x$, with or without Neumann patches. The
local-time part is not normalised. The representation needs four assumptions. The diffusivity $\kappa$ is constant;
for a varying isotropic $\kappa(x)$, the drift becomes $\nabla\kappa-\mathbf{b}$, the diffusion coefficient
$\sqrt{2\kappa(x)}$, and the reflection stays normal. The drift $\mathbf{b}$ is bounded and Lipschitz. The set $D\cup
R$ has positive measure. The domain is bounded and Lipschitz~\cite{freidlin,papanicolaou,basshsu}. The
conditions $\nabla\cdot\mathbf{b}=0$ in the domain and $\mathbf{b}\cdot\mathbf{n}=0$ on walls are not needed for the
representation itself. They identify the operator with its conservative form and the Neumann data with the total flux,
as in the finite-volume scheme. For boundary data that jump at the junctions of patches, the solution is not smooth
there. The representation then holds in the weak sense. Bass and Hsu~\cite{basshsu} give it for reflecting Brownian motion on
Lipschitz domains. The bounded drift $-\mathbf{b}$ enters by a Girsanov change of measure. The walk thus runs
upstream, and the measure of a patch at $x$ is largest downstream of it. Consider now the discrete controlled benchmark.
Its finite-volume operator has cell P\'eclet numbers well below two and discretely divergence-free face fluxes. The same
Dirichlet and Robin coefficient appears on the diagonal and in the right-hand side. The operator is therefore a
non-singular M-matrix. Its weights $\mathcal{W}=A_h^{-1}B_h$ are non-negative. Since it reproduces constants, the
Dirichlet and Robin weights sum to one in every cell.

\emph{Checks} (measure representation, Table~\ref{tab:theory-checks}, and Proposition~\ref{prop:records}(iv)). We use a
graded two-dimensional test geometry with a Neumann patch. The operator is an M-matrix. All weights are non-negative.
The Dirichlet and Robin weights sum to one. A Monte Carlo simulation of the walk with drift $-\mathbf{b}$ reproduces the
finite-volume weights. With drift $+\mathbf{b}$, it does not.

Remark~\ref{prop:jump} follows in the same way. Consider the temperature equation with the flow frozen. On the exhaust
$\Gamma_s$, the server condition reads $T(x)=\int_{\Gamma^{\mathrm{in}}_s}T\,\mathrm{d}\nu_s+\Delta T_s$ for
$x\in\Gamma_s$. Here $\nu_s$ is the normalised area measure of the server's inlet. In the solver, the exhaust carries a
uniform fixed value. This value is the area-weighted mean temperature of the paired inlet patch plus $\Delta T_s$. The
condition is the pure-jump case of the general boundary condition of Wentzell~\cite{ventcel}. A path that hits
$\Gamma_s$ is restarted at a point of the inlet drawn from $\nu_s$. It collects $\Delta T_s$ on the way. Hence
$T(x)=(\text{exit and killing terms})+\sum_s\Delta T_s\,\mathbb{E}_x[n_s]$, where $n_s$ is the number of passages
through $\Gamma_s$ before absorption. Since $\nu_s$ has mass one, the exit and killing measures still sum to one. The
exhaust carries no exit mass. $\Delta T_s$ enters with the unnormalised weight $\mathbb{E}_x[n_s]$. Mass one also
requires that every path is eventually absorbed at a supply of the cooling units. The existence theorem we cite does not
cover a jump measure spread uniformly over an inlet. For this case we rely on the well-posedness of the CFD model.

\subsection{Proposition~\ref{prop:visibility}}
\label{app:prop-a}

Condition on the sample $S$ and on the data $g_V$ of the visible patches. The input, and hence $\hat u$, is then fixed. We
write $u=u_V+\sum_{p\in H(S)}K_pg_p$. The hidden data are independent of the conditioning and of each other. With
$\bar g_p$ the mean of $g_p$,
\begin{equation*}
  \mathbb{E}\bigl[\|\hat u-u\|_Q^2\,\big|\,S,g_V\bigr]
=\Bigl\|\hat u-u_V-\sum_{p\in H(S)}K_p\bar g_p\Bigr\|_Q^2+\sum_{p\in
H(S)}\operatorname{tr}\bigl(K_p\Sigma_pK_p^{*}\bigr)\big|_Q ,
\end{equation*}
since the cross terms of centred, independent hidden data vanish. Dropping the first term and taking the expectation
gives \eqref{eq:visibility}. Equality needs the visible data to be determined by the visible records. Sometimes the
visible faces of a patch do not determine its data, for example a linear profile observed on one face. The bound is then
strict. Patch data may also be correlated. The input may also contain a quantity that depends on the hidden data, such
as the first-iteration field of DCTS. In both cases the hidden part has less variance given the input. The formula is
then no longer a bound. Consider now the two-point form. If $p\in H(S)$, the inputs under $g$ and $g'$ coincide, so $\hat u$
is the same. The triangle inequality gives $\|\hat u-u_g\|_Q+\|\hat u-u_{g'}\|_Q\ge\|K_p(g_p-g'_p)\|_Q$. Averaging over the two
data sets and over $S$ gives the bound in any norm. Dividing by $\max(\|u_g\|_Q,\|u_{g'}\|_Q)$ gives its relative form.
Suppose faces are included independently with probability $\rho$. A uniform random subset of the faces does this
approximately. A patch with $n_p$ faces is then hidden with probability $(1-\rho)^{n_p}$. The expected number of hidden
patches is $\sum_p(1-\rho)^{n_p}$. The training sampler picks the points of a case along an arithmetic progression of a
fixed permutation. Its inclusions are therefore not independent and do not follow this law exactly.

\emph{Checks} (training-free checks on exact kernels and on the controlled benchmark). With independent constant data,
the Bayes risk meets the bound. When the visible faces of a patch do not determine its data, the risk lies above it.
With correlated data or data-dependent designs, it falls below the formula, as stated. The bound on the controlled
benchmark is computed from the exact response basis. The hidden-patch count on one industrial case is compared with the
prediction in Table~\ref{tab:theory-checks}.

\subsection{Proposition~\ref{prop:count}}
\label{app:prop-b}

With $\lambda_x=\log K(x,\cdot)$, the area-weighted aggregation is $\sum_ya_yK(x,y)g_y\big/\sum_ya_yK(x,y)$. Its
numerator and denominator are Riemann sums of $\int K(x,\cdot)g\,\mathrm{d}a$ and $\int K(x,\cdot)\,\mathrm{d}a=1$. The
ratio therefore converges at the order of the rule. For smooth data on a regular family of faces, this is second order.
Without $a_y$, a record density $\varrho$ turns the sums into $\int K(x,\cdot)g\varrho\,\mathrm{d}a$ and $\int
K(x,\cdot)\varrho\,\mathrm{d}a$. The limit is then $\int g\varrho\,\mathrm{d}\mu_x\big/\int\varrho\,\mathrm{d}\mu_x$.
Subtracting $\int g\,\mathrm{d}\mu_x$ gives
$(\mathbb{E}_{\mu_x}[g\varrho]-\mathbb{E}_{\mu_x}[g]\,\mathbb{E}_{\mu_x}[\varrho])/\mathbb{E}_{\mu_x}[\varrho]$. This is
\eqref{eq:count-error}. The measure $\mu_x$ is the exact boundary measure only if the learned logits equal its log
density. Closer to the wall than the record spacing, $K(x,\cdot)$ varies within one face. The Riemann sum is then in
error even with the exact kernel. For piecewise data, the error is of order one. This is a nearly singular quadrature.

\emph{Checks} (count against area and Proposition~\ref{prop:records}(iii), Table~\ref{tab:theory-checks}, and the
sequence of record sets on the controlled benchmark). The identity holds to round-off on a graded two-dimensional test
geometry with Dirichlet patches. There it also gives the spatial distribution of the count error. With the exact Poisson
kernel, the area sum is accurate to round-off away from the wall. It is visibly in error at distances well below the
record spacing. Part~(iii) of Proposition~\ref{prop:records} and the sequence of record sets on the controlled benchmark
show second-order convergence for area weighting. They also show the predicted plateau for count weighting.

\subsection{Proposition~\ref{prop:drift}}
\label{app:prop-g}

The numerator and denominator of $s_j(S)$ have expectations $\sum_x\pi_xc(x)\omega_jf$ and $\sum_x\pi_xc(x)\omega_j$.
The state therefore tends to their ratio, with a ratio-estimator bias of order $1/|S|$. Consider two designs with
induced measures $\varpi_x=\pi_xc(x)$ and $\varpi'_x=\pi'_xc(x)$. Then
$s_j(\varpi')-s_j(\varpi)=\sum_x(\varpi'_x-\varpi_x)\omega_j(x)\bigl(f(x)-s_j(\varpi)\bigr)\big/\sum_x\varpi_x\omega_j(x)
+O(\|\varpi'-\varpi\|^2)$. This is linear in the difference of the measures. The drift of the prediction follows by the
chain rule. With $c(x)\propto w_x/\pi_x$, the measure is $w_x$ for every design, and the state does not move.
Conversely, suppose the limit equals the whole-mesh state for every feature $f$. Then the normalised measures
$\varpi_x\omega_j(x)/\sum_{x'}\varpi_{x'}\omega_j(x')$ and $w_x\omega_j(x)/\sum_{x'}w_{x'}\omega_j(x')$ coincide. Hence
$\varpi_x\propto w_x$ wherever $\omega_j>0$. The soft slice weights are positive everywhere, so $\pi_xc(x)\propto w_x$.
Anchors drawn independently of the sample give a H\'ajek estimate. Its fluctuation scales as
$(1/N_a-1/|\Omega_h|)^{1/2}$. Its bias is of order $1/N_a$.

\subsection{Propositions~\ref{prop:code} and~\ref{prop:unmix}}
\label{app:prop-de}

\begin{proposition}[unmixing after pooling]
\label{prop:unmix}
In a one-layer model linear in the data and modulated linearly by a code of width $d$, two patches $p$, $q$ at distance
$r_{pq}$ can be recovered from the code only through a left inverse of norm at least $\sqrt2/(L_\psi r_{pq})$. Here
$L_\psi$ is the Lipschitz constant of the record feature map $y\mapsto\tilde\psi(y)$, and the columns of $\Psi$ are the
features $\tilde\psi(y_p)$ of the patches. Recovering adjacent small patches pooled into one component is therefore
ill-conditioned, and their errors are coupled, as for the rows of server exhausts in a room.
\end{proposition}
The statement is exact for the linearised one-layer model and a heuristic beyond it.

\emph{Proposition~\ref{prop:code}, worst case.} Restrict the data to the sphere of radius $r$ around $g_0$ in the span
of the first $d+1$ right singular vectors of $\mathcal{G}'$, the derivative of the solution operator $\mathcal{G}$ at
$g_0$. The encoder maps this $d$-sphere continuously into $\mathbb{R}^d$. By the Borsuk--Ulam theorem, there is an
$\eta$ with $E(g_0+\eta)=E(g_0-\eta)$. The model therefore gives the same prediction $\hat u$ for both. Then
$2\max_\pm\|\hat
u-\mathcal{G}(g_0\pm\eta)\|_Q\ge\|\mathcal{G}(g_0+\eta)-\mathcal{G}(g_0-\eta)\|_Q=\|2\mathcal{G}'\eta+O(r^3)\|_Q\ge
2\sigma_{d+1}r-O(r^3)$. The even-order terms cancel in the symmetric difference. The remainder $O(r^3)$ assumes a solution operator that is three
times differentiable, and the singular values assume Hilbert norms on the data and the outputs. \emph{Proposition~\ref{prop:code}, mean
case.} A linearised model has a Jacobian $J$ of rank at most $d$. By Eckart--Young--Mirsky,
$\|J-\mathcal{G}'\|_F^2\ge\sum_{i>d}\sigma_i^2$. For isotropic perturbations $\eta$ with covariance $t^2I$ on the discretised data,
$\mathbb{E}\|(J-\mathcal{G}')\eta\|^2=t^2\|J-\mathcal{G}'\|_F^2$. For an arbitrary continuous encoder, a very steep encoder can make the mean
error over a finite-amplitude distribution small. Only the worst-case bound is a hard limit. Bounds on the mean error
under a Lipschitz constraint exist~\cite{cohenstable,lipschitzwidths}. Sum-pooled set representations need a latent
dimension at least the set size, for a related reason~\cite{wagstaff}.

\emph{Proposition~\ref{prop:unmix}.} Take a code $\zeta=\sum_p\xi_p\tilde\psi(y_p)$ and linear modulation. Then $\hat
u(x)=\sum_p\xi_p\tilde K(x,y_p)$ with the separable kernel of the proposition. Recovering the $\xi_p$ needs $\Psi$ of
full column rank. For two patches at distance $r_{pq}$, $\|\Psi(e_p-e_q)\|/\sqrt2\le L_\psi r_{pq}/\sqrt2$. Smooth features
have a numerical rank well below $d$. What counts is therefore the numerical rank of $\Psi$, not $d$. Per-record
attention forms no code. Two nearby records receive weights whose ratio is the exponential of the difference of their
logits. They are therefore separated only if the logits change by order one over their distance.

\emph{Checks} (training-free checks on exact kernels and on the controlled benchmark). The antipodal coincidence is
found, and the worst error there exceeds $\sigma_{d+1}r$. The rank-$d$ Jacobian bound is attained. A steep continuous
encoder drives the mean error down, while the worst case stays above the bound. $\sigma_{\min}(\Psi)$ falls in
proportion to $r_{pq}$. The response spectrum and its tail energy on the controlled benchmark are computed from the
exact response basis.

\section{Proofs of the record and slice statements}
\label{supp:records}

\subsection{Proposition~\ref{prop:records}}
\label{supp:app-records}

\begin{proposition}[boundary records]
\label{prop:records}
(i) Under area quadrature a single record per planar patch is
the monopole term of the patch integral, with far-field error $(h_\Gamma/r)^2$ (and $(h_\Gamma/r)^4$ with second
moments for a centrally symmetric patch; $h_\Gamma$ the size of the patch, $r$ the distance), while records on
$n_\Gamma\times n_\Gamma$ faces converge as $n_\Gamma^{-2}$. (ii) For a non-planar patch, the area times the normalised
mean normal $\mathbf{A}/|\mathbf{A}|$ leaves a bias $|\Gamma|/|\mathbf{A}|-1$ that the area vector $\mathbf{A}$
removes. (iii) Copying a record into pieces that share its area leaves area weighting exactly unchanged
while count weights drift. (iv) The weights of Dirichlet data form a probability measure in every cell, while a flux
enters additively. A readout whose weights sum to one over the records therefore cannot represent both, and it must be
paired with an area-weighted unnormalised readout. (v) Records without
types map problems with different solutions to the same input.
\end{proposition}
Part~(i) is classical. Parts (ii) and (iv) are new for boundary records. Part (iii) checks Proposition~\ref{prop:count}.
Part (v) is a corollary of the measure representation. Training-free checks confirm all five.

\emph{(i).} Expand the kernel about the centroid $\bar y$ of a planar patch $\Gamma$. Since the first moment about $\bar
y$ vanishes, $\int_\Gamma K(x,y)\,\mathrm{d}a(y)=|\Gamma|\,K(x,\bar y)+\tfrac12\nabla_y^2K(x,\bar
y):\mathbf{M}_2+\dots$. The relative error of the monopole term is of order $(h_\Gamma/r)^2$. With the second moment
$\mathbf{M}_2$, it is of order $(h_\Gamma/r)^4$; the third moment vanishes for a centrally symmetric patch. Near the
patch, the expansion fails and the error is of order one. Face records with area weights are a midpoint rule on $n_\Gamma\times n_\Gamma$ faces. Their error is of order
$n_\Gamma^{-2}$. These orders assume derivatives of the kernel whose $n$-th order scales as $K/r^n$. With drift, $r$ is replaced by the
smaller of $r$ and the P\'eclet length $2\kappa/|\mathbf{b}|$. \emph{(ii).} Take the double-layer kernel
$\nabla_y\Phi(x,y)\cdot\mathbf{n}(y)$, where $\Phi$ is the fundamental solution. Then
$\int_\Gamma\nabla_y\Phi\cdot\mathbf{n}\,\mathrm{d}a=\nabla_y\Phi(x,\bar y)\cdot\mathbf{A}+\nabla^2_y\Phi(x,\bar
y):\mathbf{M}_1+\dots$. Here $\mathbf{A}=\int_\Gamma\mathbf{n}\,\mathrm{d}a$ is the area vector and
$\mathbf{M}_1=\int_\Gamma\mathbf{n}(y-\bar y)^{\mathsf T}\mathrm{d}a$ the first-moment tensor. Replacing $\mathbf{A}$ by
$|\Gamma|\,\mathbf{A}/|\mathbf{A}|$, the area times the normalised mean normal, scales the leading term by
$|\Gamma|/|\mathbf{A}|$. This bias does not decay. With
$\mathbf{A}$, the remainder is the $\mathbf{M}_1$ term, of relative order $h_\Gamma/r$. Adding $\mathbf{M}_1$ leaves
$(h_\Gamma/r)^2$. \emph{(iii).} Replacing a record of area $a$ by $n$ copies of area $a/n$ leaves every area-weighted
sum unchanged, term by term. Refinement is a midpoint rule, of second order. Under count weights, the share of a half whose faces are
split $n$ times grows as $4^n/(4^n+1)$ for surface faces in three dimensions, which split into four, and as $2^n/(2^n+1)$ for
boundary segments in two dimensions, which split into two. A patch-mean record ignores the variation of the kernel across the patch. This
error does not depend on the face size. \emph{(iv).} The M-matrix argument above gives the first statement. Linearity in
the flux gives the additive Neumann response. Suppose a normalised readout $\hat
u(x)=\sum_i\nu_i(x)\,\chi_{\tau(i)}\,g_i$ were exact for all data, with $\nu_i\ge0$, $\sum_i\nu_i=1$ and weights
independent of the values. Shifting the Dirichlet data requires $\sum_{i\in D}\nu_i\chi_D=1$. The Neumann response
requires $\sum_{j\in N}\nu_j\chi_N=m(x)$, the Neumann mass at $x$. Since the weights sum to one,
$1/\chi_D+m(x)/\chi_N=1$ for every $x$. Then $m$ would be constant in $x$, which it is not. The argument uses weights that sum to one over the records. A
readout with an extra token of value zero escapes it, at the cost of a sink weight that varies with $x$. A normalised
readout over
the Dirichlet records together with the unnormalised sum $\sum_{j\in N}a_je^{\lambda_j(x)}q_j$, with
$\lambda_j=\log(\mathcal{W}_{N}[x,j]/a_j)$, is exact. Its logits do not depend on the patch size. \emph{(v).} Two
problems with the same values and different types give the same untyped input. By the two-point argument of
\ref{app:prop-a}, $\max\|\hat u-u\|\ge\tfrac12\|u-u'\|>0$.

\emph{Checks} (Table~\ref{tab:theory-checks}): all confirmed.

\subsection{Proposition~\ref{prop:slices}}
\label{supp:app-slices}

\begin{proposition}[slice tokens]
\label{prop:slices}
(i) With the area in the slice means and the readouts, both readouts are exactly invariant to copying a record into
pieces that share its area; hard assignment by patch and one slice per face are the two limits; scaling the constant of
the denominator with the total area, $M_g+\epsilon M_{\mathrm{tot}}$, also makes the tokens invariant to the unit of
area. (ii) In one layer and attention head with a linear value map and $\epsilon=0$, the normalised readout is a convex
combination of the $G$ slice measures and the unnormalised readout a non-negative combination additive in area; for
$\epsilon>0$, the weights of a slice sum to $M_g/(M_g+\epsilon M_{\mathrm{tot}})$.
\end{proposition}
Part (i) holds exactly. Part (ii) holds exactly for $\epsilon=0$, and with $\epsilon=10^{-8}$ the deficit matters only
for a nearly empty slice. The checks verify both parts to round-off.
Records that share a slice fall under Proposition~\ref{prop:unmix}.

\emph{(i).} The slice weights $\omega_g(y)$ depend on the lifted record $h_y$ only. Consider $n$ copies of a record,
each with area $a_y/n$. They have the same weights. In total they contribute $a_y\omega_g(y)$ to the slice mass $M_g$
and $a_y\omega_g(y)h_y$ to the numerator of $z_g$. The total area $M_{\mathrm{tot}}$ is unchanged. With the denominator
$M_g$ or $M_g+\epsilon M_{\mathrm{tot}}$, the tokens and masses are therefore unchanged. Both readouts depend on the
records only through them. Copying changes no mass, so a denominator $M_g+\epsilon$ with an absolute
$\epsilon$ is copy-invariant too. Rescaling every area by $\varkappa>0$ multiplies $M_g$ and $M_{\mathrm{tot}}$ by
$\varkappa$. The tokens are then unchanged with $M_g+\epsilon M_{\mathrm{tot}}$. With $M_g+\epsilon$, they change by a
relative amount of order $\epsilon/M_g$ that depends on the unit of area. Consider hard assignment by
patch, $\omega_g(y)=\mathbf{1}\{y\in\Gamma_g\}$. It gives $M_g=|\Gamma_g|$, and $z_g$ is the area mean of $h_y$ over the
patch. With one slice per face, $M_g=a_y$ and $z_g=h_y$. The normalised readout then becomes
$\sum_y\operatorname{softmax}_y(\lambda_y+\log a_y)\,v(h_y)$. This is the area-weighted attention of~\cite{neo}.

\emph{(ii).} Take a linear value map $v$ and the denominator $M_g$. Then $v(z_g)=\sum_y(a_y\omega_g(y)/M_g)\,v(h_y)$.
With the denominator $M_g+\epsilon M_{\mathrm{tot}}$, the weights of a slice sum to $M_g/(M_g+\epsilon
M_{\mathrm{tot}})$. This is a mass deficit of relative order $\epsilon M_{\mathrm{tot}}/M_g$. The normalised readout
weights slice $g$ by $\upsilon_g=M_ge^{\lambda_g}\big/\sum_{g'}M_{g'}e^{\lambda_{g'}}$. It therefore equals
$\sum_y\tilde\upsilon_yv(h_y)$ with
$\tilde\upsilon_y=a_y\sum_g\omega_g(y)e^{\lambda_g}\big/\sum_{g'}M_{g'}e^{\lambda_{g'}}\ge0$ and
$\sum_y\tilde\upsilon_y=\sum_g\upsilon_g=1$. This is a convex combination of the $G$ slice measures $a_y\omega_g(y)/M_g$. In
the limit of one slice per face, it is the per-record area-weighted softmax. The unnormalised readout equals
$\sum_y(a_y/M_{\mathrm{ref}})\bigl(\sum_g\omega_g(y)e^{\lambda^N_g}\bigr) v^N(h_y)$. It is non-negative and additive in
area.

\section{Scope of the theoretical statements and details of the design}
\label{supp:scope}

\subsection{What is classical and what is new}
The propositions of Section~\ref{sec:theory} apply classical tools to sampled training. Table~\ref{tab:propositions}
records which part of each statement is known and which part is new. Four ingredients are classical, and we cite them as
such. They are the Horvitz--Thompson identity behind Proposition~\ref{prop:sampling}(i), the ratio-estimator limit
behind Proposition~\ref{prop:drift}, the Feynman--Kac representation behind Eq.~\eqref{eq:measure} and the width bounds
behind Proposition~\ref{prop:code}. The paper adds a reading of boundary conditioning under sampling through these
tools. This reading says which weights a boundary record should carry. It says which part of a model must not depend on
its sample. It also names the diagnostic that exposes each violation before or after training. The four conditions are
therefore design requirements derived from these statements, not new theorems. Section~\ref{sec:results} tests each of
them on trained models.

\subsection{\texorpdfstring{Proposition~\ref{prop:sampling}(iv) and Proposition~\ref{prop:drift} stated precisely}{Two propositions stated precisely}}
Part~(iv) of Proposition~\ref{prop:sampling} concerns a model that is not sample-invariant. For such a model, consider
point weights $\varsigma_x$ that depend only on the point and the design. In general, no such weights make
$\mathbb{E}_S\sum_{x\in S}\varsigma_x\,\ell(\hat u_\theta(x;S),u(x))$ equal to the whole-mesh loss~\eqref{eq:loss-mesh}
for every $\theta$. The reason is part~(ii). The expectation weights the conditional loss $\mathbb{E}[\ell(\hat
u_\theta(x;S),u(x))\mid x\in S]$. Its dependence on $\theta$ differs from that of $\ell(\hat
u_\theta(x;\Omega_h),u(x))$, whatever the weights. The exception is a degenerate model whose conditional loss is a
multiple of the whole-mesh loss that does not depend on $\theta$. Weights inversely proportional to $\pi_x$ and to that
multiple then work.
Proposition~\ref{prop:drift} contains two statements of different kind. The first is the limit of the ratio estimator
$s_j(S)$ under a growing design, the state with weights $\pi_xc(x)$. For a given design, this limit equals the
volume-weighted whole-mesh state for every feature if and only if $\pi_xc(x)\propto w_x$. The second is a first-order
expansion of the change of the state between two deployments. The expansion is in the difference of their measures
$\pi_xc(x)$. Part~(iii) of Proposition~\ref{prop:sampling} and Proposition~\ref{prop:drift} also refer to different
quantities. The gap between the expected training objective and the deployment objective is of order $1/N_a$. A loss
twice differentiable in the state responds to an unbiased state error only through its variance. The state itself
fluctuates with fresh anchors as $N_a^{-1/2}$. These rates hold for anchors drawn as single cells. For anchors grown as
subgraphs, the number of seeds replaces $N_a$.

\subsection{\texorpdfstring{Lemma~\ref{lem:decomposition} is an organising decomposition}{The decomposition lemma}}
The lemma follows from the triangle inequality. The Lipschitz constants of a trained network are not controlled a
priori. The role of the lemma is to name the four terms. Each term can then be isolated by a separate experiment, as the
paragraph after the lemma describes. The bounded-Lipschitz distance is taken between the weighted empirical measures
$\beta$ on the record space $\mathcal{Y}$ of position, normal, type and value. The metric is
$d_{\mathcal{Y}}((\mathbf{x},\mathbf{n},\tau,g),(\mathbf{x}',\mathbf{n}',\tau',g')) =
|\mathbf{x}-\mathbf{x}'|/L_\Omega+|\mathbf{n}-\mathbf{n}'|+|g-g'|/\sigma_g+\mathbf{1}\{\tau\neq\tau'\}$. Here $L_\Omega$
is the domain size and $\sigma_g$ the scale of the values. Two records of different type are thus always a unit apart.

\subsection{Where the representation holds and how it is used beyond}
Eq.~\eqref{eq:measure} assumes a linear operator with constant diffusivity. It assumes a divergence-free drift that is
tangential on walls and a Lipschitz domain. Section~\ref{app:measure} gives the conventions. The reflection is inward,
with local time normalised by the diffusion $\sqrt{2\kappa}$. The Robin condition is
$-\kappa\partial_nu=\alpha(u-u_\infty)$. The Neumann flux is $q=\kappa\partial_nu$ with the outward normal, so a
positive $q$ heats the domain. Junctions of patches of different types are allowed. The representation needs only
bounded, piecewise continuous boundary data on a Lipschitz boundary. On DCTS, three assumptions do not hold exactly.
First, the effective diffusivity $\kappa+\kappa_t$ varies in space, with the turbulent diffusivity $\kappa_t$. This replaces the drift by
$\nabla\kappa_{\mathrm{eff}}-\mathbf{U}$, with a normal reflection. Second, the supplies, returns and server faces are
inflow and outflow boundaries, not walls. A path that reaches an inflow face stops there. An inflow with prescribed
temperature therefore acts as a Dirichlet patch. A server exhaust acts as the jump of Remark~\ref{prop:jump}. Third, the
velocity itself depends on the boundary data, through buoyancy and the supply flow rates. The measures of a room
therefore change with its data. For the temperature under a given flow, the structure of Eq.~\eqref{eq:measure} is
preserved. For the coupled nonlinear problem, the four conditions serve as a design guide. Their value is established
empirically. Section~\ref{sec:dcts} uses them in this way.

\subsection{How hidden patches harm a model at whole-mesh deployment}
At whole-mesh deployment, every boundary face is in the input. The bound of Proposition~\ref{prop:visibility} therefore
applies to the training samples and to the regional deployments, not to the whole-mesh pass. Its effect on the whole
mesh arises through training. Consider a model that reads the boundary only as sampled nodes. It is fitted to samples in
which the patches that determine a point are often absent. The minimiser of the sampled objective therefore learns the
conditional mean of the solution given the visible patches. This weakens its dependence on every patch that is often
hidden. At deployment the patch is present, but the learned map underuses it. The visibility bound thus appears as a
shift between the input distributions of training and deployment. The response of such models to patches that were often
hidden measures this shift directly (Table~\ref{tab:r0s-main-resp}).

\subsection{Sample consistency of a message-passing backbone}
C3 requires that the prediction at a point not depend on the other points of its sample. For a backbone with message
passing, the receptive field of a supervised point must then lie inside its training sample. Otherwise the point sees a
truncated neighbourhood in training and the full neighbourhood at deployment. Section~\ref{sec:record-theory} removes
the truncation at deployment with a halo as deep as the message passing. In training, a halo of the same depth removes
it too. This halo enters the forward pass but not the loss. Its cost is the memory of the larger subgraph.
Section~\ref{supp:r7-halo} measures both its effect on the sampling gap and this cost.

\subsection{C4 as a requirement on the readout}
Proposition~\ref{prop:code} motivates C4 through the dimension of a global code. Figure~\ref{fig:predictions}, however, shows
that global modulation fails well above that floor. The limitation lies less in the width of the code than in
broadcasting one summary to every point. We therefore read C4 as a requirement on the readout. The readout of the
boundary must depend on the point. Its resolution budget must not compress the records below the scale of the patches.
Per-record attention on the benchmarks meets this exactly. With $G$ slice tokens, the budget is $G$
(Proposition~\ref{prop:unmix}). Section~\ref{supp:r7-design} examines this budget on the controlled benchmark.

\subsection{Details of the readouts}
The term $\epsilon M_{\mathrm{tot}}$ in the token mean~(Section~\ref{sec:slices}) keeps an empty slice from dividing by
zero. It scales with the total boundary area, so the tokens do not depend on the unit of area. Copy invariance
(Proposition~\ref{prop:slices}(i)) holds with or without this scaling. With $\epsilon=10^{-8}$, the term changes a token
by a relative amount $\epsilon M_{\mathrm{tot}}/M_g$, which matters only for a nearly empty slice.
Inside the logarithm of the normalised readout, the mass is bounded below by the smallest positive floating-point
number. An empty slice therefore receives a vanishing weight. The exponent of the unnormalised readout is clipped at 30.
Its output projection starts at zero. The path thus begins silent and grows only as training requires totals.
$M_{\mathrm{ref}}$ fixes the unit of area in which the masses are counted. It is set once, on the order of a room's
boundary area ($4{,}000\,\mathrm{m}^2$ on DCTS), and is not tuned. Copy invariance is exact. Splitting a face into
smaller faces also moves the centroids at which the records are read. The readouts then converge to the integral as the
faces are refined. They do not stay identical. This is the sense of the main text's statement that splitting refines the
quadrature. The tokens are computed from the records alone. The domain does not update them. The coupling between a
server's exhaust and its inlet enters through the records of both patches. It also enters through the domain field,
which every layer updates.

\subsection{The records of a DCTS room}
On every role, $g_y$ holds the five boundary values $(U_x,U_y,U_z,p',T)$ that the solver writes at its first iteration.
Every room starts from the same uniform initial field. Where a condition is imposed, these are the prescribed values.
Examples are the supply velocity and temperature of a cooling unit and the flow of a server. Where the value follows
from the interior, they are the first iterate. Examples are a zero-gradient wall temperature and a server exhaust
temperature equal to its inlet mean plus $\Delta T_s$. No interior field of any iteration enters the input. The weight
of an exhaust record is its face area times the exhaust speed divided by a fixed reference speed. It is the same for
every method that reads records. The weights therefore remain areas in unit and proportional to the mass flux through
the face. On DCTS, BQA replaces GNOT's own record attention, as on the controlled benchmark. On the public benchmark, it
sits beside it. Section~\ref{supp:r7-design} reports both arrangements on both benchmarks.

\subsection{The global-modulation baseline}
Global modulation lifts every record with the same encoder as BQA. It averages the lifted records by area into one
summary. It modulates every layer with this summary through FiLM or adaLN-Zero. Its strongest variant replaces the
average by an attention pooling with learned queries in every layer. This is a query-independent readout of the same
records that BQA reads at each point. It remains less accurate than BQA on the controlled benchmark
(Section~\ref{supp:r7-design}).

\subsection{Sampling, losses and metrics}
The public benchmark has about $1.7\times10^4$ points per case and fits in memory. Its sampling regimes emulate the
large-mesh setting to test the mechanism. Every model is trained with a loss that weights the sampled points equally.
The models trained on the whole mesh use the same loss over all points. The sampling gap thus compares objectives of the
same form. The metrics weight cells by volume on volumetric meshes. The gap is reported in that metric. Let $\mathcal{D}$ be a
deployment and $Q_{\mathcal{D}}$ its deployed cells, and let $\hat u_{\mathrm{ref}}$ be the model's own reference prediction. On the
two benchmarks, the drift is $\delta=\|\hat u_{\mathcal{D}}-\hat u_{\mathrm{ref}}\|_{w,Q_{\mathcal{D}}}/\|u\|_{w,Q_{\mathcal{D}}}$. On DCTS, it is $\|\hat
u_{\mathcal{D}}-\hat u_{\mathrm{ref}}\|_{w,Q_{\mathcal{D}}}/\|u-\bar u\|_{w,Q_{\mathcal{D}}}$, where $\bar u$ is the training mean. Drift and error thus
share one unit on every data set. The reference is the whole-mesh pass. For DCTS models whose state follows the sample,
it is each model's own deployment. On another mesh, the drift counts only the excess over the change of the reference solution.
``Seven backbones'' counts the six external backbones and the SCANO backbone.

\subsection{Symbols}
Table~\ref{tab:notation} lists every symbol. A few letters carry two related meanings, and the context separates them.
$\omega_g(y)$ assigns boundary records to boundary slices. $\omega_{l,j}(x)$ assigns domain points to domain slices.
$g_y$ is a record's value. The subscript $g$ indexes a slice. $\delta$ denotes the drift, and $\delta_{\phi_y}$ a
point mass at a record. $\vartheta$ is the slice temperature, distinct from the parameters $\theta$. $\|\cdot\|_Q$
restricts a norm to the cells of a region or sample $Q$.

\subsection{Units of the cost figures}
The dashed line of Figure~\ref{fig:bc-stats}a and Figure~\ref{fig:cost}a marks the memory that the device reports as
available. It is in GiB, like the axis. The medians of the stages in Table~\ref{tab:dcts-time} are taken over the rooms
separately. They therefore need not add up to the median of the totals.

\section{SCANO's components on seven backbones: every value}
\label{supp:hosts}

Table~\ref{tab:hosts} gives the values of Figure~\ref{fig:hosts-bars} (Section~\ref{sec:hosts-results}). It also
includes the combinations of backbone and data set that the figure does not draw.

\begin{table*}[t]
\centering
\scriptsize
\setlength{\tabcolsep}{2.6pt}
\caption{Effect of BQA and AGS on seven operator backbones and three data sets. The table gives the values of Figure~\ref{fig:hosts-bars} in percent ($100\times$ the relative $L^2$ error), averaged over the training seeds, on the whole mesh in a single pass and in a compact region that holds 2\% of the cells (Section~\ref{sec:setup-deploy}). Each backbone is evaluated in its original (native) form, with BQA, and with BQA + AGS. On the SCANO backbone, the last form is SCANO (Section~\ref{sec:hosts}). Bold marks the best value of a column. Underlining marks the best variant of each backbone. $^{\mathrm{f}}$Includes a run above the pre-registered validation threshold. --: not applicable or not run. AeroGTO and MeshGraphNets need subgraph connectivity and run on $k$-hop samples only. Their BQA variants are not run on DCTS (native AeroGTO on DCTS is in Table~\ref{tab:dcts-main}).}
\label{tab:hosts}
\begin{tabular}{llrrrrrrrrrr}
\toprule
 & & \multicolumn{4}{c}{Controlled benchmark} & \multicolumn{4}{c}{Public benchmark} & \multicolumn{2}{c}{DCTS} \\
\cmidrule(lr){3-6} \cmidrule(lr){7-10} \cmidrule(lr){11-12}
 & & \multicolumn{2}{c}{$k$-hop} & \multicolumn{2}{c}{uniform} & \multicolumn{2}{c}{76\%} & \multicolumn{2}{c}{5\%} & \multicolumn{2}{c}{} \\
\cmidrule(lr){3-4}\cmidrule(lr){5-6}\cmidrule(lr){7-8}\cmidrule(lr){9-10}\cmidrule(lr){11-12}
Backbone & Variant & whole & region & whole & region & whole & region & whole & region & whole & region \\
\midrule
Transolver & original & 80.0 & 272 & 77.8 & 253 & 46.8 & 166 & 80.9$^{\mathrm{f}}$ & 91.6$^{\mathrm{f}}$ & 55.4 & 134 \\
 & + BQA & 18.9 & 63.6 & \underline{17.7} & 56.8 & 3.06 & 47.1 & 4.48 & 19.4 & 43.1 & 94.1 \\
 & + BQA + AGS & \underline{18.0} & \underline{23.8} & 18.6 & \underline{23.9} & \underline{2.86} & \underline{4.01} & \underline{3.78} & \underline{5.53} & \underline{41.5} & \underline{52.0} \\
\midrule
AB-UPT & original & 30.9 & 288 & 22.1 & 284 & 3.35 & 118 & 26.5 & 82.8 & -- & -- \\
 & + BQA & 7.78 & 110 & \underline{\textbf{7.73}} & 102 & 1.84 & 38.4 & 3.48 & 21.3 & -- & -- \\
 & + BQA + AGS & \underline{\textbf{7.35}} & \underline{\textbf{11.6}} & 8.29 & \underline{\textbf{12.6}} & \underline{\textbf{1.71}} & \underline{2.98} & \underline{\textbf{2.01}} & \underline{\textbf{3.31}} & \textbf{37.3} & \textbf{46.1} \\
\midrule
AeroGTO & original & 64.8 & 259 & -- & -- & 13.0 & 174 & -- & -- & -- & -- \\
 & + BQA & \underline{16.6} & 150 & -- & -- & \underline{2.71} & 185 & -- & -- & -- & -- \\
 & + BQA + AGS & 17.2 & \underline{24.7} & -- & -- & 2.80 & \underline{4.30} & -- & -- & -- & -- \\
\midrule
GNOT & original & 26.4 & 70.0 & 26.8 & 65.7 & 9.04 & 53.0 & 10.1 & 32.0 & 51.8 & 90.9 \\
 & + BQA & \underline{14.8} & 93.7 & \underline{13.9} & 85.7 & \underline{5.81} & 33.7 & 9.82 & 27.1 & 47.7 & 94.4 \\
 & + BQA + AGS & 14.9 & \underline{19.5} & 15.0 & \underline{19.5} & 8.10 & \underline{11.9} & \underline{9.10} & \underline{13.9} & \underline{44.0} & \underline{53.9} \\
\midrule
GeoTransolver & original & 47.9 & 135 & 50.0 & 101 & 12.0 & 89.2 & 13.6 & 61.4 & 39.9 & 87.8 \\
 & + BQA & \underline{14.1} & 60.2 & \underline{14.3} & 50.8 & 2.65 & 20.5 & 3.48 & 10.5 & 41.1 & 85.0 \\
 & + BQA + AGS & 15.2 & \underline{20.8} & 14.7 & \underline{19.4} & \underline{2.54} & \underline{3.38} & \underline{3.45} & \underline{4.56} & \underline{39.5} & \underline{49.5} \\
\midrule
MeshGraphNets & original & 99.3$^{\mathrm{f}}$ & 119$^{\mathrm{f}}$ & -- & -- & 53.5 & 71.7 & -- & -- & -- & -- \\
 & + BQA & \underline{23.9} & \underline{33.2} & -- & -- & \underline{4.92} & \underline{7.37} & -- & -- & -- & -- \\
\midrule
SCANO backbone & + BQA & \underline{10.9} & 75.0 & \underline{12.9} & 67.0 & 2.08 & 29.9 & 3.21 & 15.8 & \underline{39.3} & 63.7 \\
 & + BQA + AGS & 11.1 & \underline{14.5} & 13.8 & \underline{18.9} & \underline{1.97} & \underline{\textbf{2.93}} & \underline{2.48} & \underline{3.89} & 39.5 & \underline{48.7} \\
\bottomrule
\end{tabular}
\end{table*}

\section{The controlled benchmark: drift and response, uniform samples, sparse sampling, the cost of sampling}
\label{supp:r0s}

Table~\ref{tab:r0s-main-resp} gives the region drift, the response errors and the size of the models of
Table~\ref{tab:r0s-main}. Figure~\ref{fig:drift-response} plots them. Table~\ref{tab:r0s-uniform} repeats the main
comparison for uniform random samples at the same ratio. Table~\ref{tab:r0s-uniform-resp} gives the drift and response
there. Table~\ref{tab:r0s-strong} repeats the main comparison at the sparse end of both families (ratio 0.02).
Table~\ref{tab:r0s-strong-resp} gives the drift and response there. Table~\ref{tab:r0s-tax} gives the sampling gap of
the main methods that were also trained on the whole mesh (Section~\ref{sec:test-c3}).

\begin{table*}[!tbp]
\centering
\scriptsize
\setlength{\tabcolsep}{3pt}
\caption{Controlled benchmark: drift, response and size of the models of Table~\ref{tab:r0s-main}. Values are averaged over the training seeds, per case and, for $E_{\mathrm{resp}}$, per intervention pair. $\pm$ is half the width of the 95\% case-bootstrap interval; for $E_{\mathrm{resp}}$, whole base cases are resampled. The region drift $\delta$ is measured in a compact region that holds the sampling ratio's share of the cells. It is the distance of the prediction there from the model's own whole-mesh prediction on the region's cells, relative to the reference. A model with AGS draws fresh anchors in the region, as in training. A message-passing model reads the halo of its receptive field. $E_{\mathrm{resp}}$ (Eq.~\eqref{eq:eresp}) is measured on one sample drawn as in training. It covers the intervention pairs whose true response exceeds five percent of the base solution. The columns give all pairs, the pairs on the smallest patches (tiny) and the far pairs, whose intervened patch has no face in the sample. $^{\ddagger}$Equal to 1 by construction. The model has no boundary input, or it reads no boundary record while the intervened patch lies outside the sample. Parameters are in thousands. Training families and marks as in Table~\ref{tab:r0s-main}.}
\label{tab:r0s-main-resp}
\begin{tabular}{lccccc}
\toprule
 & & \multicolumn{3}{c}{$E_{\mathrm{resp}}$ on a sample} & \\
\cmidrule(lr){3-5}
Method & Region drift $\delta$ & All & Tiny & Far & Parameters ($10^3$) \\
\midrule
\multicolumn{6}{l}{\textbf{$k$-hop samples, ratio 0.12}} \\
\multicolumn{6}{l}{\emph{SCANO}} \\
SCANO & 0.028\,{\scriptsize$\pm$\,0.003} & 0.169\,{\scriptsize$\pm$\,0.025} & 0.198\,{\scriptsize$\pm$\,0.036} & 0.179\,{\scriptsize$\pm$\,0.034} & 777 \\
SCANO backbone + BQA & 0.559\,{\scriptsize$\pm$\,0.065} & 0.176\,{\scriptsize$\pm$\,0.024} & 0.216\,{\scriptsize$\pm$\,0.044} & 0.197\,{\scriptsize$\pm$\,0.039} & 777 \\
\midrule
\multicolumn{6}{l}{\emph{Boundary injection on the SCANO backbone}} \\
SCANO, anchors outside the message passing & 0.023\,{\scriptsize$\pm$\,0.003} & 0.190\,{\scriptsize$\pm$\,0.027} & 0.225\,{\scriptsize$\pm$\,0.047} & 0.206\,{\scriptsize$\pm$\,0.044} & 777 \\
SCANO without message passing & 0.024\,{\scriptsize$\pm$\,0.002} & 0.230\,{\scriptsize$\pm$\,0.035} & 0.270\,{\scriptsize$\pm$\,0.051} & 0.245\,{\scriptsize$\pm$\,0.044} & 787 \\
Global modulation & 0.409\,{\scriptsize$\pm$\,0.024} & 0.696\,{\scriptsize$\pm$\,0.091} & 0.776\,{\scriptsize$\pm$\,0.091} & 0.792\,{\scriptsize$\pm$\,0.071} & 851 \\
No boundary values$^{\S}$ & 0.010\,{\scriptsize$\pm$\,0.001} & 1$^{\ddagger}$ & 1$^{\ddagger}$ & 1$^{\ddagger}$ & 837 \\
\midrule
\multicolumn{6}{l}{\emph{Other boundary encoders}} \\
Transolver + LFE & 0.383\,{\scriptsize$\pm$\,0.036} & 0.299\,{\scriptsize$\pm$\,0.061} & 0.348\,{\scriptsize$\pm$\,0.056} & 0.322\,{\scriptsize$\pm$\,0.048} & 778 \\
Transolver + LFE, area & 0.348\,{\scriptsize$\pm$\,0.039} & 0.276\,{\scriptsize$\pm$\,0.073} & 0.312\,{\scriptsize$\pm$\,0.045} & 0.291\,{\scriptsize$\pm$\,0.039} & 778 \\
GNOT & 0.437\,{\scriptsize$\pm$\,0.038} & 0.410\,{\scriptsize$\pm$\,0.082} & 0.478\,{\scriptsize$\pm$\,0.067} & 0.434\,{\scriptsize$\pm$\,0.051} & 790 \\
GeoTransolver & 0.955\,{\scriptsize$\pm$\,0.077} & 0.716\,{\scriptsize$\pm$\,0.110} & 0.732\,{\scriptsize$\pm$\,0.090} & 0.703\,{\scriptsize$\pm$\,0.070} & 796 \\
Slice aggregation & 0.981\,{\scriptsize$\pm$\,0.076} & 0.540\,{\scriptsize$\pm$\,0.120} & 0.568\,{\scriptsize$\pm$\,0.072} & 0.543\,{\scriptsize$\pm$\,0.057} & 808 \\
BENO & 0.002\,{\scriptsize$\pm$\,0.000} & 0.638\,{\scriptsize$\pm$\,0.071} & 0.695\,{\scriptsize$\pm$\,0.070} & 0.673\,{\scriptsize$\pm$\,0.059} & 790 \\
\midrule
\multicolumn{6}{l}{\emph{Backbones}} \\
Transolver & 1.890\,{\scriptsize$\pm$\,0.170} & 0.850\,{\scriptsize$\pm$\,0.075} & 0.904\,{\scriptsize$\pm$\,0.050} & 1$^{\ddagger}$ & 803 \\
Transolver + BQA & 0.514\,{\scriptsize$\pm$\,0.043} & 0.297\,{\scriptsize$\pm$\,0.050} & 0.350\,{\scriptsize$\pm$\,0.052} & 0.319\,{\scriptsize$\pm$\,0.042} & 800 \\
\addlinespace[0.35em]
AB-UPT & 2.036\,{\scriptsize$\pm$\,0.160} & 0.502\,{\scriptsize$\pm$\,0.077} & 0.714\,{\scriptsize$\pm$\,0.074} & 1$^{\ddagger}$ & 817 \\
AB-UPT + BQA & 0.758\,{\scriptsize$\pm$\,0.075} & 0.137\,{\scriptsize$\pm$\,0.044} & 0.160\,{\scriptsize$\pm$\,0.046} & 0.140\,{\scriptsize$\pm$\,0.042} & 760 \\
AB-UPT, whole-mesh anchors by count & 0.386\,{\scriptsize$\pm$\,0.049} & 0.347\,{\scriptsize$\pm$\,0.070} & 0.393\,{\scriptsize$\pm$\,0.078} & 0.360\,{\scriptsize$\pm$\,0.070} & 817 \\
AB-UPT + BQA, whole-mesh anchors by count & 0.034\,{\scriptsize$\pm$\,0.003} & 0.152\,{\scriptsize$\pm$\,0.050} & 0.171\,{\scriptsize$\pm$\,0.047} & 0.152\,{\scriptsize$\pm$\,0.045} & 760 \\
AB-UPT + BQA + AGS & 0.032\,{\scriptsize$\pm$\,0.002} & 0.121\,{\scriptsize$\pm$\,0.024} & 0.157\,{\scriptsize$\pm$\,0.044} & 0.139\,{\scriptsize$\pm$\,0.040} & 760 \\
\addlinespace[0.35em]
AeroGTO & 2.085\,{\scriptsize$\pm$\,0.182} & 0.767\,{\scriptsize$\pm$\,0.072} & 0.864\,{\scriptsize$\pm$\,0.047} & 1$^{\ddagger}$ & 795 \\
AeroGTO + BQA & 0.974\,{\scriptsize$\pm$\,0.052} & 0.315\,{\scriptsize$\pm$\,0.120} & 0.289\,{\scriptsize$\pm$\,0.044} & 0.262\,{\scriptsize$\pm$\,0.042} & 799 \\
\bottomrule
\end{tabular}
\end{table*}

\begin{table*}[!tbp]
\centering
\scriptsize
\setlength{\tabcolsep}{3pt}
\caption{The uniform-sampling family of Table~\ref{tab:r0s-main}. Controlled benchmark, 512 test cases. The table gives the volume-weighted relative $L^2$ error over the cells in each deployment. Errors are averaged per case over the training seeds and then over the cases. $\pm$ is half the width of the 95\% case-bootstrap interval. \emph{Sample} is one sample drawn as in training. \emph{Whole mesh} is one pass. \emph{Region} is one compact region of interest that holds the given share of the cells (Section~\ref{sec:setup-deploy}). It is predicted from a pass over the region. Models with AGS draw fresh anchors there, and message-passing models read the halo of their receptive field. \emph{Other mesh} is the reverse-graded $40^3$ mesh. On the whole mesh, SCANO uses its whole-mesh state. Bold marks the lowest error of a column within a training family. Methods are compared within a training family, never across families (Table~\ref{tab:r0s-sampling}). $^{\S}$The model has no boundary input.}
\label{tab:r0s-uniform}
\begin{tabular}{P{4.3cm}ccccccc}
\toprule
 & & & \multicolumn{4}{c}{Region, share of the cells} & \\
\cmidrule(lr){4-7}
Method & Sample & Whole mesh & 0.3 & 0.12 & 0.05 & 0.02 & Other mesh \\
\midrule
\multicolumn{8}{l}{\textbf{Uniform samples, ratio 0.12}} \\
\multicolumn{8}{l}{\emph{SCANO}} \\
SCANO & 0.138\,{\tiny$\pm$\,0.009} & 0.138\,{\tiny$\pm$\,0.009} & 0.149\,{\tiny$\pm$\,0.011} & 0.158\,{\tiny$\pm$\,0.013} & 0.172\,{\tiny$\pm$\,0.018} & 0.189\,{\tiny$\pm$\,0.025} & 0.138\,{\tiny$\pm$\,0.009} \\
SCANO backbone + BQA & 0.132\,{\tiny$\pm$\,0.006} & 0.129\,{\tiny$\pm$\,0.006} & 0.378\,{\tiny$\pm$\,0.035} & 0.503\,{\tiny$\pm$\,0.056} & 0.606\,{\tiny$\pm$\,0.066} & 0.670\,{\tiny$\pm$\,0.071} & 0.129\,{\tiny$\pm$\,0.006} \\
\midrule
\multicolumn{8}{l}{\emph{Boundary injection on the SCANO backbone}} \\
Global modulation & 0.680\,{\tiny$\pm$\,0.035} & 0.680\,{\tiny$\pm$\,0.035} & 0.686\,{\tiny$\pm$\,0.040} & 0.746\,{\tiny$\pm$\,0.050} & 0.798\,{\tiny$\pm$\,0.065} & 0.863\,{\tiny$\pm$\,0.069} & 0.680\,{\tiny$\pm$\,0.035} \\
No boundary values$^{\S}$ & 1.013\,{\tiny$\pm$\,0.033} & 1.013\,{\tiny$\pm$\,0.033} & 1.050\,{\tiny$\pm$\,0.042} & 1.100\,{\tiny$\pm$\,0.054} & 1.147\,{\tiny$\pm$\,0.067} & 1.217\,{\tiny$\pm$\,0.071} & 1.013\,{\tiny$\pm$\,0.033} \\
\midrule
\multicolumn{8}{l}{\emph{Other boundary encoders}} \\
Transolver + LFE & 0.196\,{\tiny$\pm$\,0.022} & 0.196\,{\tiny$\pm$\,0.022} & 0.323\,{\tiny$\pm$\,0.031} & 0.429\,{\tiny$\pm$\,0.045} & 0.502\,{\tiny$\pm$\,0.062} & 0.577\,{\tiny$\pm$\,0.089} & 0.443\,{\tiny$\pm$\,0.038} \\
Transolver + LFE, area & 0.168\,{\tiny$\pm$\,0.016} & 0.167\,{\tiny$\pm$\,0.015} & 0.305\,{\tiny$\pm$\,0.028} & 0.406\,{\tiny$\pm$\,0.042} & 0.466\,{\tiny$\pm$\,0.054} & 0.528\,{\tiny$\pm$\,0.060} & 0.174\,{\tiny$\pm$\,0.015} \\
GNOT & 0.268\,{\tiny$\pm$\,0.014} & 0.268\,{\tiny$\pm$\,0.014} & 0.415\,{\tiny$\pm$\,0.028} & 0.501\,{\tiny$\pm$\,0.036} & 0.563\,{\tiny$\pm$\,0.051} & 0.657\,{\tiny$\pm$\,0.066} & 0.504\,{\tiny$\pm$\,0.037} \\
GeoTransolver & 0.501\,{\tiny$\pm$\,0.032} & 0.500\,{\tiny$\pm$\,0.032} & 0.688\,{\tiny$\pm$\,0.047} & 0.809\,{\tiny$\pm$\,0.061} & 0.905\,{\tiny$\pm$\,0.105} & 1.009\,{\tiny$\pm$\,0.118} & 0.559\,{\tiny$\pm$\,0.027} \\
Slice aggregation & 0.377\,{\tiny$\pm$\,0.026} & 0.379\,{\tiny$\pm$\,0.027} & 0.737\,{\tiny$\pm$\,0.047} & 0.975\,{\tiny$\pm$\,0.079} & 1.023\,{\tiny$\pm$\,0.086} & 1.129\,{\tiny$\pm$\,0.110} & 0.497\,{\tiny$\pm$\,0.029} \\
\midrule
\multicolumn{8}{l}{\emph{Backbones}} \\
Transolver & 0.822\,{\tiny$\pm$\,0.045} & 0.778\,{\tiny$\pm$\,0.044} & 1.364\,{\tiny$\pm$\,0.107} & 1.970\,{\tiny$\pm$\,0.173} & 2.263\,{\tiny$\pm$\,0.187} & 2.528\,{\tiny$\pm$\,0.225} & 0.720\,{\tiny$\pm$\,0.031} \\
Transolver + BQA & 0.178\,{\tiny$\pm$\,0.011} & 0.177\,{\tiny$\pm$\,0.011} & 0.346\,{\tiny$\pm$\,0.029} & 0.447\,{\tiny$\pm$\,0.050} & 0.514\,{\tiny$\pm$\,0.056} & 0.568\,{\tiny$\pm$\,0.071} & 0.215\,{\tiny$\pm$\,0.019} \\
\addlinespace[0.35em]
AB-UPT & 0.413\,{\tiny$\pm$\,0.033} & 0.221\,{\tiny$\pm$\,0.022} & 1.159\,{\tiny$\pm$\,0.097} & 1.871\,{\tiny$\pm$\,0.174} & 2.321\,{\tiny$\pm$\,0.223} & 2.845\,{\tiny$\pm$\,0.252} & 0.474\,{\tiny$\pm$\,0.038} \\
AB-UPT + BQA & 0.084\,{\tiny$\pm$\,0.010} & \textbf{0.077}\,{\tiny$\pm$\,0.008} & 0.466\,{\tiny$\pm$\,0.039} & 0.704\,{\tiny$\pm$\,0.066} & 0.846\,{\tiny$\pm$\,0.080} & 1.024\,{\tiny$\pm$\,0.114} & 0.221\,{\tiny$\pm$\,0.017} \\
AB-UPT, whole-mesh anchors by count & 0.241\,{\tiny$\pm$\,0.022} & 0.241\,{\tiny$\pm$\,0.022} & 0.459\,{\tiny$\pm$\,0.043} & 0.496\,{\tiny$\pm$\,0.066} & 0.533\,{\tiny$\pm$\,0.079} & 0.589\,{\tiny$\pm$\,0.096} & 0.459\,{\tiny$\pm$\,0.031} \\
AB-UPT + BQA, whole-mesh anchors by count & \textbf{0.079}\,{\tiny$\pm$\,0.010} & 0.079\,{\tiny$\pm$\,0.010} & \textbf{0.094}\,{\tiny$\pm$\,0.014} & \textbf{0.099}\,{\tiny$\pm$\,0.015} & \textbf{0.110}\,{\tiny$\pm$\,0.025} & 0.129\,{\tiny$\pm$\,0.038} & 0.209\,{\tiny$\pm$\,0.015} \\
AB-UPT + BQA + AGS & 0.083\,{\tiny$\pm$\,0.010} & 0.083\,{\tiny$\pm$\,0.010} & 0.095\,{\tiny$\pm$\,0.012} & 0.101\,{\tiny$\pm$\,0.014} & 0.112\,{\tiny$\pm$\,0.021} & \textbf{0.126}\,{\tiny$\pm$\,0.032} & \textbf{0.114}\,{\tiny$\pm$\,0.011} \\
\bottomrule
\end{tabular}
\end{table*}

\begin{table*}[!tbp]
\centering
\scriptsize
\setlength{\tabcolsep}{3pt}
\caption{The uniform-sampling family of Table~\ref{tab:r0s-main}. Controlled benchmark: drift, response and size of the models of Table~\ref{tab:r0s-uniform}. Values are averaged over the training seeds, per case and, for $E_{\mathrm{resp}}$, per intervention pair. $\pm$ is half the width of the 95\% case-bootstrap interval; for $E_{\mathrm{resp}}$, whole base cases are resampled. The region drift $\delta$ is measured in a compact region that holds the sampling ratio's share of the cells. It is the distance of the prediction there from the model's own whole-mesh prediction on the region's cells, relative to the reference. A model with AGS draws fresh anchors in the region, as in training. A message-passing model reads the halo of its receptive field. $E_{\mathrm{resp}}$ (Eq.~\eqref{eq:eresp}) is measured on one sample drawn as in training. It covers the intervention pairs whose true response exceeds five percent of the base solution. The columns give all pairs, the pairs on the smallest patches (tiny) and the far pairs, whose intervened patch has no face in the sample. $^{\ddagger}$Equal to 1 by construction. The model has no boundary input, or it reads no boundary record while the intervened patch lies outside the sample. Parameters are in thousands. Training families and marks as in Table~\ref{tab:r0s-uniform}.}
\label{tab:r0s-uniform-resp}
\begin{tabular}{lccccc}
\toprule
 & & \multicolumn{3}{c}{$E_{\mathrm{resp}}$ on a sample} & \\
\cmidrule(lr){3-5}
Method & Region drift $\delta$ & All & Tiny & Far & Parameters ($10^3$) \\
\midrule
\multicolumn{6}{l}{\textbf{Uniform samples, ratio 0.12}} \\
\multicolumn{6}{l}{\emph{SCANO}} \\
SCANO & 0.023\,{\scriptsize$\pm$\,0.003} & 0.242\,{\scriptsize$\pm$\,0.040} & 0.282\,{\scriptsize$\pm$\,0.053} & 0.284\,{\scriptsize$\pm$\,0.050} & 787 \\
SCANO backbone + BQA & 0.494\,{\scriptsize$\pm$\,0.057} & 0.216\,{\scriptsize$\pm$\,0.025} & 0.259\,{\scriptsize$\pm$\,0.050} & 0.251\,{\scriptsize$\pm$\,0.035} & 787 \\
\midrule
\multicolumn{6}{l}{\emph{Boundary injection on the SCANO backbone}} \\
Global modulation & 0.231\,{\scriptsize$\pm$\,0.015} & 0.996\,{\scriptsize$\pm$\,0.185} & 0.802\,{\scriptsize$\pm$\,0.098} & 0.835\,{\scriptsize$\pm$\,0.114} & 808 \\
No boundary values$^{\S}$ & 0.018\,{\scriptsize$\pm$\,0.001} & 1$^{\ddagger}$ & 1$^{\ddagger}$ & 1$^{\ddagger}$ & 783 \\
\midrule
\multicolumn{6}{l}{\emph{Other boundary encoders}} \\
Transolver + LFE & 0.373\,{\scriptsize$\pm$\,0.037} & 0.311\,{\scriptsize$\pm$\,0.078} & 0.346\,{\scriptsize$\pm$\,0.058} & 0.353\,{\scriptsize$\pm$\,0.065} & 778 \\
Transolver + LFE, area & 0.370\,{\scriptsize$\pm$\,0.041} & 0.270\,{\scriptsize$\pm$\,0.057} & 0.307\,{\scriptsize$\pm$\,0.045} & 0.316\,{\scriptsize$\pm$\,0.045} & 778 \\
GNOT & 0.402\,{\scriptsize$\pm$\,0.031} & 0.418\,{\scriptsize$\pm$\,0.095} & 0.467\,{\scriptsize$\pm$\,0.067} & 0.446\,{\scriptsize$\pm$\,0.046} & 790 \\
GeoTransolver & 0.603\,{\scriptsize$\pm$\,0.033} & 0.771\,{\scriptsize$\pm$\,0.110} & 0.798\,{\scriptsize$\pm$\,0.102} & 0.764\,{\scriptsize$\pm$\,0.082} & 796 \\
Slice aggregation & 0.875\,{\scriptsize$\pm$\,0.071} & 0.549\,{\scriptsize$\pm$\,0.113} & 0.571\,{\scriptsize$\pm$\,0.078} & 0.578\,{\scriptsize$\pm$\,0.083} & 808 \\
\midrule
\multicolumn{6}{l}{\emph{Backbones}} \\
Transolver & 1.977\,{\scriptsize$\pm$\,0.176} & 0.840\,{\scriptsize$\pm$\,0.073} & 0.880\,{\scriptsize$\pm$\,0.047} & 1$^{\ddagger}$ & 803 \\
Transolver + BQA & 0.409\,{\scriptsize$\pm$\,0.046} & 0.286\,{\scriptsize$\pm$\,0.045} & 0.339\,{\scriptsize$\pm$\,0.053} & 0.336\,{\scriptsize$\pm$\,0.046} & 800 \\
\addlinespace[0.35em]
AB-UPT & 1.818\,{\scriptsize$\pm$\,0.161} & 0.502\,{\scriptsize$\pm$\,0.080} & 0.746\,{\scriptsize$\pm$\,0.061} & 1$^{\ddagger}$ & 817 \\
AB-UPT + BQA & 0.695\,{\scriptsize$\pm$\,0.064} & 0.149\,{\scriptsize$\pm$\,0.033} & 0.185\,{\scriptsize$\pm$\,0.048} & 0.175\,{\scriptsize$\pm$\,0.046} & 760 \\
AB-UPT, whole-mesh anchors by count & 0.413\,{\scriptsize$\pm$\,0.058} & 0.354\,{\scriptsize$\pm$\,0.061} & 0.396\,{\scriptsize$\pm$\,0.071} & 0.377\,{\scriptsize$\pm$\,0.066} & 817 \\
AB-UPT + BQA, whole-mesh anchors by count & 0.034\,{\scriptsize$\pm$\,0.004} & 0.153\,{\scriptsize$\pm$\,0.048} & 0.170\,{\scriptsize$\pm$\,0.052} & 0.165\,{\scriptsize$\pm$\,0.050} & 760 \\
AB-UPT + BQA + AGS & 0.030\,{\scriptsize$\pm$\,0.002} & 0.158\,{\scriptsize$\pm$\,0.042} & 0.175\,{\scriptsize$\pm$\,0.048} & 0.175\,{\scriptsize$\pm$\,0.050} & 760 \\
\bottomrule
\end{tabular}
\end{table*}

\begin{table*}[!tbp]
\centering
\scriptsize
\setlength{\tabcolsep}{3pt}
\caption{Sparse sampling (ratio 0.02), as Table~\ref{tab:r0s-main}. Controlled benchmark, 512 test cases. The table gives the volume-weighted relative $L^2$ error over the cells in each deployment. Errors are averaged per case over the training seeds and then over the cases. $\pm$ is half the width of the 95\% case-bootstrap interval. \emph{Sample} is one sample drawn as in training. \emph{Whole mesh} is one pass. \emph{Region} is one compact region of interest that holds the given share of the cells (Section~\ref{sec:setup-deploy}). It is predicted from a pass over the region. Models with AGS draw fresh anchors there, and message-passing models read the halo of their receptive field. \emph{Other mesh} is the reverse-graded $40^3$ mesh. On the whole mesh, SCANO uses its whole-mesh state. Bold marks the lowest error of a column within a training family. Methods are compared within a training family, never across families (Table~\ref{tab:r0s-sampling}).}
\label{tab:r0s-strong}
\begin{tabular}{P{4.3cm}ccccccc}
\toprule
 & & & \multicolumn{4}{c}{Region, share of the cells} & \\
\cmidrule(lr){4-7}
Method & Sample & Whole mesh & 0.3 & 0.12 & 0.05 & 0.02 & Other mesh \\
\midrule
\multicolumn{8}{l}{\textbf{$k$-hop samples, ratio 0.02}} \\
\multicolumn{8}{l}{\emph{SCANO}} \\
SCANO & \textbf{0.126}\,{\tiny$\pm$\,0.010} & 0.126\,{\tiny$\pm$\,0.010} & 0.143\,{\tiny$\pm$\,0.012} & 0.152\,{\tiny$\pm$\,0.015} & 0.167\,{\tiny$\pm$\,0.025} & \textbf{0.176}\,{\tiny$\pm$\,0.022} & 0.126\,{\tiny$\pm$\,0.009} \\
SCANO backbone + BQA & 0.130\,{\tiny$\pm$\,0.011} & \textbf{0.124}\,{\tiny$\pm$\,0.010} & 0.287\,{\tiny$\pm$\,0.037} & 0.409\,{\tiny$\pm$\,0.057} & 0.505\,{\tiny$\pm$\,0.067} & 0.596\,{\tiny$\pm$\,0.082} & \textbf{0.124}\,{\tiny$\pm$\,0.010} \\
\midrule
\multicolumn{8}{l}{\emph{Boundary injection on the SCANO backbone}} \\
SCANO, anchors outside the message passing & 0.127\,{\tiny$\pm$\,0.011} & 0.127\,{\tiny$\pm$\,0.011} & \textbf{0.138}\,{\tiny$\pm$\,0.013} & \textbf{0.147}\,{\tiny$\pm$\,0.016} & \textbf{0.165}\,{\tiny$\pm$\,0.026} & 0.177\,{\tiny$\pm$\,0.031} & 0.127\,{\tiny$\pm$\,0.011} \\
\midrule
\multicolumn{8}{l}{\emph{Other boundary encoders}} \\
Transolver + LFE & 0.218\,{\tiny$\pm$\,0.027} & 0.216\,{\tiny$\pm$\,0.026} & 0.297\,{\tiny$\pm$\,0.032} & 0.372\,{\tiny$\pm$\,0.042} & 0.416\,{\tiny$\pm$\,0.047} & 0.476\,{\tiny$\pm$\,0.064} & 0.459\,{\tiny$\pm$\,0.042} \\
\midrule
\multicolumn{8}{l}{\emph{Backbones}} \\
Transolver & 0.917\,{\tiny$\pm$\,0.050} & 0.815\,{\tiny$\pm$\,0.037} & 1.080\,{\tiny$\pm$\,0.069} & 1.369\,{\tiny$\pm$\,0.118} & 1.758\,{\tiny$\pm$\,0.210} & 2.153\,{\tiny$\pm$\,0.213} & 0.772\,{\tiny$\pm$\,0.025} \\
Transolver + BQA & 0.195\,{\tiny$\pm$\,0.013} & 0.195\,{\tiny$\pm$\,0.013} & 0.307\,{\tiny$\pm$\,0.027} & 0.390\,{\tiny$\pm$\,0.049} & 0.436\,{\tiny$\pm$\,0.054} & 0.496\,{\tiny$\pm$\,0.064} & 0.209\,{\tiny$\pm$\,0.015} \\
\midrule[\heavyrulewidth]
\multicolumn{8}{l}{\textbf{Uniform samples, ratio 0.02}} \\
\multicolumn{8}{l}{\emph{SCANO}} \\
SCANO & \textbf{0.130}\,{\tiny$\pm$\,0.007} & \textbf{0.130}\,{\tiny$\pm$\,0.007} & \textbf{0.139}\,{\tiny$\pm$\,0.008} & \textbf{0.148}\,{\tiny$\pm$\,0.010} & \textbf{0.163}\,{\tiny$\pm$\,0.015} & \textbf{0.175}\,{\tiny$\pm$\,0.018} & \textbf{0.130}\,{\tiny$\pm$\,0.007} \\
SCANO backbone + BQA & 0.148\,{\tiny$\pm$\,0.008} & 0.143\,{\tiny$\pm$\,0.007} & 0.270\,{\tiny$\pm$\,0.024} & 0.365\,{\tiny$\pm$\,0.044} & 0.454\,{\tiny$\pm$\,0.057} & 0.540\,{\tiny$\pm$\,0.070} & 0.143\,{\tiny$\pm$\,0.007} \\
\midrule
\multicolumn{8}{l}{\emph{Other boundary encoders}} \\
Transolver + LFE & 0.204\,{\tiny$\pm$\,0.020} & 0.205\,{\tiny$\pm$\,0.020} & 0.311\,{\tiny$\pm$\,0.027} & 0.402\,{\tiny$\pm$\,0.039} & 0.461\,{\tiny$\pm$\,0.046} & 0.528\,{\tiny$\pm$\,0.058} & 0.451\,{\tiny$\pm$\,0.039} \\
\midrule
\multicolumn{8}{l}{\emph{Backbones}} \\
Transolver & 0.885\,{\tiny$\pm$\,0.045} & 0.797\,{\tiny$\pm$\,0.038} & 1.047\,{\tiny$\pm$\,0.063} & 1.332\,{\tiny$\pm$\,0.112} & 1.594\,{\tiny$\pm$\,0.151} & 1.858\,{\tiny$\pm$\,0.163} & 0.759\,{\tiny$\pm$\,0.030} \\
Transolver + BQA & 0.189\,{\tiny$\pm$\,0.010} & 0.187\,{\tiny$\pm$\,0.011} & 0.295\,{\tiny$\pm$\,0.025} & 0.382\,{\tiny$\pm$\,0.047} & 0.434\,{\tiny$\pm$\,0.055} & 0.512\,{\tiny$\pm$\,0.074} & 0.203\,{\tiny$\pm$\,0.014} \\
\bottomrule
\end{tabular}
\end{table*}

\begin{table*}[!tbp]
\centering
\scriptsize
\setlength{\tabcolsep}{3pt}
\caption{Sparse sampling (ratio 0.02), as Table~\ref{tab:r0s-main}. Controlled benchmark: drift, response and size of the models of Table~\ref{tab:r0s-strong}. Values are averaged over the training seeds, per case and, for $E_{\mathrm{resp}}$, per intervention pair. $\pm$ is half the width of the 95\% case-bootstrap interval; for $E_{\mathrm{resp}}$, whole base cases are resampled. The region drift $\delta$ is measured in a compact region that holds the sampling ratio's share of the cells. It is the distance of the prediction there from the model's own whole-mesh prediction on the region's cells, relative to the reference. A model with AGS draws fresh anchors in the region, as in training. A message-passing model reads the halo of its receptive field. $E_{\mathrm{resp}}$ (Eq.~\eqref{eq:eresp}) is measured on one sample drawn as in training. It covers the intervention pairs whose true response exceeds five percent of the base solution. The columns give all pairs, the pairs on the smallest patches (tiny) and the far pairs, whose intervened patch has no face in the sample. $^{\ddagger}$Equal to 1 by construction. The model has no boundary input, or it reads no boundary record while the intervened patch lies outside the sample. Parameters are in thousands. Training families and marks as in Table~\ref{tab:r0s-strong}.}
\label{tab:r0s-strong-resp}
\begin{tabular}{lccccc}
\toprule
 & & \multicolumn{3}{c}{$E_{\mathrm{resp}}$ on a sample} & \\
\cmidrule(lr){3-5}
Method & Region drift $\delta$ & All & Tiny & Far & Parameters ($10^3$) \\
\midrule
\multicolumn{6}{l}{\textbf{$k$-hop samples, ratio 0.02}} \\
\multicolumn{6}{l}{\emph{SCANO}} \\
SCANO & 0.053\,{\scriptsize$\pm$\,0.008} & 0.208\,{\scriptsize$\pm$\,0.035} & 0.237\,{\scriptsize$\pm$\,0.050} & 0.223\,{\scriptsize$\pm$\,0.046} & 777 \\
SCANO backbone + BQA & 0.584\,{\scriptsize$\pm$\,0.072} & 0.203\,{\scriptsize$\pm$\,0.027} & 0.250\,{\scriptsize$\pm$\,0.047} & 0.217\,{\scriptsize$\pm$\,0.032} & 777 \\
\midrule
\multicolumn{6}{l}{\emph{Boundary injection on the SCANO backbone}} \\
SCANO, anchors outside the message passing & 0.027\,{\scriptsize$\pm$\,0.004} & 0.214\,{\scriptsize$\pm$\,0.040} & 0.242\,{\scriptsize$\pm$\,0.046} & 0.233\,{\scriptsize$\pm$\,0.052} & 777 \\
\midrule
\multicolumn{6}{l}{\emph{Other boundary encoders}} \\
Transolver + LFE & 0.350\,{\scriptsize$\pm$\,0.032} & 0.326\,{\scriptsize$\pm$\,0.081} & 0.346\,{\scriptsize$\pm$\,0.042} & 0.366\,{\scriptsize$\pm$\,0.111} & 778 \\
\midrule
\multicolumn{6}{l}{\emph{Backbones}} \\
Transolver & 1.972\,{\scriptsize$\pm$\,0.176} & 0.930\,{\scriptsize$\pm$\,0.031} & 1.007\,{\scriptsize$\pm$\,0.020} & 1$^{\ddagger}$ & 803 \\
Transolver + BQA & 0.426\,{\scriptsize$\pm$\,0.053} & 0.313\,{\scriptsize$\pm$\,0.057} & 0.365\,{\scriptsize$\pm$\,0.060} & 0.349\,{\scriptsize$\pm$\,0.075} & 800 \\
\midrule[\heavyrulewidth]
\multicolumn{6}{l}{\textbf{Uniform samples, ratio 0.02}} \\
\multicolumn{6}{l}{\emph{SCANO}} \\
SCANO & 0.028\,{\scriptsize$\pm$\,0.005} & 0.221\,{\scriptsize$\pm$\,0.032} & 0.258\,{\scriptsize$\pm$\,0.045} & 0.241\,{\scriptsize$\pm$\,0.038} & 787 \\
SCANO backbone + BQA & 0.516\,{\scriptsize$\pm$\,0.060} & 0.251\,{\scriptsize$\pm$\,0.037} & 0.281\,{\scriptsize$\pm$\,0.044} & 0.275\,{\scriptsize$\pm$\,0.048} & 787 \\
\midrule
\multicolumn{6}{l}{\emph{Other boundary encoders}} \\
Transolver + LFE & 0.432\,{\scriptsize$\pm$\,0.039} & 0.317\,{\scriptsize$\pm$\,0.069} & 0.355\,{\scriptsize$\pm$\,0.051} & 0.356\,{\scriptsize$\pm$\,0.090} & 778 \\
\midrule
\multicolumn{6}{l}{\emph{Backbones}} \\
Transolver & 1.674\,{\scriptsize$\pm$\,0.130} & 0.948\,{\scriptsize$\pm$\,0.043} & 1.037\,{\scriptsize$\pm$\,0.064} & 1$^{\ddagger}$ & 803 \\
Transolver + BQA & 0.458\,{\scriptsize$\pm$\,0.081} & 0.316\,{\scriptsize$\pm$\,0.069} & 0.356\,{\scriptsize$\pm$\,0.058} & 0.359\,{\scriptsize$\pm$\,0.092} & 800 \\
\bottomrule
\end{tabular}
\end{table*}

\begin{table}[t]
\centering
\small
\caption{Sampling gap on the controlled benchmark. The gap is the whole-mesh error of a model trained on samples minus that of the same method trained on the whole mesh, paired per test case. $\pm$ is half the width of the 95\% case-bootstrap interval. The range gives the gap of each training seed of the sampled model against the model trained on the whole mesh. Errors are as in Table~\ref{tab:r0s-main}. A negative gap means that the model trained on samples is the more accurate on the whole mesh. Trained on the whole mesh, SCANO and the SCANO backbone with BQA alone are one model, because the anchor sample is the mesh itself. Both are therefore compared with it, on $k$-hop samples only. On uniform samples, the backbone runs without its message-passing layers. The whole-mesh models chose their learning rate on validation.}
\label{tab:r0s-tax}
\begin{tabular}{lccc}
\toprule
Training family & Sampled model & Gap & Range over seeds \\
\midrule
\multicolumn{4}{l}{\emph{SCANO}: trained on the whole mesh 0.078\,{\scriptsize$\pm$\,0.005}} \\
\quad $k$-hop samples, ratio 0.12 & 0.111\,{\scriptsize$\pm$\,0.009} & $+0.032$\,{\scriptsize$\pm$\,0.006} & $+0.027$ to $+0.037$ \\
\quad $k$-hop samples, ratio 0.02 & 0.126\,{\scriptsize$\pm$\,0.010} & $+0.048$\,{\scriptsize$\pm$\,0.006} & $+0.042$ to $+0.054$ \\
\midrule
\multicolumn{4}{l}{\emph{SCANO backbone + BQA}: trained on the whole mesh 0.078\,{\scriptsize$\pm$\,0.005}} \\
\quad $k$-hop samples, ratio 0.12 & 0.109\,{\scriptsize$\pm$\,0.008} & $+0.031$\,{\scriptsize$\pm$\,0.005} & $+0.021$ to $+0.043$ \\
\quad $k$-hop samples, ratio 0.02 & 0.124\,{\scriptsize$\pm$\,0.010} & $+0.045$\,{\scriptsize$\pm$\,0.007} & $+0.041$ to $+0.050$ \\
\midrule
\multicolumn{4}{l}{\emph{Transolver + LFE}: trained on the whole mesh 0.195\,{\scriptsize$\pm$\,0.021}} \\
\quad $k$-hop samples, ratio 0.12 & 0.207\,{\scriptsize$\pm$\,0.026} & $+0.013$\,{\scriptsize$\pm$\,0.007} & $-0.004$ to $+0.023$ \\
\quad Uniform samples, ratio 0.12 & 0.196\,{\scriptsize$\pm$\,0.022} & $+0.001$\,{\scriptsize$\pm$\,0.004} & $-0.007$ to $+0.007$ \\
\quad $k$-hop samples, ratio 0.02 & 0.216\,{\scriptsize$\pm$\,0.026} & $+0.021$\,{\scriptsize$\pm$\,0.007} & $+0.000$ to $+0.038$ \\
\quad Uniform samples, ratio 0.02 & 0.205\,{\scriptsize$\pm$\,0.020} & $+0.010$\,{\scriptsize$\pm$\,0.004} & $+0.000$ to $+0.025$ \\
\bottomrule
\end{tabular}
\end{table}

\section{The public benchmark: sparse sampling and the sampling gap}
\label{supp:olbc-strong}

Table~\ref{tab:olbc-strong} repeats the comparison of Table~\ref{tab:olbc-main} for training samples of about 25\% and
5\% of the mesh. Table~\ref{tab:olbc-tax} gives the sampling gap against training on the whole mesh
(Section~\ref{sec:test-c3}).

\begin{table*}[t]
\centering
\small
\caption{Public benchmark under sparse sampling, 256 test cases. The table gives the relative $L^2$ error in percent. Errors are averaged per case over the training seeds and then over the cases. $\pm$ is half the width of the 95\% case-bootstrap interval. \emph{Whole mesh} is one pass over the whole mesh, with SCANO in its whole-mesh state. \emph{Subgraph} is one subgraph drawn as in training. \emph{Region} is one compact region of interest that holds 5\% of the points (Section~\ref{sec:setup-deploy}). The last two are averaged over three draws per case. Models with AGS draw fresh anchors, and message-passing models read the halo of their receptive field. The region drift $\delta$ is the distance of the region prediction from the model's own whole-mesh prediction on the same points, relative to the reference, in percent. Bold marks the lowest value of a column within a setting. On this small mesh, SCANO's anchor sample is the whole mesh, passed through the message passing. For samples of about 5\% of the mesh, SCANO draws its anchors stratified over the mesh, outside the message passing. So does the row with anchors outside the message passing. $^{\ddagger}$Zero by construction (up to round-off) with whole-mesh anchors. These values are not ranked. A method marked (failed runs) has runs whose training failed by a criterion fixed before the test results were read. The criterion is a validation error of at least 0.85 times that of a model without boundary values. The row below it leaves these runs out.}
\label{tab:olbc-strong}
\begin{tabular}{lcccc}
\toprule
Method & Whole mesh & Subgraph & Region & Region drift $\delta$ \\
\midrule
\multicolumn{5}{l}{\textbf{Sparse sampling: one training sample covers about 25\% of the mesh}} \\
\multicolumn{5}{l}{\emph{SCANO}} \\
SCANO & 2.23\,{\scriptsize$\pm$\,0.10} & 2.49\,{\scriptsize$\pm$\,0.11} & 2.91\,{\scriptsize$\pm$\,0.20} & 0$^{\ddagger}$ \\
SCANO, anchors outside the message passing & 2.32\,{\scriptsize$\pm$\,0.11} & 2.70\,{\scriptsize$\pm$\,0.12} & 3.19\,{\scriptsize$\pm$\,0.23} & \textbf{0.92}\,{\scriptsize$\pm$\,0.07} \\
SCANO backbone + BQA & 2.65\,{\scriptsize$\pm$\,0.13} & 3.27\,{\scriptsize$\pm$\,0.15} & 14.09\,{\scriptsize$\pm$\,1.31} & 13.84\,{\scriptsize$\pm$\,1.27} \\
\midrule
\multicolumn{5}{l}{\emph{Boundary injection on the SCANO backbone without message passing}} \\
SCANO without message passing & 2.79\,{\scriptsize$\pm$\,0.12} & 3.07\,{\scriptsize$\pm$\,0.13} & 3.72\,{\scriptsize$\pm$\,0.26} & 1.73\,{\scriptsize$\pm$\,0.15} \\
\midrule
\multicolumn{5}{l}{\emph{Other boundary encoders and message passing}} \\
Transolver + LFE & 5.31\,{\scriptsize$\pm$\,0.24} & 5.36\,{\scriptsize$\pm$\,0.25} & 18.03\,{\scriptsize$\pm$\,1.28} & 16.17\,{\scriptsize$\pm$\,1.28} \\
GNOT & 9.53\,{\scriptsize$\pm$\,0.43} & 10.07\,{\scriptsize$\pm$\,0.46} & 28.26\,{\scriptsize$\pm$\,2.03} & 25.72\,{\scriptsize$\pm$\,1.91} \\
GeoTransolver & 12.93\,{\scriptsize$\pm$\,0.58} & 13.53\,{\scriptsize$\pm$\,0.60} & 56.58\,{\scriptsize$\pm$\,3.63} & 54.82\,{\scriptsize$\pm$\,3.72} \\
Slice aggregation & 13.06\,{\scriptsize$\pm$\,0.55} & 14.11\,{\scriptsize$\pm$\,0.55} & 66.43\,{\scriptsize$\pm$\,3.97} & 65.38\,{\scriptsize$\pm$\,4.22} \\
\midrule
\multicolumn{5}{l}{\emph{Backbones}} \\
Transolver (failed runs) & 68.29\,{\scriptsize$\pm$\,1.53} & 73.61\,{\scriptsize$\pm$\,1.81} & 110.32\,{\scriptsize$\pm$\,8.59} & 157.98\,{\scriptsize$\pm$\,11.68} \\
\quad without failed runs & 65.14\,{\scriptsize$\pm$\,1.59} & 71.14\,{\scriptsize$\pm$\,1.92} & 117.72\,{\scriptsize$\pm$\,9.90} & 167.71\,{\scriptsize$\pm$\,13.28} \\
Transolver + BQA & 3.39\,{\scriptsize$\pm$\,0.15} & 3.88\,{\scriptsize$\pm$\,0.17} & 17.88\,{\scriptsize$\pm$\,1.53} & 17.46\,{\scriptsize$\pm$\,1.56} \\
\addlinespace[0.35em]
AB-UPT & 10.19\,{\scriptsize$\pm$\,0.53} & 31.57\,{\scriptsize$\pm$\,1.68} & 95.58\,{\scriptsize$\pm$\,8.99} & 95.37\,{\scriptsize$\pm$\,8.94} \\
AB-UPT + BQA & 3.22\,{\scriptsize$\pm$\,0.15} & 3.63\,{\scriptsize$\pm$\,0.18} & 24.15\,{\scriptsize$\pm$\,2.22} & 24.00\,{\scriptsize$\pm$\,2.24} \\
AB-UPT + AGS & 2.84\,{\scriptsize$\pm$\,0.13} & 6.09\,{\scriptsize$\pm$\,0.34} & 7.19\,{\scriptsize$\pm$\,0.59} & 6.18\,{\scriptsize$\pm$\,0.57} \\
AB-UPT + BQA + AGS & \textbf{1.65}\,{\scriptsize$\pm$\,0.09} & \textbf{1.88}\,{\scriptsize$\pm$\,0.09} & \textbf{2.45}\,{\scriptsize$\pm$\,0.17} & 1.10\,{\scriptsize$\pm$\,0.07} \\
\midrule[\heavyrulewidth]
\multicolumn{5}{l}{\textbf{Sparse sampling: one training sample covers about 5\% of the mesh}} \\
\multicolumn{5}{l}{\emph{SCANO}} \\
SCANO & 2.48\,{\scriptsize$\pm$\,0.11} & 2.75\,{\scriptsize$\pm$\,0.12} & 3.35\,{\scriptsize$\pm$\,0.22} & \textbf{0.91}\,{\scriptsize$\pm$\,0.06} \\
SCANO backbone + BQA & 3.21\,{\scriptsize$\pm$\,0.15} & 3.75\,{\scriptsize$\pm$\,0.18} & 12.51\,{\scriptsize$\pm$\,1.16} & 12.37\,{\scriptsize$\pm$\,1.18} \\
\midrule
\multicolumn{5}{l}{\emph{Boundary injection on the SCANO backbone without message passing}} \\
SCANO without message passing & 2.85\,{\scriptsize$\pm$\,0.13} & 3.14\,{\scriptsize$\pm$\,0.13} & 3.81\,{\scriptsize$\pm$\,0.26} & 1.71\,{\scriptsize$\pm$\,0.15} \\
\midrule
\multicolumn{5}{l}{\emph{Other boundary encoders and message passing}} \\
Transolver + LFE & 5.68\,{\scriptsize$\pm$\,0.29} & 5.65\,{\scriptsize$\pm$\,0.30} & 11.81\,{\scriptsize$\pm$\,0.77} & 8.31\,{\scriptsize$\pm$\,0.65} \\
GNOT & 10.10\,{\scriptsize$\pm$\,0.46} & 10.96\,{\scriptsize$\pm$\,0.47} & 25.66\,{\scriptsize$\pm$\,2.02} & 21.90\,{\scriptsize$\pm$\,2.00} \\
GeoTransolver & 13.58\,{\scriptsize$\pm$\,0.56} & 14.11\,{\scriptsize$\pm$\,0.57} & 49.90\,{\scriptsize$\pm$\,2.74} & 48.97\,{\scriptsize$\pm$\,2.87} \\
Slice aggregation & 13.53\,{\scriptsize$\pm$\,0.58} & 14.81\,{\scriptsize$\pm$\,0.60} & 43.49\,{\scriptsize$\pm$\,2.51} & 40.53\,{\scriptsize$\pm$\,2.45} \\
\midrule
\multicolumn{5}{l}{\emph{Backbones}} \\
Transolver (failed runs) & 80.94\,{\scriptsize$\pm$\,1.39} & 85.71\,{\scriptsize$\pm$\,1.65} & 83.25\,{\scriptsize$\pm$\,5.28} & 117.05\,{\scriptsize$\pm$\,8.57} \\
\quad without failed runs & 72.89\,{\scriptsize$\pm$\,1.52} & 80.89\,{\scriptsize$\pm$\,1.85} & 83.15\,{\scriptsize$\pm$\,5.42} & 114.03\,{\scriptsize$\pm$\,8.49} \\
Transolver + BQA & 4.48\,{\scriptsize$\pm$\,0.20} & 5.06\,{\scriptsize$\pm$\,0.23} & 15.74\,{\scriptsize$\pm$\,1.27} & 14.84\,{\scriptsize$\pm$\,1.31} \\
\addlinespace[0.35em]
AB-UPT & 26.52\,{\scriptsize$\pm$\,1.11} & 63.87\,{\scriptsize$\pm$\,2.05} & 73.38\,{\scriptsize$\pm$\,4.38} & 69.63\,{\scriptsize$\pm$\,4.40} \\
AB-UPT + BQA & 3.48\,{\scriptsize$\pm$\,0.15} & 3.75\,{\scriptsize$\pm$\,0.19} & 17.97\,{\scriptsize$\pm$\,1.72} & 18.01\,{\scriptsize$\pm$\,1.73} \\
AB-UPT + AGS & 2.83\,{\scriptsize$\pm$\,0.13} & 6.07\,{\scriptsize$\pm$\,0.37} & 7.29\,{\scriptsize$\pm$\,0.60} & 6.18\,{\scriptsize$\pm$\,0.56} \\
AB-UPT + BQA + AGS & \textbf{2.01}\,{\scriptsize$\pm$\,0.10} & \textbf{2.17}\,{\scriptsize$\pm$\,0.10} & \textbf{2.95}\,{\scriptsize$\pm$\,0.21} & 1.07\,{\scriptsize$\pm$\,0.07} \\
\bottomrule
\end{tabular}
\end{table*}

\begin{table}[t]
\centering
\small
\caption{Sampling gap on the public benchmark, in percent. The gap is the whole-mesh error of a model trained on samples minus that of the same method trained on the whole mesh, paired per test case. $\pm$ is half the width of the 95\% case-bootstrap interval. The range gives the gap of each training seed of the sampled model against the model trained on the whole mesh. Errors are as in Table~\ref{tab:olbc-main}. A negative gap means that the model trained on samples is the more accurate on the whole mesh. Trained on the whole mesh, SCANO is the same function as the SCANO backbone with BQA alone, because the anchor sample is the mesh itself. This is its model trained on the whole mesh. For samples of about 5\% of the mesh, SCANO is the form with anchors outside the message passing.}
\label{tab:olbc-tax}
\begin{tabular}{lccc}
\toprule
Training samples & Sampled model & Gap & Range over seeds \\
\midrule
\multicolumn{4}{l}{\emph{SCANO}: trained on the whole mesh 2.08\,{\scriptsize$\pm$\,0.09}} \\
\quad about 76\% of the mesh per sample & 1.97\,{\scriptsize$\pm$\,0.09} & $-0.11$\,{\scriptsize$\pm$\,0.03} & $-0.14$ to $-0.10$ \\
\quad about 25\% of the mesh per sample & 2.23\,{\scriptsize$\pm$\,0.10} & $+0.15$\,{\scriptsize$\pm$\,0.05} & $+0.12$ to $+0.19$ \\
\midrule
\multicolumn{4}{l}{\emph{SCANO, anchors outside the message passing}: trained on the whole mesh 2.15\,{\scriptsize$\pm$\,0.11}} \\
\quad about 76\% of the mesh per sample & 2.22\,{\scriptsize$\pm$\,0.11} & $+0.07$\,{\scriptsize$\pm$\,0.05} & $-0.06$ to $+0.17$ \\
\quad about 25\% of the mesh per sample & 2.32\,{\scriptsize$\pm$\,0.11} & $+0.17$\,{\scriptsize$\pm$\,0.05} & $+0.08$ to $+0.35$ \\
\quad about 5\% of the mesh per sample & 2.48\,{\scriptsize$\pm$\,0.11} & $+0.33$\,{\scriptsize$\pm$\,0.06} & $+0.29$ to $+0.35$ \\
\midrule
\multicolumn{4}{l}{\emph{SCANO without message passing}: trained on the whole mesh 2.75\,{\scriptsize$\pm$\,0.12}} \\
\quad about 76\% of the mesh per sample & 2.76\,{\scriptsize$\pm$\,0.12} & $+0.01$\,{\scriptsize$\pm$\,0.03} & $-0.07$ to $+0.12$ \\
\quad about 25\% of the mesh per sample & 2.79\,{\scriptsize$\pm$\,0.12} & $+0.04$\,{\scriptsize$\pm$\,0.03} & $-0.00$ to $+0.12$ \\
\quad about 5\% of the mesh per sample & 2.85\,{\scriptsize$\pm$\,0.13} & $+0.10$\,{\scriptsize$\pm$\,0.03} & $+0.08$ to $+0.13$ \\
\midrule
\multicolumn{4}{l}{\emph{Transolver + LFE}: trained on the whole mesh 5.06\,{\scriptsize$\pm$\,0.24}} \\
\quad about 76\% of the mesh per sample & 5.09\,{\scriptsize$\pm$\,0.25} & $+0.02$\,{\scriptsize$\pm$\,0.06} & $-0.55$ to $+0.36$ \\
\quad about 25\% of the mesh per sample & 5.31\,{\scriptsize$\pm$\,0.24} & $+0.25$\,{\scriptsize$\pm$\,0.11} & $-0.24$ to $+0.58$ \\
\quad about 5\% of the mesh per sample & 5.68\,{\scriptsize$\pm$\,0.29} & $+0.62$\,{\scriptsize$\pm$\,0.09} & $+0.18$ to $+0.88$ \\
\bottomrule
\end{tabular}
\end{table}

\section{The public benchmark: LFE at its published settings}
\label{app:olbc}

The LFE variants of Section~\ref{sec:olbc-data} are re-implementations trained under our protocol. As a reference, we
also trained LFE with its official code, unchanged, at the settings of its paper (Table~\ref{tab:olbc-lfe}). We trained
both published models. Their processors follow RIGNO~\cite{rigno} and GAOT~\cite{gaot}. Training used 8,192 samples and
the paper's optimiser, learning-rate schedule, gradient clipping, number of epochs and extender size. The paper leaves
two details open, and the code settles them. First, the paper's batch of eight on two GPUs is read as eight per GPU, as
the code defines its batch option. This gives a global batch of sixteen, run as sixteen on one GPU. Second, the code
reads the two Robin input fields in swapped order. We keep this order, since the merged encoding remains invertible. The
reproduction counts as LFE at its published settings if its test error is of the paper's order. The error is measured in
the paper's metric, the median relative $L^2$ error over the test cases. Both models reach this level. The RIGNO-type
model comes within a few tenths of a percentage point of the error that the paper reports for it. The paper gives only a
bar chart for the GAOT-type model. Our reproduction of it is slightly more accurate.

We compare our models on the same 8,192 training cases and in the same metric. They are trained under the protocol of
Section~\ref{sec:olbc-data}, on samples covering about 76\% of the mesh for 60 epochs. SCANO reaches 1.10\%. This is the
mean over the training seeds of the per-seed medians, with a 95\% case-bootstrap interval of 1.04--1.16\%. It equals the
error of the GAOT-type LFE model trained on whole meshes for 2,048 epochs. Transolver reaches 1.78\% with BQA and 2.95\%
with the LFE extender under the same protocol.

\begin{table}[t]
\centering
\small
\caption{Settings of the official LFE reproduction (from the paper; details it leaves open from the code).}
\label{tab:olbc-lfe}
\begin{tabular}{P{0.3\linewidth}P{0.63\linewidth}}
\toprule
Data & 8,192 training and 256 validation samples of the official training file; the 256 official test samples \\
Optimiser & AdamW, weight decay $10^{-4}$; adaptive gradient clipping 0.5 \\
Learning rate & linear from $10^{-5}$ to $2\times10^{-4}$ over the first 5\% of training, then cosine to $10^{-5}$ \\
Epochs and batch & 2,048 epochs; global batch 16 (the paper's batch 8 read per GPU on its two GPUs), run on one GPU \\
Extender & 6 layers, 4 heads, latent size 128, 16 output features, no masking \\
Processors & RIGNO type: 12 steps, latent size 128, edge masking 0.5; GAOT type: $32\times32$ grid, patch size 2, 4
layers, 4 heads, latent size 128 \\
Parameters & 3.6 million and 3.9 million, as in the paper \\
\bottomrule
\end{tabular}
\end{table}

\section{DCTS: further tables}
\label{supp:dcts}

Table~\ref{tab:dcts-phys} gives the mean absolute error of each field in physical units. It covers every method of
Table~\ref{tab:dcts-main}, each in its own deployment. Table~\ref{tab:dcts-budget} varies the number of slices $G$ of
BQA in SCANO from 16 to 1,024 (Section~\ref{sec:slices}). Table~\ref{tab:dcts-tax} gives the sampling gap of Transolver,
Transolver + BQA and Transolver + BQA + AGS. These are the models that can also be trained on whole rooms
(Section~\ref{sec:test-c3}).

\begin{table*}[t]
\centering
\small
\caption{DCTS test rooms (50), models of Table~\ref{tab:dcts-main}. The table gives the mean absolute error of each field in physical units, in each model's own deployment. Errors are averaged over the rooms and then over the training seeds. $\pm$ is the standard deviation over the seeds. $p'$ is the kinematic pressure without its hydrostatic part; multiply it by the reference density to obtain Pa.}
\label{tab:dcts-phys}
\begin{tabular}{lccccc}
\toprule
Method & $U_x$ (m\,s$^{-1}$) & $U_y$ (m\,s$^{-1}$) & $U_z$ (m\,s$^{-1}$) & $p'$ (m$^2$\,s$^{-2}$) & $T$ (K) \\
\midrule
\multicolumn{6}{l}{\emph{SCANO}} \\
SCANO & 0.081\,{\scriptsize$\pm$\,0.000} & 0.058\,{\scriptsize$\pm$\,0.000} & 0.104\,{\scriptsize$\pm$\,0.000} & 0.138\,{\scriptsize$\pm$\,0.003} & 0.554\,{\scriptsize$\pm$\,0.006} \\
SCANO, anchors outside the message passing & 0.082\,{\scriptsize$\pm$\,0.000} & 0.059\,{\scriptsize$\pm$\,0.000} & 0.106\,{\scriptsize$\pm$\,0.000} & 0.134\,{\scriptsize$\pm$\,0.001} & 0.552\,{\scriptsize$\pm$\,0.003} \\
SCANO backbone + BQA & 0.078\,{\scriptsize$\pm$\,0.000} & 0.056\,{\scriptsize$\pm$\,0.000} & 0.100\,{\scriptsize$\pm$\,0.001} & 0.131\,{\scriptsize$\pm$\,0.001} & 0.548\,{\scriptsize$\pm$\,0.013} \\
\midrule
\multicolumn{6}{l}{\emph{Backbones without and with BQA}} \\
Transolver & 0.094\,{\scriptsize$\pm$\,0.002} & 0.064\,{\scriptsize$\pm$\,0.001} & 0.112\,{\scriptsize$\pm$\,0.002} & 0.199\,{\scriptsize$\pm$\,0.022} & 0.744\,{\scriptsize$\pm$\,0.063} \\
Transolver + BQA & 0.084\,{\scriptsize$\pm$\,0.000} & 0.061\,{\scriptsize$\pm$\,0.001} & 0.107\,{\scriptsize$\pm$\,0.001} & 0.144\,{\scriptsize$\pm$\,0.001} & 0.591\,{\scriptsize$\pm$\,0.001} \\
AB-UPT + AGS & 0.082\,{\scriptsize$\pm$\,0.003} & 0.056\,{\scriptsize$\pm$\,0.001} & 0.099\,{\scriptsize$\pm$\,0.000} & 0.141\,{\scriptsize$\pm$\,0.014} & 0.632\,{\scriptsize$\pm$\,0.027} \\
AB-UPT + BQA + AGS & 0.079\,{\scriptsize$\pm$\,0.001} & 0.055\,{\scriptsize$\pm$\,0.001} & 0.099\,{\scriptsize$\pm$\,0.002} & 0.145\,{\scriptsize$\pm$\,0.003} & 0.578\,{\scriptsize$\pm$\,0.056} \\
AeroGTO & 0.089\,{\scriptsize$\pm$\,0.010} & 0.061\,{\scriptsize$\pm$\,0.005} & 0.108\,{\scriptsize$\pm$\,0.005} & 0.230\,{\scriptsize$\pm$\,0.134} & 0.877\,{\scriptsize$\pm$\,0.439} \\
\midrule
\multicolumn{6}{l}{\emph{Other boundary encoders}} \\
Transolver + LFE & 0.087\,{\scriptsize$\pm$\,0.000} & 0.063\,{\scriptsize$\pm$\,0.000} & 0.109\,{\scriptsize$\pm$\,0.000} & 0.158\,{\scriptsize$\pm$\,0.007} & 0.627\,{\scriptsize$\pm$\,0.014} \\
GeoTransolver & 0.079\,{\scriptsize$\pm$\,0.000} & 0.057\,{\scriptsize$\pm$\,0.000} & 0.098\,{\scriptsize$\pm$\,0.001} & 0.139\,{\scriptsize$\pm$\,0.002} & 0.601\,{\scriptsize$\pm$\,0.005} \\
GNOT & 0.093\,{\scriptsize$\pm$\,0.002} & 0.067\,{\scriptsize$\pm$\,0.002} & 0.118\,{\scriptsize$\pm$\,0.004} & 0.163\,{\scriptsize$\pm$\,0.008} & 0.698\,{\scriptsize$\pm$\,0.055} \\
Slice aggregation & 0.081\,{\scriptsize$\pm$\,0.001} & 0.058\,{\scriptsize$\pm$\,0.000} & 0.102\,{\scriptsize$\pm$\,0.001} & 0.160\,{\scriptsize$\pm$\,0.010} & 0.647\,{\scriptsize$\pm$\,0.045} \\
\bottomrule
\end{tabular}
\end{table*}

\begin{table}[!htb]
\centering
\small
\caption{DCTS: the number of slices $G$ of BQA in SCANO, with everything else as in Table~\ref{tab:dcts-main} (50 test rooms, multiple training seeds). The error is measured in the model's own deployment, as the mean over the five fields of the volume-weighted relative $L^2$ error. The response error $E_{\mathrm{resp}}$ is that of the room's temperature to one rack's doubled power, over the ten held-out pairs of Table~\ref{tab:dcts-decision}. Both are relative to $G = 64$, the setting of the main text. $\pm$ is half the width of the 95\% bootstrap interval over the test rooms and over the pairs. The last column is the median training time of one epoch on one GPU.}
\label{tab:dcts-budget}
\begin{tabular}{cccc}
\toprule
$G$ & Error, own deployment & Response error $E_{\mathrm{resp}}$ & Time per epoch (min) \\
\midrule
16 & 1.000\,{\scriptsize$\pm$\,0.002} & 0.985\,{\scriptsize$\pm$\,0.026} & 11 \\
64 & 1 (reference) & 1 (reference) & 12 \\
256 & 1.002\,{\scriptsize$\pm$\,0.003} & 0.999\,{\scriptsize$\pm$\,0.014} & 16 \\
1,024 & 1.009\,{\scriptsize$\pm$\,0.002} & 0.995\,{\scriptsize$\pm$\,0.025} & 22 \\
\bottomrule
\end{tabular}
\end{table}

\begin{table}[!htb]
\centering
\small
\caption{Sampling gap on DCTS. The error is that of a single pass over the room, the metric of Table~\ref{tab:dcts-main}. The gap is the error of a model trained on samples (seeds at 5\% of the cells, two hops) minus that of the same method trained on the whole room. Everything else is unchanged, and the gap is paired per test room. $\pm$ is half the width of the 95\% case-bootstrap interval. The gap is the mean over multiple training seeds, and the last column gives the range of the per-seed gaps.}
\label{tab:dcts-tax}
\begin{tabular}{lccc}
\toprule
Training & Single pass over the room & Gap & Range over seeds \\
\midrule
\multicolumn{4}{l}{\emph{Transolver}: trained on the whole room 0.426\,{\scriptsize$\pm$\,0.027}} \\
\quad trained on samples & 0.554\,{\scriptsize$\pm$\,0.027} & $+0.128$\,{\scriptsize$\pm$\,0.017} & $+0.067$ to $+0.212$ \\
\midrule
\multicolumn{4}{l}{\emph{Transolver + BQA}: trained on the whole room 0.419\,{\scriptsize$\pm$\,0.024}} \\
\quad trained on samples & 0.431\,{\scriptsize$\pm$\,0.023} & $+0.012$\,{\scriptsize$\pm$\,0.002} & $+0.007$ to $+0.014$ \\
\midrule
\multicolumn{4}{l}{\emph{Transolver + BQA + AGS}: trained on the whole room 0.423\,{\scriptsize$\pm$\,0.023}} \\
\quad trained on samples & 0.415\,{\scriptsize$\pm$\,0.024} & $-0.008$\,{\scriptsize$\pm$\,0.002} & $-0.011$ to $-0.002$ \\
\bottomrule
\end{tabular}
\end{table}

\section{Ablations and checks behind Section 5}
\label{supp:ablations}

Table~\ref{tab:ablations} gives the readings that Section~\ref{sec:results} cites without a table in the main text. It
holds two ablations of SCANO on the controlled benchmark: records without types and records weighted by count. It gives
the drift of SCANO's prediction against the number of fresh anchors. It reports BQA on Transolver on DCTS without the
per-point heat-source input. It also gives the training time per epoch on samples and on whole rooms.

\begin{table}[!htb]
\centering
\footnotesize
\setlength{\tabcolsep}{3pt}
\caption{Ablations and checks behind Section~\ref{sec:results}. Each variant is trained with multiple seeds and paired by seed.
Intervals are 95\% case-bootstrap intervals, taken over the intervention pairs for the response. Ratios divide the
variant by SCANO or by the original backbone. The controlled benchmark is evaluated in both sampling families at ratio
0.12. For the anchor count, the table gives the range of the slopes over the training seeds. It is the slope of the mean squared drift of
the whole-mesh prediction against the number $N_a$ of fresh single-cell anchors (256 to 16,384, log axes). The reference
is the slope $-1.16$ of the variance $1/N_a-1/|\Omega_h|$ for anchors drawn without replacement. The table also gives
the error with 16,384 fresh anchors relative to the error with the whole-mesh state.}
\label{tab:ablations}
\begin{tabular}{P{0.36\linewidth}P{0.26\linewidth}P{0.15\linewidth}P{0.15\linewidth}}
\toprule
Variant & Quantity & $k$-hop & Uniform \\
\midrule
\multicolumn{4}{l}{\emph{Controlled benchmark (Sections~\ref{sec:test-c1}, \ref{sec:test-c2}, \ref{sec:test-c3})}} \\
SCANO without record types / SCANO & response error, interventions that switch the type of a patch & 3.40 {\scriptsize[2.67, 4.37]} & 3.03 {\scriptsize[2.16, 4.30]} \\
SCANO, records weighted by count / SCANO & error on the reverse-graded mesh & 3.44 {\scriptsize[3.21, 3.66]} & 2.80 {\scriptsize[2.65, 2.95]} \\
 & error on the training mesh & 1.28 {\scriptsize[1.23, 1.33]} & 1.03 {\scriptsize[1.00, 1.07]} \\
SCANO, anchors as single cells & slope of the mean squared drift & $-1.20$ to $-1.13$ & $-1.35$ to $-1.30$ \\
 & error with 16,384 anchors $-$ 1 & $-0.4$ to $+0.1$\% & $+0.1$ to $+0.4$\% \\
\midrule
\multicolumn{4}{l}{\emph{DCTS (Sections~\ref{sec:test-hosts}, \ref{sec:test-c3}), two-hop samples}} \\
Transolver + BQA / Transolver, both without the per-point heat-source input & error, own deployment & \multicolumn{2}{c}{0.94 {\scriptsize[0.92, 0.96]}} \\
 & error, one pass over the room & \multicolumn{2}{c}{0.69 {\scriptsize[0.64, 0.74]}} \\
 & error, compact region of 2\% of the cells & \multicolumn{2}{c}{0.67 {\scriptsize[0.59, 0.74]}} \\
Transolver + BQA + AGS, on samples / on whole rooms & training time per epoch & \multicolumn{2}{c}{0.30} \\
Transolver, on samples / on whole rooms & training time per epoch & \multicolumn{2}{c}{0.31} \\
\bottomrule
\end{tabular}
\end{table}

\section{Further figures}
\label{supp:figures}

\begin{figure}[t]
\centering
\includegraphics[width=\textwidth]{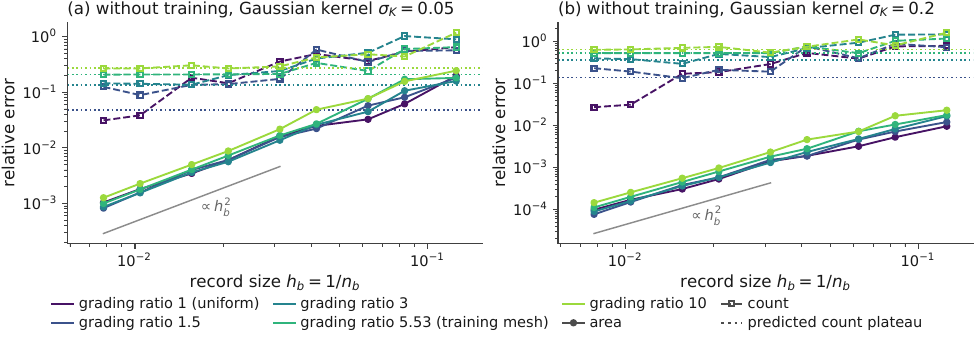}
\caption{Training-free comparison of count and area weights on the controlled benchmark. A fixed Gaussian kernel aggregates
refined boundary records. Its width is $\sigma_K = 0.05$ in (a) and $\sigma_K = 0.2$ in (b). The records have spacing
(\emph{record size}) $h_b = 1/n_b$ and grading ratios from 1 to 10 (dark to light). The error is relative to a fine uniform reference.
Area-weighted aggregation (solid, filled markers) converges at second order for all gradings. The grey guide is $\propto
h_b^2$. Count-weighted aggregation (dashed, open markers) on graded records stalls at the plateau predicted by
Proposition~\ref{prop:count} (dotted). At a grading ratio of 1, patch-boundary intersections create short edge fragments. Their
count-weighting error decays under refinement.}
\label{fig:k1}
\end{figure}

\begin{figure}[t]
\centering
\includegraphics{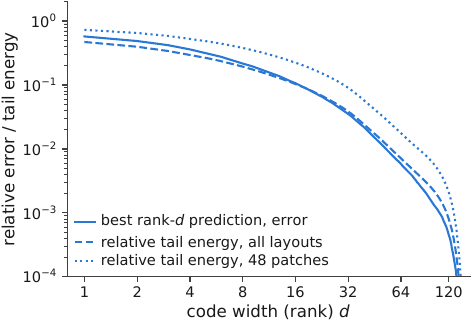}
\caption{Response spectrum of the controlled benchmark. The solid line is the optimal rank-$d$ approximation error (\emph{best rank-$d$ prediction, error}) of the exact
response operator. The other lines give the relative tail energy, for all layouts (dashed) and for 48 patches (dotted).
These curves set the theoretical error floor of a global code of width $d$ (Proposition~\ref{prop:code}). At the backbone's full channel width
($d = 96$), the trained global modulation model of Table~\ref{tab:r0s-main} remains over two orders of magnitude above
this floor.}
\label{fig:predictions}
\end{figure}

\begin{figure*}[t]
\centering
\includegraphics[width=\textwidth]{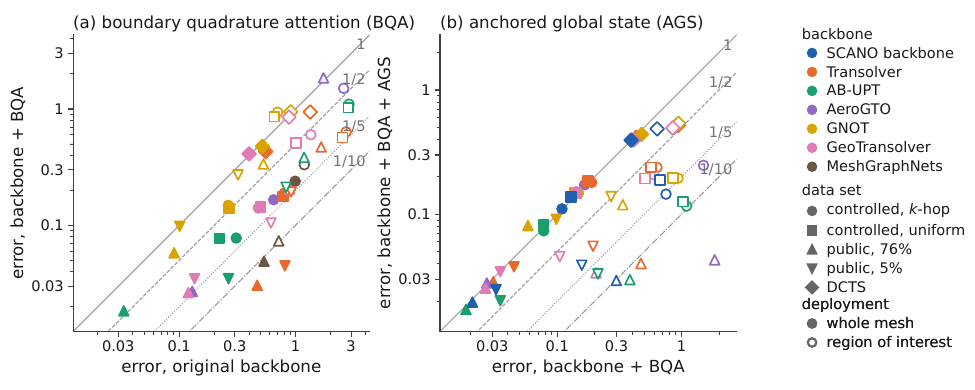}
\caption{Effect of BQA and AGS on all backbones (colour), data sets (marker shape) and deployments. Filled markers are the whole
mesh. Open markers are a compact region of 2\% of the cells. (a) Relative $L^2$ error with BQA against the original (native)
backbone. (b) Relative $L^2$ error with BQA + AGS against BQA alone. Grey reference lines mark $1\times$, $1/2\times$,
$1/5\times$ and $1/10\times$ the error. All configurations with completed training seeds are shown.}
\label{fig:hosts-scatter}
\end{figure*}

\begin{figure*}[t]
\centering
\includegraphics[width=\textwidth]{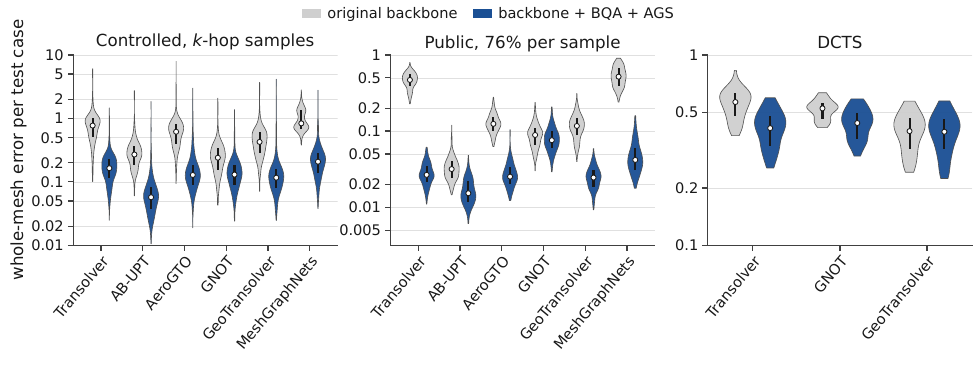}
\caption{Per-case whole-mesh error distributions of each original (native) backbone (grey) and of the same backbone with BQA + AGS (blue).
Where AGS does not apply, the blue distribution is BQA alone. The panels are the controlled benchmark ($k$-hop samples,
512 test cases), the public benchmark (76\% coverage, 256 test cases) and DCTS (50 test rooms). Bars mark interquartile
ranges, and white dots mark medians (log scale).}
\label{fig:hosts-violin}
\end{figure*}

\begin{figure}[!htb]
\centering
\includegraphics{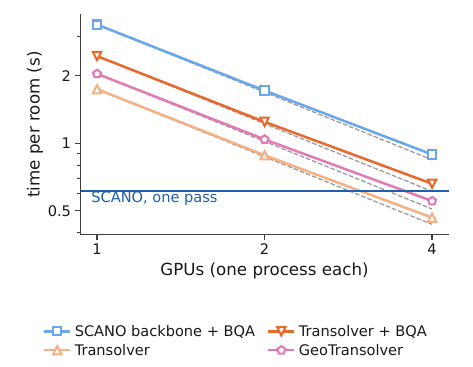}
\caption{The multi-pass average of the models whose state follows the sample on one, two and four GPUs. Each GPU runs
one process and takes its share of the passes in turn. The grey dashed lines mark the ideal speed-up. The solid line is SCANO's single
pass on one GPU. Values are medians over the four rooms of Table~\ref{tab:dcts-time}.}
\label{fig:cost-gpus}
\end{figure}

This section collects five further figures. Figure~\ref{fig:k1} belongs to Section~\ref{sec:test-c2}. It compares count
and area weights without training. Figure~\ref{fig:predictions} belongs to Section~\ref{sec:test-c4}. It shows the
response spectrum of the controlled benchmark. Figures~\ref{fig:hosts-scatter} and~\ref{fig:hosts-violin} belong to
Section~\ref{sec:test-hosts}. They show all backbones and data sets in one plot and their per-case errors.
Figure~\ref{fig:cost-gpus} belongs to Section~\ref{sec:dcts-cost}. It shows the multi-pass average on several GPUs.

\section{Further experiments and analyses}
\label{supp:r7-experiments}

This section collects the experiments and analyses that support Sections~\ref{sec:results} and~\ref{sec:dcts}. It first
examines the sampling design and the visibility of patches. It then turns to the sample consistency of the SCANO
backbone and to the origin of the gain of BQA. Next come the capacity and learning rate of the models and the deployment
protocols. The section ends with further evidence on DCTS, the statistics of the main comparisons and the DCTS reference
solutions. Every model follows the training protocol of Section~\ref{sec:setup-training} unless stated otherwise.

\subsection{Sampling design and the visibility of patches}
\label{supp:r7-sampling}

The $k$-hop sampler grows each training sample from seed cells along the cell graph. A boundary face enters a sample
only when it is drawn as a seed. A sample of the controlled benchmark at ratio 0.12 therefore holds most of the cells
but only a small share of the faces (Table~\ref{tab:r0s-sampling}). Two alternatives remove this asymmetry. In the
first, every training sample contains at least one face of every patch. No patch is then ever hidden. The second is the
face-inclusive sampler. Here a sample contains every boundary face owned by one of its cells. Face coverage then follows
cell coverage. At the training ratio of the controlled benchmark, the face-inclusive sampler raises the share of faces
in a sample from 12 to 58 per cent. This equals the share of the cells. The share of patches that a sample misses
entirely falls from 17 to about one per cent. The cost is 14 per cent more nodes per sample
(Table~\ref{tab:r0s-sampling}). Table~\ref{tab:r7-sampling} compares the backbones that read the boundary only as
sampled nodes under these designs. It also lists the same backbones with BQA.
On the controlled benchmark, the face-inclusive sampler lowers the whole-mesh error of every backbone. The gain is
largest for AB-UPT, whose anchors are drawn from the sample (from 0.31 to 0.12). It is smallest for the backbones with
BQA, which already read every face as an area-weighted record. Transolver + BQA goes from 0.19 to 0.18, and AB-UPT + BQA
+ AGS from 0.073 to 0.070. The gain of BQA remains on face-inclusive samples. Transolver + BQA is about four times as
accurate as the original Transolver. Its response is also far better. AB-UPT + BQA + AGS is 1.7 times as accurate as
AB-UPT. The records that BQA reads at every point therefore carry more than the faces that a sample would otherwise
hide.

\begin{table}[!htb]
\centering
\scriptsize
\setlength{\tabcolsep}{2.5pt}
\caption{Controlled benchmark: the backbones trained on $k$-hop samples at ratio 0.12, as in the main text, and on face-inclusive samples. A face-inclusive sample adds every boundary face owned by a sampled cell. The table gives the relative $L^2$ error on one sample drawn as in training and on the whole mesh, and the response error $E_{\mathrm{resp}}$ on the whole mesh. Values are seed means $\pm$ half the width of the 95\% case-bootstrap interval. The last row is the original Transolver with every patch present in every sample.}
\label{tab:r7-sampling}
\begin{tabular}{P{0.28\linewidth}cccccc}
\toprule
 & \multicolumn{3}{c}{$k$-hop samples} & \multicolumn{3}{c}{Face-inclusive samples} \\
\cmidrule(lr){2-4}\cmidrule(lr){5-7}
Method & Sample & Whole mesh & $E_{\mathrm{resp}}$ & Sample & Whole mesh & $E_{\mathrm{resp}}$ \\
\midrule
Transolver & 0.846\,{\scriptsize$\pm$\,0.055} & 0.800\,{\scriptsize$\pm$\,0.045} & 0.808\,{\scriptsize$\pm$\,0.075} & 0.702\,{\scriptsize$\pm$\,0.045} & 0.694\,{\scriptsize$\pm$\,0.046} & 0.788\,{\scriptsize$\pm$\,0.104} \\
Transolver + BQA & 0.186\,{\scriptsize$\pm$\,0.012} & 0.189\,{\scriptsize$\pm$\,0.012} & 0.299\,{\scriptsize$\pm$\,0.051} & 0.176\,{\scriptsize$\pm$\,0.010} & 0.176\,{\scriptsize$\pm$\,0.010} & 0.271\,{\scriptsize$\pm$\,0.040} \\
AB-UPT & 0.436\,{\scriptsize$\pm$\,0.044} & 0.309\,{\scriptsize$\pm$\,0.020} & 0.390\,{\scriptsize$\pm$\,0.062} & 0.177\,{\scriptsize$\pm$\,0.013} & 0.119\,{\scriptsize$\pm$\,0.009} & 0.193\,{\scriptsize$\pm$\,0.033} \\
AB-UPT, whole-mesh anchors by count & 0.256\,{\scriptsize$\pm$\,0.029} & 0.256\,{\scriptsize$\pm$\,0.029} & 0.344\,{\scriptsize$\pm$\,0.070} & 0.234\,{\scriptsize$\pm$\,0.023} & 0.235\,{\scriptsize$\pm$\,0.022} & 0.337\,{\scriptsize$\pm$\,0.067} \\
AB-UPT + BQA + AGS & 0.074\,{\scriptsize$\pm$\,0.009} & 0.073\,{\scriptsize$\pm$\,0.009} & 0.120\,{\scriptsize$\pm$\,0.024} & 0.070\,{\scriptsize$\pm$\,0.007} & 0.070\,{\scriptsize$\pm$\,0.007} & 0.122\,{\scriptsize$\pm$\,0.028} \\
\midrule
Transolver, every patch in every sample & 0.838\,{\scriptsize$\pm$\,0.059} & 0.794\,{\scriptsize$\pm$\,0.050} & 0.837\,{\scriptsize$\pm$\,0.091} & & & \\
\bottomrule
\end{tabular}
\end{table}

On DCTS, the face-inclusive sampler lowers the whole-room error of the original Transolver toward that of Transolver +
BQA (0.43 against 0.55 in one pass, and 0.43 against 0.46 in its own deployment, Table~\ref{tab:r7-sampling-dcts}). It
changes that of Transolver + BQA little (0.42 against 0.43 in one pass, and 0.42 against 0.41 in its own deployment). However, it breaks the response of
both models to a rack's power. The predicted change of the room temperature becomes two to seven times the true one. The
response errors are 5.5 and 2.6, against 0.72 and 0.73 on $k$-hop samples. The response of Transolver + BQA to a halved
cooling flow also worsens, from 0.38 to 0.85. The cause is as follows. A face-inclusive sample carries every outlet face
of its cells as a point of its own. The slice attention of the sample weighs these many small faces by their count. A
change at one rack is thus amplified in the state. On $k$-hop samples the faces are rare, and this path barely acts. On
DCTS, the faces are therefore no substitute for boundary records read with their areas. BQA provides such records.
The two forms of AB-UPT are not affected in this way. Their state comes from anchors drawn over the whole room, not from
the sample. On face-inclusive samples, their errors change by about three per cent. Their responses to a doubled rack do
not change. BQA keeps its advantage in the response to a halved cooling flow on both kinds of sample. This holds for
AB-UPT as for Transolver (Table~\ref{tab:r7-sampling-dcts}).

\begin{table}[!htb]
\centering
\scriptsize
\setlength{\tabcolsep}{2.5pt}
\caption{DCTS: the backbones trained on $k$-hop samples (Table~\ref{tab:dcts-main}) and on face-inclusive samples (50 test rooms, seed means). The table gives the error in the model's own deployment and the error on one region draw of the training sampler at the training budget. Such a draw fills a box of random size and tops it up with seeds from the whole room. It differs from the compact regions of Table~\ref{tab:dcts-main}. It also gives the response error $E_{\mathrm{resp}}$ of the room temperature to one rack's doubled power and to one cooling unit's halved flow, as the median over the ten held-out pairs of each kind.}
\label{tab:r7-sampling-dcts}
\begin{tabular}{P{0.2\linewidth}cccccccc}
\toprule
 & \multicolumn{4}{c}{$k$-hop samples} & \multicolumn{4}{c}{Face-inclusive samples} \\
\cmidrule(lr){2-5}\cmidrule(lr){6-9}
 & & & \multicolumn{2}{c}{$E_{\mathrm{resp}}$} & & & \multicolumn{2}{c}{$E_{\mathrm{resp}}$} \\
\cmidrule(lr){4-5}\cmidrule(lr){8-9}
Method & Error & Regional & Rack & Flow & Error & Regional & Rack & Flow \\
\midrule
Transolver & 0.457 & 0.782 & 0.72 & 0.85 & 0.428 & 0.742 & 5.53 & 1.03 \\
Transolver + BQA & 0.410 & 0.599 & 0.73 & 0.38 & 0.417 & 0.576 & 2.58 & 0.85 \\
AB-UPT + AGS & 0.378 & 0.410 & 0.62 & 0.86 & 0.390 & 0.421 & 0.63 & 0.84 \\
AB-UPT + BQA + AGS & 0.373 & 0.403 & 0.61 & 0.40 & 0.384 & 0.418 & 0.62 & 0.46 \\
\bottomrule
\end{tabular}
\end{table}

Even when every patch is present in every sample, the original Transolver remains far less accurate than Transolver +
BQA. The whole-mesh errors are 0.79 against 0.19 on $k$-hop samples and 0.74 against 0.18 on uniform samples. The gain
of BQA on this benchmark therefore does not come from patches that the sampler hides. Two differences remain. The faces
are weighted by count, and they enter as node values of the sampled points. BQA replaces both by area-weighted records
read at every point.

\subsection{Sample consistency of the SCANO backbone}
\label{supp:r7-halo}

A message-passing backbone trained on samples sees a truncated neighbourhood at points near the border of its sample. At
deployment, every point sees its full receptive field (Section~\ref{supp:scope}). A halo removes this difference in
training. The halo is twelve hops deep, the depth of SCANO's message passing. It enters the forward pass but not the
loss. Its effect on the sampling gap therefore isolates the truncation.
We trained SCANO in this way and evaluated it over the whole mesh, with its state drawn as in training. It reaches a
whole-mesh error of 0.109, against 0.112 without the halo. The difference lies inside the interval over the test cases.
The model trained on the whole mesh reaches 0.078. The truncated neighbourhood therefore accounts for little of SCANO's
sampling gap. Twelve hops of message passing train well on one-hop samples. The halo model must be evaluated in this way. In
training, its points always saw their full receptive field. Its anchor state, however, still came from a one-hop sample.
A state cached from every node was never encountered during training. It raises the error by about two fifths.
We also weighted the sampled anchors of AGS by $w_x/\pi_x$, where $\pi_x$ is the inclusion probability of the $k$-hop
design. Their estimate of the volume mean is then design-unbiased in the sense of Proposition~\ref{prop:drift}. On the
controlled benchmark, this leaves SCANO's error unchanged. The whole-mesh error is 0.112 against 0.111, and every
deployment is a tie. This is expected. The inclusion probabilities of this design are nearly uniform. They vary by about
four per cent across the cells and fall by a tenth only next to the boundary. DCTS has refined meshes, so its cells have
different numbers of neighbours. There, the same probabilities vary by about fourteen per cent.
In training, a model with AGS reads a state estimated from its anchors. At deployment, it reads the state of the whole
mesh. Consider SCANO with its anchors outside the message-passing layers. On the controlled benchmark, the state from 4,096
anchors drawn as in training differs from the whole-mesh state by 1.6 to 3 per cent of its norm. This is the variation
between two independent draws of the anchors. The deployed errors change by about one per cent between the two states.
The model therefore trains on a nearly unbiased and only mildly noisy version of the state it is deployed with.
Weights in the training loss alone do not help, as Proposition~\ref{prop:sampling}(iv) states. We put the
design-unbiased weights $w_x/\pi_x$ into the loss. The whole-mesh error of SCANO with its anchors outside the message-passing
layers then rises from 0.118 to 0.140. We also weighted the state of the SCANO backbone + BQA by $w_x/\pi_x$. Its
whole-mesh error changes little (0.116 against 0.109). Its error on a one-hop regional subgraph of 5\% of the cells
stays at 0.24 (0.241 against 0.238). The reason is that the state still follows the sample.

\subsection{Where the gain of BQA comes from}
\label{supp:r7-design}

The gain of BQA rests on reading the records at every point, not through one summary. To test this, we built a global
code from exactly the records that BQA reads. The code is pooled in every layer by attention with learned queries and
broadcast to all points. It reaches a whole-mesh error of 0.48 on the controlled benchmark. This is about four times
that of SCANO. The point dependence of the readout therefore carries the gain, not the records alone (C4).
Reading the records only in the first layer of Transolver keeps almost all of the gain of BQA. The whole-mesh error is
0.195, against 0.189 when the records are read in every layer and 0.80 without BQA. This variant has a quarter fewer
parameters (Table~\ref{tab:r7-design}). The gain thus lies in area-weighted records read at every point, not in reading
them repeatedly. The LFE encoder with area weights reads the same records at the input. It is as accurate on this
benchmark and reaches 0.167. Attaching AGS to it leaves this unchanged (0.172). Transolver + BQA reaches 0.166 at the
larger of the two learning rates of the selection (Table~\ref{tab:r7-capacity}). Both area-weighted readouts thus agree
on the controlled benchmark. There, every record is a face of a small patch. BQA brings the same area-weighted reading
to every backbone. It does so inside the attention layers and on all three data sets.

\begin{table}[!htb]
\centering
\footnotesize
\setlength{\tabcolsep}{3pt}
\caption{Controlled benchmark, $k$-hop samples at ratio 0.12: where the records are read. The table gives the relative $L^2$ error and the response error $E_{\mathrm{resp}}$, both on the whole mesh. Values are seed means $\pm$ half the width of the 95\% case-bootstrap interval.}
\label{tab:r7-design}
\begin{tabular}{P{0.46\linewidth}ccc}
\toprule
Method & Parameters ($10^3$) & Whole mesh & $E_{\mathrm{resp}}$ \\
\midrule
Transolver & 803 & 0.800\,{\scriptsize$\pm$\,0.045} & 0.808\,{\scriptsize$\pm$\,0.075} \\
Transolver + BQA, read in every layer & 800 & 0.189\,{\scriptsize$\pm$\,0.012} & 0.299\,{\scriptsize$\pm$\,0.051} \\
Transolver + BQA, read in the first layer only & 601 & 0.195\,{\scriptsize$\pm$\,0.011} & 0.314\,{\scriptsize$\pm$\,0.063} \\
Transolver + LFE, area & 778 & 0.167\,{\scriptsize$\pm$\,0.014} & 0.276\,{\scriptsize$\pm$\,0.075} \\
Transolver + LFE, area, + AGS & 778 & 0.172\,{\scriptsize$\pm$\,0.016} & 0.274\,{\scriptsize$\pm$\,0.064} \\
\bottomrule
\end{tabular}
\end{table}

A second way to isolate the readout is to remove the sampling altogether. The public benchmark allows this, because its
cases fit in memory. Trained on the whole mesh, the original backbones reach the accuracy of their sampled training.
They do not reach that of the same backbones with BQA. Transolver reaches a whole-mesh error of 46.6\%, against 46.9\%
on samples and 3.1\% with BQA. AB-UPT reaches 2.9\%, against 2.8\% on samples with whole-mesh anchors. AB-UPT + BQA +
AGS reaches 1.7\%. On this benchmark, the gain of BQA is therefore the gain of the boundary readout. On the controlled
benchmark, BQA replaces GNOT's own attention to the boundary records. Placed in the same way here, BQA lowers GNOT's
error to 3.1\%. Added beside that attention, it gives 5.8\%. GNOT alone reaches 9.0\%.

On DCTS, Transolver fits on a whole room. The same separation between the two modules holds there. Trained on samples
instead of whole rooms, the original Transolver's error rises by about a third. That of Transolver + BQA rises by about
three per cent. That of Transolver + BQA + AGS does not rise (Table~\ref{tab:dcts-tax}). The boundary readout thus
closes most of the sampling gap. The sample-independent state closes the rest.

The number of slice tokens sets how finely the boundary is resolved. The patches of the controlled benchmark are small
and adjacent. There, compressing the per-face records into 64 area-consistent slice tokens keeps most of the gain of BQA
on Transolver. The whole-mesh error is 0.31, against 0.19 with one token per face and 0.80 without BQA. Eight tokens
keep less. The loss concentrates near the walls and on the smallest patches, as Proposition~\ref{prop:unmix} predicts.
On DCTS, accuracy and response do not change from 16 to 1,024 learned slices (Table~\ref{tab:dcts-budget}). The value of
the main text, 64, was set before this sweep. The runs and criteria of the sweep were registered before any of them was
trained. The learned soft assignment never isolates a single rack on DCTS. The outlet records of a doubled rack make up
less than one per cent of the mass of their slice, for every $G$. We therefore also gave the server outlets 256 slices
of their own. This isolates every rack by construction. The error, the regional error and the response to a doubled rack
still do not change. This holds with and without the per-point heat-source input (Table~\ref{tab:r7-outlet}).
GeoTransolver reads every boundary record in every layer. Without that input, it shows the same response
(Table~\ref{tab:r7-heat}). The resolution at which the records are read therefore does not set the rack response on
DCTS. Each point receives as input the load of its nearest server and its offset to that exhaust. The response to a
single rack reaches the domain through this input, which every model shares.

\begin{table}[!htb]
\centering
\footnotesize
\setlength{\tabcolsep}{3pt}
\caption{DCTS: the outlet records of the servers in 256 slices of their own, with and without the per-point heat-source input (50 test rooms, seed means). The table gives the error in the model's own deployment and its ratio to the row above, with the 95\% interval over the test rooms. It gives the error on one region draw of the training sampler at the training budget. Such a draw fills a box of random size and tops it up with seeds from the whole room. It differs from the compact regions of Table~\ref{tab:dcts-main}. It also gives the response error $E_{\mathrm{resp}}$ of the room temperature to one rack's doubled power, with its ratio and the 95\% interval over the held-out pairs.}
\label{tab:r7-outlet}
\begin{tabular}{P{0.34\linewidth}ccccc}
\toprule
 & \multicolumn{2}{c}{Error} & Regional & \multicolumn{2}{c}{$E_{\mathrm{resp}}$, rack doubled} \\
\cmidrule(lr){2-3}\cmidrule(lr){5-6}
Method & Value & Ratio & error & Value & Ratio \\
\midrule
SCANO without heat-source input & 0.407 & --- & 0.443 & 0.86 & --- \\
\quad with server-outlet slices & 0.415 & 1.02\,{\scriptsize[1.00, 1.05]} & 0.448 & 0.88 & 1.02\,{\scriptsize[0.99, 1.14]} \\
\midrule
SCANO & 0.395 & --- & 0.425 & 0.64 & --- \\
\quad with server-outlet slices & 0.398 & 1.01\,{\scriptsize[1.00, 1.01]} & 0.428 & 0.66 & 1.03\,{\scriptsize[0.99, 1.08]} \\
\bottomrule
\end{tabular}
\end{table}

On DCTS, the two readout paths and the measures behind them matter less for the whole-room error. These ablations
start from SCANO with its anchors outside the message-passing layers. Removing the unnormalised readout leaves its whole-room
error unchanged (ratio 1.003). Its response to a doubled rack stays
nearly the same (0.62 against 0.61). Weighting the server records by area instead of by their heat flux changes the
error as little (1.002). The rack response worsens slightly (0.64). We also conditioned the servers only through their records,
without the per-point heat-source input. The error then rises by about seven per cent, and the rack response error rises
to 0.83. On DCTS, the rack-level information that a doubled rack needs thus reaches the domain mainly through that
input. The slice analysis above found the same.

The per-point heat-source input plays the same role for every model on DCTS. Without it, the error in each model's own
deployment rises by five to ten per cent. The response error to a doubled rack rises from 0.6--0.7 to 0.83--0.87. This
holds for Transolver, Transolver + BQA, GeoTransolver and SCANO with anchors outside the message-passing layers alike
(Table~\ref{tab:r7-heat}). The response to a
halved cooling flow changes little. The boundary values of the cooling units carry that response. The rack-level
information therefore reaches the domain through the heat-source input, which every model shares. This also explains why
BQA adds little to GeoTransolver on DCTS. GeoTransolver already reads every boundary record in each layer. The two
benchmarks have no such input. There, the boundary records carry the whole forcing, and BQA helps every backbone.

\begin{table}[!htb]
\centering
\footnotesize
\setlength{\tabcolsep}{3pt}
\caption{DCTS models with and without the per-point heat-source input (50 test rooms, seed means). The table gives the error in the model's own deployment. It also gives the response error $E_{\mathrm{resp}}$ of the room temperature to one rack's doubled power and to one cooling unit's halved flow, as the median over the ten held-out pairs of each kind.}
\label{tab:r7-heat}
\begin{tabular}{P{0.34\linewidth}cccccc}
\toprule
 & \multicolumn{2}{c}{Error} & \multicolumn{2}{c}{Rack power doubled} & \multicolumn{2}{c}{Cooling flow halved} \\
\cmidrule(lr){2-3}\cmidrule(lr){4-5}\cmidrule(lr){6-7}
Method & with & without & with & without & with & without \\
\midrule
Transolver & 0.457 & 0.478 & 0.72 & 0.84 & 0.85 & 0.79 \\
Transolver + BQA & 0.410 & 0.448 & 0.73 & 0.85 & 0.38 & 0.38 \\
GeoTransolver & 0.392 & 0.419 & 0.65 & 0.87 & 0.37 & 0.42 \\
SCANO, anchors outside the message passing & 0.397 & 0.423 & 0.61 & 0.83 & 0.39 & 0.45 \\
\bottomrule
\end{tabular}
\end{table}

Weighting the records of BQA by count instead of area changes SCANO's error on DCTS by about one per cent. Its response
errors do not change. On the controlled benchmark, count weighting costs accuracy mainly when the deployment mesh is
graded differently from the training mesh (Section~\ref{sec:test-c2}; Table~\ref{tab:ablations}). The DCTS test rooms
contain no such change. They are meshed by the same procedure as the training rooms. Removing BQA from the SCANO
backbone with AGS has a large effect. The error rises by about half (ratio 1.48, interval [1.31, 1.66] over the test
rooms). The response to a halved cooling flow disappears ($E_{\mathrm{resp}}=1.00$). On DCTS too, the boundary records
that BQA reads carry the accuracy of the field. They also carry the response to what the cooling units impose.

\begin{table}[!htb]
\centering
\footnotesize
\setlength{\tabcolsep}{3pt}
\caption{DCTS ablations (50 test rooms, seed means). The table gives the error in each model's own deployment and its ratio to the reference row above it, with the 95\% interval over the test rooms. It also gives the response error $E_{\mathrm{resp}}$ of the room temperature to one rack's doubled power and to one cooling unit's halved flow, as the median over the ten held-out pairs of each kind. $^{a}$One run stayed on the initial loss plateau and is left out of the mean, as in the main tables.}
\label{tab:r7-dcts-ablation}
\begin{tabular}{P{0.36\linewidth}ccccc}
\toprule
 & Parameters & & Ratio to & \multicolumn{2}{c}{$E_{\mathrm{resp}}$} \\
\cmidrule(lr){5-6}
Method & (M) & Error & reference & Rack doubled & Flow halved \\
\midrule
SCANO & 0.81 & 0.395 & --- & 0.64 & 0.40 \\
SCANO, BQA weighted by count$^{a}$ & 0.81 & 0.399 & 1.01\,{\scriptsize[1.01, 1.02]} & 0.64 & 0.39 \\
SCANO backbone + AGS, without BQA & 0.62 & 0.584 & 1.48\,{\scriptsize[1.31, 1.66]} & 0.66 & 1.00 \\
\midrule
Transolver & 0.81 & 0.457 & --- & 0.72 & 0.85 \\
Transolver, about 3 M parameters & 3.04 & 0.419 & 0.92\,{\scriptsize[0.90, 0.94]} & 0.69 & 0.94 \\
\midrule
Transolver + BQA & 0.86 & 0.410 & --- & 0.73 & 0.38 \\
Transolver + BQA, about 3 M parameters$^{a}$ & 2.98 & 0.395 & 0.96\,{\scriptsize[0.96, 0.97]} & 0.67 & 0.37 \\
\bottomrule
\end{tabular}
\end{table}

\subsection{Model capacity and learning rate}
\label{supp:r7-capacity}

We also trained larger models. At 1.37 million parameters, about 1.7 times the common budget, Transolver + BQA improves
from 0.19 to 0.17 on the controlled benchmark. Transolver + LFE improves from 0.21 to 0.20. SCANO with anchors outside the message-passing layers stays at 0.12 to 0.13.
The ranking of the methods therefore does not follow from the budget.
At about three million parameters, four times the common budget, the picture is the same (Table~\ref{tab:r7-capacity}).
The original Transolver and AB-UPT with whole-mesh anchors by count do not improve (0.84 and 0.25, against 0.80 and 0.26). Transolver + BQA and AB-UPT +
BQA + AGS stay where they were (0.18 and 0.076). SCANO reaches 0.148. It was trained at the learning rate selected for
its main width, without a new selection. None of the comparisons is therefore set by the parameter budget. We also tried
two further learning rates. The error of every method is then at least that of the selected rate.
The one exception is Transolver + BQA, which gains from the larger rate (0.17 against 0.19).

\begin{table}[!htb]
\centering
\scriptsize
\setlength{\tabcolsep}{2.5pt}
\caption{Controlled benchmark, $k$-hop samples at ratio 0.12. The table gives the whole-mesh relative $L^2$ error at the common budget of the main text and at about three million parameters. The wider models use the main learning rate. These values are seed means $\pm$ half the width of the 95\% case-bootstrap interval. The last column gives the error at two further learning rates. $^{a}$One run stayed on the initial loss plateau and is left out of the mean, as in the main tables. $^{b}$The run stayed on the initial loss plateau.}
\label{tab:r7-capacity}
\begin{tabular}{P{0.25\linewidth}ccccP{0.24\linewidth}}
\toprule
 & \multicolumn{2}{c}{Common budget} & \multicolumn{2}{c}{About 3 M parameters} & \\
\cmidrule(lr){2-3}\cmidrule(lr){4-5}
Method & Parameters (M) & Whole mesh & Parameters (M) & Whole mesh & Learning rate: whole mesh \\
\midrule
Transolver & 0.80 & 0.800\,{\scriptsize$\pm$\,0.045} & 3.02 & 0.840\,{\scriptsize$\pm$\,0.045}$^{a}$ & 3e-4 (main); 1e-3: 0.890$^{b}$; 1e-4: 0.810 \\
Transolver + BQA & 0.80 & 0.189\,{\scriptsize$\pm$\,0.012} & 2.97 & 0.181\,{\scriptsize$\pm$\,0.010} & 3e-4 (main); 1e-3: 0.166; 1e-4: 0.309 \\
AB-UPT, whole-mesh anchors by count & 0.82 & 0.256\,{\scriptsize$\pm$\,0.029} & 2.92 & 0.250\,{\scriptsize$\pm$\,0.026} & 3e-4 (main) \\
AB-UPT + BQA + AGS & 0.76 & 0.073\,{\scriptsize$\pm$\,0.009} & 3.02 & 0.076\,{\scriptsize$\pm$\,0.007} & 3e-4 (main); 1e-3: 0.075; 1e-4: 0.081 \\
SCANO & 0.78 & 0.111\,{\scriptsize$\pm$\,0.009} & 3.08 & 0.148\,{\scriptsize$\pm$\,0.009} & 1e-3 (main); 3e-4: 0.141; 1e-4: 0.238 \\
\bottomrule
\end{tabular}
\end{table}

On DCTS, the larger budget brings the error of the original Transolver closer to that of Transolver + BQA at the common
budget (0.419 against 0.410). It does not bring its response to a halved cooling flow closer. That response error stays
near one (0.94 against 0.38). Transolver + BQA itself gains a few per cent from the same budget. It then reaches SCANO's
error (0.395; Table~\ref{tab:r7-dcts-ablation}). On the industrial data too, the boundary information that the cooling
units impose comes from BQA, not from capacity.

\subsection{Deployment protocols}
\label{supp:r7-deploy}

The comparisons in the main text evaluate every model in the deployment regime defined by its training
(Section~\ref{sec:setup-deploy}). A model with AGS predicts the whole mesh in one pass. Its state is computed once for
the case. A model whose state follows the sample predicts the mesh as the average over training-sized samples that cover
it. Each of its points is thus predicted from a context of the kind it was trained on. On DCTS, this is the column ``Own
deployment'' of Table~\ref{tab:dcts-main}. The adjacent ``Whole mesh'' column evaluates every model in a single pass.
Table~\ref{tab:r7-stats} compares Transolver + BQA with Transolver in both deployments.

The same choice of deployment applies to the response. Table~\ref{tab:r0s-main-resp} measures it on one sample drawn as
in training. There, a model meets the incomplete boundary information it was trained with. We also evaluated the response on
the whole mesh, where every model sees every boundary face. On the controlled benchmark, the response error of every
model then stays within a few per cent of its value on a sample. The order of the methods does not change
(Table~\ref{tab:r7-resp-whole}). Transolver and AeroGTO, whose state follows the sample, gain about five per cent from
the complete boundary. Their response errors remain two to three times those of the same backbones with BQA. The gain of
BQA in the response is therefore not an artefact of evaluating on a subsample.

\begin{table}[!htb]
\centering
\footnotesize
\setlength{\tabcolsep}{4pt}
\caption{Controlled benchmark, $k$-hop samples at ratio 0.12. The table gives the response error $E_{\mathrm{resp}}$ of the models of Table~\ref{tab:r0s-main} on one sample drawn as in training and on the whole mesh. The first column repeats the column of Table~\ref{tab:r0s-main-resp}. The pairs are those whose true response exceeds five percent of the base solution: all pairs, and the pairs on the smallest patches. Values are seed means $\pm$ half the width of the 95\% interval over whole base cases.}
\label{tab:r7-resp-whole}
\begin{tabular}{P{0.40\linewidth}ccc}
\toprule
 & Sample & \multicolumn{2}{c}{Whole mesh} \\
\cmidrule(lr){3-4}
Method & All & All & Smallest patches \\
\midrule
SCANO & 0.169\,{\scriptsize$\pm$\,0.025} & 0.168\,{\scriptsize$\pm$\,0.025} & 0.191\,{\scriptsize$\pm$\,0.037} \\
SCANO backbone + BQA & 0.176\,{\scriptsize$\pm$\,0.024} & 0.171\,{\scriptsize$\pm$\,0.024} & 0.203\,{\scriptsize$\pm$\,0.045} \\
Transolver + LFE & 0.299\,{\scriptsize$\pm$\,0.061} & 0.300\,{\scriptsize$\pm$\,0.063} & 0.346\,{\scriptsize$\pm$\,0.056} \\
Transolver + LFE, area & 0.276\,{\scriptsize$\pm$\,0.073} & 0.276\,{\scriptsize$\pm$\,0.075} & 0.307\,{\scriptsize$\pm$\,0.047} \\
GNOT & 0.410\,{\scriptsize$\pm$\,0.082} & 0.404\,{\scriptsize$\pm$\,0.082} & 0.464\,{\scriptsize$\pm$\,0.068} \\
GeoTransolver & 0.716\,{\scriptsize$\pm$\,0.110} & 0.712\,{\scriptsize$\pm$\,0.112} & 0.718\,{\scriptsize$\pm$\,0.088} \\
Slice aggregation & 0.540\,{\scriptsize$\pm$\,0.120} & 0.540\,{\scriptsize$\pm$\,0.119} & 0.570\,{\scriptsize$\pm$\,0.076} \\
Transolver & 0.850\,{\scriptsize$\pm$\,0.075} & 0.808\,{\scriptsize$\pm$\,0.075} & 0.823\,{\scriptsize$\pm$\,0.068} \\
Transolver + BQA & 0.297\,{\scriptsize$\pm$\,0.050} & 0.299\,{\scriptsize$\pm$\,0.051} & 0.353\,{\scriptsize$\pm$\,0.058} \\
AB-UPT, whole-mesh anchors by count & 0.347\,{\scriptsize$\pm$\,0.070} & 0.344\,{\scriptsize$\pm$\,0.070} & 0.381\,{\scriptsize$\pm$\,0.080} \\
AB-UPT + BQA + AGS & 0.121\,{\scriptsize$\pm$\,0.024} & 0.120\,{\scriptsize$\pm$\,0.024} & 0.156\,{\scriptsize$\pm$\,0.046} \\
AeroGTO & 0.767\,{\scriptsize$\pm$\,0.072} & 0.724\,{\scriptsize$\pm$\,0.074} & 0.751\,{\scriptsize$\pm$\,0.066} \\
AeroGTO + BQA & 0.315\,{\scriptsize$\pm$\,0.120} & 0.312\,{\scriptsize$\pm$\,0.122} & 0.276\,{\scriptsize$\pm$\,0.042} \\
\bottomrule
\end{tabular}
\end{table}

For the regional deployments, a region of interest is an axis-aligned box centred on a cell drawn uniformly at random.
Its sides start as the extents of the mesh along the three axes. Each side is multiplied by a factor drawn log-uniformly
between one half and two. The box is then scaled by the smallest common factor at which it holds the prescribed share of
the cells. The region contains every cell and boundary face inside the box and nothing outside. The message-passing
models also read a halo as deep as their receptive field. Every method is evaluated on the same three regions per case and coverage share.
A model whose state follows the sample could also be deployed on a region with a state computed once from the whole mesh
of the case. For a backbone without message passing, this reproduces the whole-mesh prediction inside the region
exactly. The region columns of the main tables instead compute the state from the region itself. They therefore measure
the deployment that a regional query is meant to provide, one that does not pass over the whole mesh. The cached
alternative restores the whole-mesh accuracy only at the cost of that pass for every case. A sample-independent state
avoids this cost by construction. We also compared the regional error with the whole-mesh error on the same cells. On
the controlled benchmark at a share of 0.02, the error of the original Transolver is then 2.8 times larger. That of the
SCANO backbone + BQA is 5.1 times larger. SCANO and Transolver + BQA + AGS keep their whole-mesh accuracy within one per
cent.

AB-UPT can also be deployed over the whole room at a lower cost. Its whole-mesh results on DCTS
(Table~\ref{tab:dcts-main}) use every cell as an anchor. In that deployment, the quadratic anchor attention takes under
three minutes per room (Section~\ref{sec:dcts-cost}). The anchors of AGS do not depend on the sample. The same
trained models can therefore predict the whole room using a single fresh draw of anchors at the training count. The cost is
then linear in the number of cells. The pass takes about two seconds, three times SCANO's single pass in the timing
of Section~\ref{sec:dcts-cost}. The whole-room error rises by about one per cent for AB-UPT + BQA + AGS and two
per cent for AB-UPT + AGS. SCANO's error changes by less than one per cent. The sample-independent state thus gives
AB-UPT, too, a deployment of linear cost at nearly the accuracy of its quadratic one.

\subsection{Further evidence on DCTS}
\label{supp:r7-dcts}

Beyond the field error, the predictions can be checked against conservation. The reference solutions close the energy
balance of a room to within a fraction of a per cent (Section~\ref{supp:r7-reference}). The same boundary integral can
be evaluated on a model's whole-room prediction. It uses the predicted velocity and temperature on every boundary face.
None of the compared models is trained with a conservation constraint. Their predicted fields close the balance to about
14 to 20 per cent in the median test room. The methods differ little (Table~\ref{tab:r7-energy}). The imbalance enters
mainly through the velocities predicted on the boundary faces. Their volume fluxes miss the mass balance by 7 to 16 per
cent. The miss is largest at the cooling-unit returns. There the solver prescribes the pressure, and the velocity
follows the outflow. Scaling the predicted return fluxes to carry the supplied flow removes only a small part of the
imbalance. The temperatures themselves carry the heat far more accurately. We weight the predicted return temperatures
by the face fluxes of the reference solution. This gives the heat removed by the cooling units to within 4 to 7 per cent
in the median room. The one exception is AB-UPT without BQA, at 12 per cent. BQA lowers this error on Transolver from
about 7 to 5 per cent. On AB-UPT, it lowers it from 12 to 6 per cent. A conservative output is a natural extension for
applications that need the global balance of the predicted field. One example is projecting the boundary fluxes onto the
prescribed flows.

\begin{table}[!htb]
\centering
\footnotesize
\setlength{\tabcolsep}{3pt}
\caption{Global balances of the predicted DCTS fields over the 50 test rooms. They are computed from the volume and enthalpy fluxes through the boundary faces of each model's whole-room prediction in its own deployment. The first column is the relative imbalance between the heat removed by the cooling units and the heat added by the servers. The second is the net volume flux relative to the supplied flow. The third is the relative error of the heat removed by the cooling units, computed from the predicted temperatures at the face fluxes of the reference solution. Values are medians over the rooms and means over the seeds, with the range over the seeds in brackets.}
\label{tab:r7-energy}
\begin{tabular}{P{0.40\linewidth}ccc}
\toprule
Method & Energy balance & Mass balance & Heat removed, known flows \\
\midrule
\multicolumn{4}{l}{\emph{SCANO}} \\
SCANO & 0.164\,{\scriptsize[0.139, 0.208]} & 0.102\,{\scriptsize[0.085, 0.122]} & 0.057\,{\scriptsize[0.035, 0.072]} \\
SCANO, anchors outside the message passing & 0.157\,{\scriptsize[0.126, 0.205]} & 0.127\,{\scriptsize[0.094, 0.152]} & 0.054\,{\scriptsize[0.028, 0.079]} \\
SCANO backbone + BQA & 0.156\,{\scriptsize[0.122, 0.187]} & 0.110\,{\scriptsize[0.100, 0.119]} & 0.054\,{\scriptsize[0.033, 0.073]} \\
\midrule
\multicolumn{4}{l}{\emph{Backbones without and with BQA}} \\
Transolver & 0.165\,{\scriptsize[0.101, 0.209]} & 0.158\,{\scriptsize[0.090, 0.225]} & 0.071\,{\scriptsize[0.050, 0.099]} \\
Transolver + BQA & 0.147\,{\scriptsize[0.122, 0.168]} & 0.114\,{\scriptsize[0.087, 0.129]} & 0.045\,{\scriptsize[0.035, 0.058]} \\
AB-UPT + AGS & 0.177\,{\scriptsize[0.125, 0.205]} & 0.120\,{\scriptsize[0.106, 0.148]} & 0.122\,{\scriptsize[0.108, 0.137]} \\
AB-UPT + BQA + AGS & 0.198\,{\scriptsize[0.167, 0.242]} & 0.068\,{\scriptsize[0.066, 0.071]} & 0.062\,{\scriptsize[0.037, 0.080]} \\
AeroGTO & 0.174\,{\scriptsize[0.154, 0.195]} & 0.119\,{\scriptsize[0.081, 0.157]} & 0.066\,{\scriptsize[0.043, 0.099]} \\
\midrule
\multicolumn{4}{l}{\emph{Other boundary encoders}} \\
Transolver + LFE & 0.185\,{\scriptsize[0.162, 0.203]} & 0.106\,{\scriptsize[0.076, 0.163]} & 0.067\,{\scriptsize[0.050, 0.079]} \\
GeoTransolver & 0.142\,{\scriptsize[0.132, 0.157]} & 0.106\,{\scriptsize[0.096, 0.125]} & 0.039\,{\scriptsize[0.031, 0.053]} \\
GNOT & 0.179\,{\scriptsize[0.156, 0.197]} & 0.139\,{\scriptsize[0.115, 0.175]} & 0.059\,{\scriptsize[0.053, 0.066]} \\
Slice aggregation & 0.171\,{\scriptsize[0.146, 0.193]} & 0.111\,{\scriptsize[0.100, 0.131]} & 0.057\,{\scriptsize[0.047, 0.072]} \\
\midrule
CFD reference & 0.006 & 0.011 & --- \\
\bottomrule
\end{tabular}
\end{table}

Table~\ref{tab:c132-tta} gives the error of the multi-pass average at each share of the room per pass.
Figure~\ref{fig:cost}f gives its time. On these four rooms, the error of Transolver without BQA rises steeply as one pass covers more of
the room. With BQA, every model keeps its error within a few percent at every share. Figure~\ref{fig:cost-gpus}
(Section~\ref{supp:figures}) deals the passes of the average to several GPUs.

\begin{table}[!htb]
\centering
\footnotesize
\setlength{\tabcolsep}{4pt}
\caption{Error of the multi-pass average against the share of the room per pass on DCTS. Each pass draws seed cells at the ratio $\rho$ and grows them by two hops. The second header row gives the share of the room in a full pass, counting its cells and boundary faces. The values are the relative error averaged over the five fields. Each is the median over the four rooms of Table~\ref{tab:dcts-time}. The training setting is $\rho = 0.05$ (bold). At $\rho = 1$, one pass covers the whole room. Figure~\ref{fig:cost}f gives the time of the same averages.}
\label{tab:c132-tta}
\begin{tabular}{lccccccc}
\toprule
Seed ratio $\rho$ & 0.01 & 0.02 & \textbf{0.05} & 0.1 & 0.2 & 0.5 & 1 \\
Share of the room per pass & 0.07 & 0.13 & 0.27 & 0.42 & 0.59 & 0.80 & 1.00 \\
\midrule
\multicolumn{8}{l}{\emph{State that follows the sample}} \\
SCANO backbone + BQA & 0.394 & 0.392 & 0.390 & 0.390 & 0.392 & 0.394 & 0.395 \\
Transolver & 0.443 & 0.441 & 0.438 & 0.438 & 0.469 & 0.557 & 0.840 \\
Transolver + BQA & 0.423 & 0.423 & 0.423 & 0.424 & 0.425 & 0.430 & 0.446 \\
GeoTransolver & 0.404 & 0.404 & 0.404 & 0.403 & 0.403 & 0.404 & 0.408 \\
\midrule
\multicolumn{8}{l}{\emph{Anchored global state}} \\
SCANO & 0.399 & 0.397 & 0.394 & 0.395 & 0.396 & 0.395 & 0.399 \\
Transolver + BQA + AGS & 0.439 & 0.439 & 0.438 & 0.438 & 0.437 & 0.439 & 0.442 \\
AB-UPT + AGS & 0.389 & 0.388 & 0.390 & 0.388 & 0.393 & 0.408 & 0.394 \\
AB-UPT + BQA + AGS & 0.381 & 0.380 & 0.381 & 0.381 & 0.381 & 0.382 & 0.383 \\
\bottomrule
\end{tabular}
\end{table}

\subsection{Statistics of the main comparisons}
\label{supp:r7-stats}

Every configuration is trained with multiple seeds. The samples of a training run depend only on the case and the epoch.
The seeds therefore differ only in their initialisation and in the order of the minibatches. The intervals of the main
tables resample the test cases. Table~\ref{tab:r7-stats} adds the paired ratio for the comparisons on which the text
relies. Its interval also resamples the training seeds of each model, so the variability between seeds enters it. Every
gain and gap that the text reports stays on its side of one under this wider interval. The one exception is the small
gain of BQA on AB-UPT on DCTS, which lies inside it. The comparison of SCANO with the
SCANO backbone + BQA on the controlled benchmark remains a tie. On DCTS, SCANO and GeoTransolver differ by less than one
per cent. This difference lies at the edge of the case interval and inside the wider one.

\begin{table}[!htb]
\centering
\footnotesize
\setlength{\tabcolsep}{3pt}
\caption{Paired comparisons behind the main text. Each ratio is the error of the first model over that of the second. On the
controlled benchmark, the errors are evaluated on the whole mesh. On DCTS, each model is evaluated in its own deployment unless
noted. The 95\% intervals come from 10,000 bootstrap draws over the test cases. In the last column, the draws also
resample the training seeds of each model.}
\label{tab:r7-stats}
\begin{tabular}{P{0.56\linewidth}ccc}
\toprule
Comparison & Ratio & Cases & Cases and seeds \\
\midrule
\multicolumn{4}{l}{\emph{Controlled benchmark, $k$-hop samples}} \\
Transolver + BQA / Transolver & 0.24 & {\scriptsize[0.22, 0.25]} & {\scriptsize[0.22, 0.25]} \\
AB-UPT + BQA + AGS / AB-UPT, whole-mesh anchors by count & 0.29 & {\scriptsize[0.26, 0.32]} & {\scriptsize[0.25, 0.33]} \\
SCANO / Transolver + LFE, area & 0.66 & {\scriptsize[0.63, 0.70]} & {\scriptsize[0.61, 0.72]} \\
SCANO / SCANO backbone + BQA & 1.01 & {\scriptsize[0.98, 1.05]} & {\scriptsize[0.90, 1.12]} \\
SCANO / AB-UPT + BQA + AGS & 1.50 & {\scriptsize[1.34, 1.66]} & {\scriptsize[1.32, 1.70]} \\
Transolver + BQA / Transolver + LFE, area & 1.13 & {\scriptsize[1.05, 1.21]} & {\scriptsize[1.04, 1.24]} \\
\midrule
\multicolumn{4}{l}{\emph{DCTS}} \\
Transolver + BQA / Transolver & 0.90 & {\scriptsize[0.87, 0.92]} & {\scriptsize[0.86, 0.94]} \\
Transolver + BQA / Transolver, one pass over the room & 0.78 & {\scriptsize[0.75, 0.81]} & {\scriptsize[0.67, 0.88]} \\
AB-UPT + BQA + AGS / AB-UPT + AGS & 0.99 & {\scriptsize[0.97, 1.01]} & {\scriptsize[0.95, 1.02]} \\
SCANO / GeoTransolver$^{\ast}$ & 1.006 & {\scriptsize[1.000, 1.013]} & {\scriptsize[0.998, 1.015]} \\
SCANO / SCANO backbone + BQA & 1.02 & {\scriptsize[1.02, 1.03]} & {\scriptsize[1.01, 1.03]} \\
SCANO / AB-UPT + BQA + AGS & 1.06 & {\scriptsize[1.04, 1.07]} & {\scriptsize[1.03, 1.08]} \\
\bottomrule
\end{tabular}
\par\smallskip\footnotesize
$^{\ast}$Three decimals: the case interval ends at one, and the wider interval contains it.
\end{table}

The engineering readings of Table~\ref{tab:dcts-decision} rest on fewer samples. There are ten held-out pairs of each
kind of intervention and 50 rooms for the server inlets. Table~\ref{tab:r7-decision-ci} gives their intervals. With ten
pairs, the interval of a response error spans 0.1 to 0.25. The table therefore resolves the separation that the text
draws from it. Under a halved cooling flow, the models with BQA reach about 0.4 and those without about 0.85. These
intervals lie far apart. The models with BQA differ among themselves by a few hundredths. They remain within one
another's intervals for both kinds of intervention. Over the 50 rooms, the two variants of AB-UPT predict the server-inlet temperatures most accurately. Their intervals of the rank
correlation and of the mean absolute error barely touch those of the next models. The readings of the other
models lie within one another's intervals.

\begin{table}[!htb]
\centering
\footnotesize
\setlength{\tabcolsep}{3pt}
\caption{The readings of Table~\ref{tab:dcts-decision} with 95\% bootstrap intervals. For the response error, the intervals come from 10,000 draws of the ten held-out pairs of each kind; each reading is the median over the pairs, averaged over the seeds. For the rank correlation and the mean absolute error of the server-inlet temperatures, they come from 10,000 draws of the 50 test rooms.}
\label{tab:r7-decision-ci}
\begin{tabular}{P{0.30\linewidth}cccc}
\toprule
 & \multicolumn{2}{c}{Response $E_{\mathrm{resp}}$} & \multicolumn{2}{c}{Server inlets} \\
\cmidrule(lr){2-3}\cmidrule(lr){4-5}
Method & Rack power doubled & Cooling flow halved & Rank correlation & MAE (K) \\
\midrule
\multicolumn{5}{l}{\emph{SCANO}} \\
SCANO & 0.643\,{\scriptsize[0.593, 0.759]} & 0.400\,{\scriptsize[0.273, 0.520]} & 0.54\,{\scriptsize[0.47, 0.59]} & 1.40\,{\scriptsize[1.20, 1.59]} \\
SCANO, anchors outside the message passing & 0.612\,{\scriptsize[0.584, 0.732]} & 0.393\,{\scriptsize[0.297, 0.522]} & 0.55\,{\scriptsize[0.49, 0.60]} & 1.39\,{\scriptsize[1.19, 1.58]} \\
SCANO backbone + BQA & 0.633\,{\scriptsize[0.598, 0.742]} & 0.386\,{\scriptsize[0.274, 0.513]} & 0.55\,{\scriptsize[0.49, 0.60]} & 1.38\,{\scriptsize[1.19, 1.58]} \\
\midrule
\multicolumn{5}{l}{\emph{Backbones without and with BQA}} \\
Transolver & 0.718\,{\scriptsize[0.648, 0.830]} & 0.854\,{\scriptsize[0.795, 0.925]} & 0.47\,{\scriptsize[0.41, 0.53]} & 1.53\,{\scriptsize[1.34, 1.71]} \\
Transolver + BQA & 0.734\,{\scriptsize[0.644, 0.807]} & 0.376\,{\scriptsize[0.268, 0.522]} & 0.49\,{\scriptsize[0.43, 0.55]} & 1.45\,{\scriptsize[1.24, 1.65]} \\
AB-UPT + AGS & 0.615\,{\scriptsize[0.588, 0.748]} & 0.860\,{\scriptsize[0.816, 0.906]} & 0.66\,{\scriptsize[0.59, 0.73]} & 1.07\,{\scriptsize[0.93, 1.20]} \\
AB-UPT + BQA + AGS & 0.608\,{\scriptsize[0.559, 0.733]} & 0.399\,{\scriptsize[0.327, 0.532]} & 0.66\,{\scriptsize[0.59, 0.72]} & 1.04\,{\scriptsize[0.90, 1.17]} \\
AeroGTO & 0.766\,{\scriptsize[0.678, 0.832]} & 0.849\,{\scriptsize[0.798, 0.885]} & 0.52\,{\scriptsize[0.47, 0.57]} & 1.60\,{\scriptsize[1.43, 1.75]} \\
\midrule
\multicolumn{5}{l}{\emph{Other boundary encoders}} \\
Transolver + LFE & 0.658\,{\scriptsize[0.622, 0.747]} & 0.587\,{\scriptsize[0.518, 0.701]} & 0.47\,{\scriptsize[0.41, 0.53]} & 1.46\,{\scriptsize[1.25, 1.67]} \\
GeoTransolver & 0.653\,{\scriptsize[0.621, 0.772]} & 0.371\,{\scriptsize[0.294, 0.535]} & 0.48\,{\scriptsize[0.42, 0.54]} & 1.43\,{\scriptsize[1.22, 1.63]} \\
GNOT & 0.746\,{\scriptsize[0.627, 0.808]} & 0.412\,{\scriptsize[0.345, 0.543]} & 0.45\,{\scriptsize[0.38, 0.51]} & 1.53\,{\scriptsize[1.31, 1.74]} \\
Slice aggregation & 0.661\,{\scriptsize[0.643, 0.761]} & 0.497\,{\scriptsize[0.439, 0.613]} & 0.48\,{\scriptsize[0.41, 0.54]} & 1.47\,{\scriptsize[1.27, 1.67]} \\
\bottomrule
\end{tabular}
\end{table}

\subsection{The DCTS reference solutions}
\label{supp:r7-reference}

The 500 rooms share one building envelope of height 5\,m. They differ in their internal layout. Each room has 18 to 273
servers (median 109). Each server draws 400 to 2,200\,W, and the total load of a room is 13 to 537\,kW. Each room has
two to six cooling units. Their supplies impose flow rates of 0.47 to 7.4\,m$^3$\,s$^{-1}$ and temperatures of 15.0 to
24.0\,$^\circ$C. \texttt{snappyHexMesh} meshes each room with a base cell size of 0.34\,m. The mesh is refined three
levels around the racks and up to two levels at the floor features. This gives 0.88 to 1.23 million cells. The solver
integrates the steady Boussinesq RANS equations with the realizable $k$--$\varepsilon$ model. Wall functions are used
for the turbulent kinetic energy, its dissipation rate, the turbulent viscosity and the turbulent thermal
diffusivity. The laminar Prandtl number is 0.696 and the
turbulent one 0.85. Convection uses bounded first-order upwinding. Diffusion is linear, with a limited non-orthogonal
correction. Under-relaxation is 0.3 for $p'$, $k$ and $\varepsilon$ and 0.5 for $\mathbf{U}$ and $T$. Every room starts
from the same uniform field and runs 5,000 SIMPLE iterations. No room was discarded. An acceptance check on the converged fields tests three things. It requires a residual plateau and plausible value
ranges of $\mathbf{U}$, $p'$ and $T$. It also rejects interior speeds far above the boundary speeds. All 500 rooms pass
it.

Walls, partitions and the outer surfaces of racks and pillars are adiabatic no-slip walls. A cooling-unit supply imposes
its volumetric flow and temperature. Its inlet values of the turbulent kinetic energy and its dissipation rate follow from a turbulence intensity and a
mixing length. A cooling-unit return imposes the pressure, and the velocity follows the outflow. A server inlet
withdraws the server's volumetric flow from the room. The server's exhaust returns the same flow. Its temperature is the
area-weighted mean temperature of the inlet raised by $\Delta T_s=P/(\rho_{\mathrm{air}}c_p\dot V)$. An expression
boundary condition imposes this temperature at every iteration (Remark~\ref{prop:jump}). Perforated floor tiles and
panels are pairs of coupled internal faces. They let the air pass without a pressure jump.

These solutions are converged and conservative. Over the 500 rooms, the initial residuals fall by about four orders of
magnitude for $\mathbf{U}$ and $T$. For $p'$ they fall by two to three orders. All residuals are flat over the last
thousand iterations. Over that stretch, $T$ at the monitor points varies by about a thousandth of a kelvin in the median
room. This is three to four orders of magnitude below its spread over the dataset. The converged solutions also conserve
energy. We integrate the enthalpy flux $\rho_{\mathrm{air}}c_p(\mathbf{U}\cdot\mathbf{n})(T-T_0)$ over the boundary
faces. Here $T_0$ is the uniform initial temperature and $\mathbf{n}$ the outward normal. The heat that the cooling
units remove then equals the heat that the servers add to the air. The match is within 0.6 per cent in the median room
and within 6 per cent in 95 per cent of the rooms. The face values close the mass balance to about one per cent of the
supplied flow. The heat added by the servers exceeds their specified load by about four per cent. The reason is a
difference in weighting. The exhaust condition raises the area-weighted mean temperature of a server's inlet. The
enthalpy that the inlet withdraws weights its temperature with the local flux. Evaluated with area-weighted means, as
the condition is written, the exhaust reproduces every server's specified load to within two thousandths.

The paired interventions inherit a small noise floor from the solver. A base room and its interventions are solved in
separate runs. The difference of the two solutions therefore also contains the run-to-run variation of a converged
solution. To measure it, we solved the ten base rooms of the evaluation pairs again. We used their unchanged settings,
exactly as for the interventions. We then compared the two converged base solutions of each room. Their temperatures
differ by 0.003 to 0.07\,K in volume-weighted root mean square. They differ by more than 0.1\,K in at most four per cent
of a room's volume. We divide this variation by the change that each intervention causes. The result is the response
error of a model that reproduces the intervention's change exactly. Its median over the ten pairs is 0.09 for a doubled
rack and 0.02 for a halved cooling-unit flow (Table~\ref{tab:r7-reference}). This is an order of magnitude below the
response errors of every method in Table~\ref{tab:dcts-decision}. The reference therefore does not set those errors.

Finally, the solutions depend only weakly on the mesh. We meshed and solved one test room of 137 servers again from its
case specification. The base cell size was 1.26 times larger and 1.26 times smaller than the dataset's. This gives 0.81,
1.10 (the dataset) and 2.35 million cells. The solver settings were those of the dataset. The dataset's room temperature
field differs from that of the finest mesh by 0.34\,K in volume-weighted mean absolute value. The coarse field differs
from the dataset's by 0.49\,K. Against the finest mesh, the return temperatures of the cooling units differ by 0.16\,K
on the dataset's mesh and 0.40\,K on the coarse one. The mean temperatures at the server inlets differ by 1.4\,K and
1.3\,K. The room field and the returns are thus resolved to a few tenths of a kelvin. This is below the surrogates' room
errors of about half a kelvin (Table~\ref{tab:dcts-phys}). The inlet temperatures are read on small patches next to the
recirculation between the racks. They change with the mesh by as much as the surrogates' inlet errors
(Table~\ref{tab:dcts-decision}). The surrogates are trained and evaluated against the dataset's discretisation. Their
inlet errors lie at the level of that discretisation's own uncertainty. The coarse mesh has a quarter fewer cells. Yet
it took almost the same computing time as the dataset's mesh. A coarser solve is therefore no cheap substitute at this
accuracy. On the dataset's mesh, the patch-averaged $y^+$ of the walls lies between 22 and 232 (median 71). This is the
range of the wall functions.

\begin{table}[!htb]
\centering
\footnotesize
\setlength{\tabcolsep}{4pt}
\caption{Quality of the DCTS reference. The top block gives the run-to-run variation of a converged solution. It comes from the
ten base rooms of the evaluation pairs, solved again as the interventions were. The block also gives the response error
that this variation implies for a model that reproduces each intervention's change exactly (median and range over the
ten pairs of each kind). The middle block gives the global balances of the 500 converged solutions from the boundary
fluxes (median over the rooms). The bottom block compares one room on three meshes from the same case specification with
the finest mesh.}
\label{tab:r7-reference}
\begin{tabular}{P{0.52\linewidth}P{0.40\linewidth}}
\toprule
Quantity & Value \\
\midrule
$T$, two converged solutions of one room, volume-weighted RMS & 0.003--0.07\,K \\
Share of the volume where they differ by more than 0.1\,K & 0--4\% \\
Implied response error, doubled rack & 0.09 (0.02--0.22) \\
Implied response error, halved cooling-unit flow & 0.02 (0.01--0.07) \\
\midrule
Heat removed by the cooling units against heat added by the servers, 500 rooms & 0.6\% in the median room (95\% of rooms within 6\%) \\
Mass balance of the boundary face values, 500 rooms & 1\% of the supplied flow \\
\midrule
Cells: coarse / dataset / fine & 0.81 / 1.10 / 2.35 million \\
Room $T$, mean absolute: dataset against fine / coarse against dataset & 0.34 / 0.49\,K \\
Return temperatures against the fine mesh: coarse / dataset & 0.40 / 0.16\,K \\
Server-inlet temperatures against the fine mesh: coarse / dataset & 1.3 / 1.4\,K \\
Wall $y^+$ on the dataset's mesh, patch averages & 22--232 (median 71) \\
\bottomrule
\end{tabular}
\end{table}

\end{document}